\documentclass{article}
\pdftrailerid{}
\usepackage{microtype}
\usepackage{graphicx}
\usepackage{subcaption}
\usepackage{booktabs} %

\usepackage{hyperref}

\usepackage[preprint]{icml2026}

\usepackage{amsmath,amssymb,amsthm,mathtools}
\usepackage{adjustbox}
\usepackage{array}
\usepackage{enumitem}
\usepackage[labelsep=period,font=small]{caption}
\usepackage{tikz}
\usetikzlibrary{arrows.meta,decorations.pathreplacing,calc}

\usepackage[capitalize,noabbrev]{cleveref}

\theoremstyle{plain}
\newtheorem{theorem}{Theorem}[section]
\newtheorem{proposition}[theorem]{Proposition}
\newtheorem{lemma}[theorem]{Lemma}
\newtheorem{corollary}[theorem]{Corollary}

\theoremstyle{definition}

\newtheorem{observation}[theorem]{Observation}
\theoremstyle{remark}
\newtheorem{remark}[theorem]{Remark}

\newcommand{\R}{\mathbb{R}}

\newcommand{\Z}{\mathbb{Z}}

\newcommand{\E}{\mathbb{E}}
\newcommand{\Id}{\mathrm{Id}}
\newcommand{\sd}{\operatorname{sd}}
\newcommand{\sign}{\operatorname{sign}}

\newcommand{\pmax}{p_{\max}}

\newcommand{\avg}[1]{\langle #1\rangle}

\usepackage[textsize=tiny]{todonotes}

\icmltitlerunning{When Does Exploration Prevent Collapse?}

\begin{document}

\twocolumn[
  \icmltitle{When Does Exploration Prevent Collapse?\texorpdfstring{\\}{ }Three Regimes and the Path of the Reward in Self-Consistency Training}

  \begin{icmlauthorlist}
    \icmlauthor{Jaeho Hwang}{a}
    \icmlauthor{Uiwon Hwang}{b}
  \end{icmlauthorlist}

  \icmlaffiliation{a}{Human-Centered Artificial Intelligence Research Institute, Ewha Womans University}
  \icmlaffiliation{b}{Department of Computer Science and Engineering, Ewha Womans University}

  \icmlcorrespondingauthor{Uiwon Hwang}{uiwon.hwang@ewha.ac.kr}

  \icmlkeywords{Machine Learning, ICML}

  \vskip 0.3in
]

\printAffiliationsAndNotice{}  %

\begin{abstract}
Test-time reinforcement learning (TTRL) adapts a model without labels by rewarding answers that agree with the majority, but it collapses the policy onto its majority answer and can remove the correct answer from the samples. Exploration can prevent this collapse, yet its strength is chosen by trial and error, and how the training algorithm changes the required strength is unclear. We model training on a single problem as a continuous flow of the logits of a softmax policy and analyze which policies are stable as the exploration coefficient varies. Only collapsed policies are stable below a lower boundary and only policies spread over many answers above an upper boundary, and we derive these boundaries in closed form for an agreement reward with smoothing toward the uniform policy, compute them for the majority reward of TTRL on policies with one leading answer and bound them with interval arithmetic for actions on ordered bins. We prove that under exact gradients, the learning rate and, in a mean-field model, the normalization of GRPO leave these boundaries unchanged when exploration passes through them together with the reward but shift them when exploration bypasses them. Exact sums on these policies then show that a bonus for rare answers needs a coefficient above $1/\log(G-1)$ for groups of $G$ samples before the normalization and about two to three times that after it. In free-form TTRL on MATH-$500$, a bonus placed before the normalization prevented collapse and kept more correct answers among the samples, although it lowered pass@1.
\end{abstract}

\section{Introduction}\label{sec:intro}

Reinforcement learning with verifiable rewards, often implemented with GRPO \citep{shao2024deepseekmath}, has improved the reasoning of large language models, but it requires a label or a verifier for every problem. Test-time reinforcement learning (TTRL) removes this requirement by sampling a group of answers to each problem and rewarding the answers that agree with the majority of the group \citep{zuo2025ttrl}. Related methods reward agreement, low entropy or self-certainty \citep{zhang2025empo,agarwal2025em,prabhudesai2025rent,zhao2025intuitor}, and vision-language-action (VLA) models, which predict each action as one of many ordered bins that split its range, are adapted with rewards computed from their own rollouts \citep{t2vla2026}. Because these self-consistency rewards need neither labels nor a verifier, they can adapt a model on the test problems themselves.

However, a self-consistency reward favors what the policy already prefers, so training collapses the policy onto its initial majority \citep{huang2025sharpening,he2026unsupervised,corderoencinar2025certified}. When the majority is wrong, the correct answer can disappear from the samples early in training \citep{lin2026extinction}, and even a verifier or a selector that chooses among the samples can no longer recover it \citep{shafayat2025srt,zhang2025nofreelunch}. Existing methods try to prevent collapse by adding entropy bonuses or novelty rewards, raising the sampling temperature, applying KL penalties or modifying the normalization of advantages \citep{evolrl2025,liu2025ettrl,liu2025drgrpo}. The effect of these terms also depends on the training algorithm, since GRPO divides the advantages by the standard deviation of the rewards in each group, and this division has been shown to change the effect of a KL penalty \citep{vojnovic2025grpo} and of shaping rewards \citep{wang2026darkroom}.

The settings of these methods, such as the coefficient of a bonus or the temperature, are chosen by trial and error for each model and task. Even for a single problem, it is still unclear which of these methods keep the policy spread over many answers, how strong they must be to prevent collapse and how the training algorithm changes that strength. For example, raising the sampling temperature changes how fast the policy collapses but not whether it collapses. Moreover, a bonus for rare answers with a fixed coefficient prevents collapse when it is added to the reward before the division of GRPO, but not when it is added after the division (Figure~\ref{fig:ttrl}). Exploration here means any term in the reward or the loss that keeps the policy spread over many answers.

\begin{figure*}[t]
\centering
\includegraphics[width=0.9\textwidth]{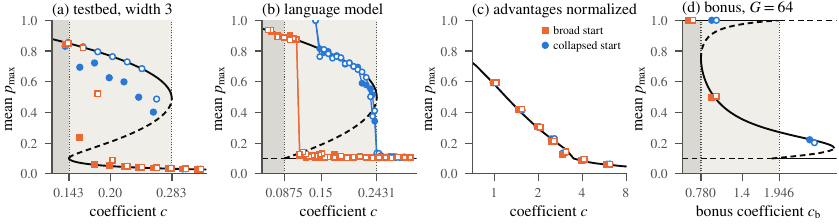}
\caption{Regimes in four settings, with no window in (c). Lines are stable (solid) and unstable (dashed) equilibria of the tabular flow, with shading where collapse cannot be prevented (dark) and in the window (light). Markers give the mean $\pmax$ of a run from broad (squares) and collapsed (circles) starts, filled and open for SGD and AdamW in (a) and (c) and for seeds 0 and 1 otherwise. (a) Testbed, Gaussian kernel of width 3 in the loss, with the folds of Theorem~\ref{thm:inloss-cap} times $\frac{62}{64}$ for the leave-one-out reward. (b) Letter test, agreement reward, coefficient ramped over 900 steps, with the ends of Proposition~\ref{prop:potts-uniform} times $\frac{14}{16}$. (c) As (a) with only the advantages normalized, against the coefficient before the division (Proposition~\ref{prop:norm}(b)). (d) Letter test, majority reward, bonus after the normalization, $G=64$, with the end of the interior branch and the threshold near collapse.}
\label{fig:regimes}\label{fig:phase}\label{fig:llm-ramp}
\end{figure*}

In this paper, we answer these questions with the tabular flow, the dynamics of the answer distribution of a single problem under the expected gradient of a softmax policy. We analyze which policies of this flow are stable as the exploration coefficient varies and find a lower and an upper boundary of the coefficient that divide it into three regimes (Figure~\ref{fig:regimes}). Below the lower boundary, only policies collapsed onto one answer are stable, above the upper boundary, only policies spread over many answers are stable, and between the two, the initial policy decides the outcome. We derive these boundaries in closed form when the reward is the probability that another sample agrees and exploration smooths the policy toward the uniform policy, compute them exactly for the majority reward of TTRL on policies with one leading answer and bound them for actions on ordered bins by computer-assisted proofs, in which interval arithmetic encloses every rounding error. We also prove that exploration treating all answers alike keeps the order of their probabilities, so it can keep the correct answer among the samples but cannot correct a wrong majority.

These boundaries also depend on where exploration enters the update. The reward reaches the parameters through the policy gradient, the optimizer and the division of GRPO, which together form the path of the reward. We prove that when exploration passes along this whole path, the equilibria that set the boundaries are the stationary points of the training objective, so under exact gradients the learning rate and the scaling of the optimizer do not move them. A term that bypasses one of these steps instead has a threshold that changes with the learning rate or with the policy. For example, exact sums on policies with one leading answer show that a bonus for rare answers prevents collapse above $1/\log(G-1)$ for groups of $G$ samples when it is added before the division of GRPO, but it needs a coefficient about two to three times larger when it is added after the division.

We test these predictions on a small network with ordered actions, on Qwen2.5-1.5B-Instruct answering multiple-choice questions with a single letter and on free-form TTRL with Qwen2.5-Math-1.5B on MATH-500, three settings that come progressively closer to practice. Across all three, the thresholds stayed fixed when exploration followed the path of the reward and moved when it did not. In free-form TTRL, the measured threshold of the bonus lay around its predicted value both before and after the division of GRPO. The bonus kept more correct answers among the samples and raised pass@32, although it lowered pass@1 compared with TTRL without the bonus, which concentrates the policy on its majority answer.

The main results are summarized as follows.

\begin{itemize}
    \item The exploration coefficient has a lower boundary, below which only collapsed policies are stable, and an upper boundary, above which only policies spread over many answers are stable.
    
    \item Under exact gradients, exploration that passes along the whole path of the reward keeps these boundaries unchanged by the learning rate and the scaling of the optimizer, whereas exploration that bypasses a step does not. On policies with one leading answer, the division of GRPO raises the coefficient that a bonus for rare answers needs by a factor of about two to three.

    \item In a network testbed, a language model with single-letter answers and free-form TTRL, the thresholds followed these predictions, and in free-form TTRL, a bonus placed before the division prevented collapse and kept more correct answers among the samples.
\end{itemize}

\section{Related work}\label{sec:related}

\paragraph{Self-referential rewards.}
TTRL turns self-consistency decoding \citep{wang2023selfconsistency} into a training signal \citep{zuo2025ttrl}. Methods against its collapse add novelty or entropy terms \citep{evolrl2025,liu2025ettrl,du2026dare,wu2025spine,ding2026collective}, weaken the self-reference \citep{zhang2025corewarding}, filter weak consensus \citep{yan2026scrl}, intervene before correct answers disappear from the samples \citep{lin2026extinction} or co-train a verifier \citep{pan2026coverrl}. Entropy also collapses with verifiable rewards \citep{cui2025entropy}.

\paragraph{Rankers and selectors.}
RISC ranks sampled answers with a learned model \citep{marina2026risc}, and CoRE takes the pseudo-label of TTRL from the dominant set of a graph of rollouts \citep{mehrish2026core}. Such selectors can recover a correct answer that the majority misses, but only while it remains among the samples. Exploration keeps such answers without choosing among them (Proposition~\ref{prop:order}).

\paragraph{Normalization.}
For verifiable rewards, \citet{bay2026identity} treat GRPO, Dr.\ GRPO \citep{liu2025drgrpo} and DAPO \citep{yu2025dapo} as three operations on the standard deviation of the group, as do our arms R, U and B (Section~\ref{sec:exploration}). REINFORCE++ normalizes advantages over the whole batch instead of the group \citep{hu2025reinforcepp}, and PAPO normalizes a rubric reward apart from the outcome reward \citep{tan2026papo}. Appendix~\ref{app:related} discusses a bonus placed before or after the division. \citet{wang2026darkroom} shows that in groups that all fail, the normalization makes the advantages of a shaping reward independent of its coefficient, which collapses agents, while the same signal helps as an auxiliary loss. For a KL penalty, \citet{vojnovic2025grpo} find that the division turns its coefficient into one proportional to the standard deviation of the rewards. This deviation falls as the policy concentrates, which is the mechanism of our arm R. Near collapse, a group of self-consistency rewards takes two values, whose gap the normalization keeps only in sign, so a bonus acts through a threshold that depends on where it enters.

\paragraph{Selection with mutation.}
Reinforcement learning over a fixed set of actions approximates replicator dynamics, a gradient flow in the Shahshahani metric \citep{borgers1997learning,hennes2020nrd,shahshahani1979,hofbauer1998,kerimkulov2025}. Under natural policy gradient, Proposition~\ref{prop:potts-uniform} is the selection-mutation equation of \citet{hofbauer1985selection} for the pure coordination game with uniform mutation, for which that work gives a potential. Its thresholds are weak-selection limits of those of \citet{komarova2001grammar}, whose fixed points \citet{mitchener2003bifurcation} classifies. The window of an entropy bonus is bounded by the two spinodals of the mean-field Potts model \citep{costeniuc2005potts,cuff2012potts}. Beyond these results, we analyze the path of the reward, compute the thresholds of the majority reward and of a bonus on either side of the normalization, certify the folds on ordered bins (Theorem~\ref{thm:inloss-cap}) and show which stability transfers to networks (Proposition~\ref{prop:unordered-networks}). In the vanilla geometry, Theorem~\ref{thm:inloss-index} shows that every equilibrium has the stability it has under natural policy gradient.

\section{Self-consistency rewards and exploration}\label{sec:setup}

\subsection{Policies, rewards and gradients}

The action space is either $K$ ordered bins or $n$ unordered answers. Ordered bins lie on a ring or on a line with reflecting ends, where out-of-range actions are clipped to the end bins as in tokenized VLA models. A policy is $p=\mathrm{softmax}(z)$ with logits $z$, and we write $S_k=\sum_ip_i^k$ and $\avg{f}_p=\sum_ip_if_i$.

\paragraph{Rewards.}
The agreement reward $R_i=p_i$ is the probability that an independent sample equals $i$, with the advantage $p_i-S_2$. In a group of $G$, a sample gets the fraction of the other $G-1$ samples that agree. The majority reward of TTRL rewards the samples equal to the most frequent one, with ties broken at random. Both rewards depend only on the answer counts of the group, symmetrically in the answers.

\paragraph{Two gradient geometries.}
We write training as a flow $z'=V$ of the logits, with the reward held fixed during differentiation. Natural policy gradient (NPG) \citep{kakade2001natural} follows the Fisher metric and gives $V_i=R_i-\avg{R}_p$ up to a constant shift, and vanilla policy gradient gives $V_i=p_i(R_i-\avg{R}_p)$. For the agreement reward, both are covered by
\begin{equation}\label{eq:qfamily-lattice}
V_i=p_i^{\,q}(p_i-S_2),
\end{equation}
with $q=0$ for NPG and $q=1$ for vanilla policy gradient. Each flow is the small-step limit of plain gradient steps $z\leftarrow z+\eta V$, as taken by SGD.

\subsection{The path of the reward}\label{sec:exploration}

\paragraph{A leak.}
In a leak, the policy is mixed with its neighbors after each step, $p_i\leftarrow(1-\deg(i)\,\varepsilon)p_i+\varepsilon\sum_{j\sim i}p_j$. As $\eta\to0$ with $\varepsilon=\omega\eta$, the algorithm converges to the lattice equation
\begin{equation}\label{eq:lattice}
\partial_tp_i=\omega(\Delta p)_i+p_i\big(V_i-\avg{V}_p\big),\qquad \omega=\frac\varepsilon\eta,
\end{equation}
where $(\Delta p)_i=\sum_{j\sim i}(p_j-p_i)$ is the graph Laplacian (Appendix~\ref{app:proofs-setup}).

\paragraph{A term of the loss.}
In the loss, exploration can enter as the term $c\cdot\mathrm{CE}(Dp_{\rm old},p)$, which pulls the policy toward its own smoothed version, with $p_{\rm old}$ the current policy held fixed and $D$ a symmetric stochastic kernel. On ordered bins, we use $D=\Id+\alpha\Delta$ with $0<\alpha\le\frac12$ and the Gaussian kernel $D_w$ of width $w$ bins, normalized to unit row sums and mirrored at reflecting ends. On unordered answers, $D=\frac1n\mathbf1\mathbf1^\top$ pulls the policy toward the uniform policy $\bar p$. Under NPG, the term is the leak $c(D-\Id)p$ of the probabilities. Under vanilla policy gradient, it acts on the logits, and Lemma~\ref{lem:explore-geometry} gives the lattice equation of the objective
\begin{equation}\label{eq:inloss}
z'=\mathcal V(z),\qquad \mathcal V=p\odot(p-S_2)+c\,(Dp-p) .
\end{equation}

\paragraph{Three transformations of the reward.}
The geometry turns the reward into a logit velocity, and the optimizer multiplies the gradient by a positive definite preconditioner $A(\theta)$, the learning rate times the identity for SGD and a coordinatewise scaling for Adam. For the leave-one-out agreement reward, the reward of one sample has the variance
\begin{equation}\label{eq:sigmaG}
\varsigma_G(p)^2=\big(S_3-S_2^2\big)+\frac{S_2-S_3}{G-1} .
\end{equation}
We model the normalization of GRPO by dividing the selection term by $\varsigma_G(p)$. Monte Carlo simulations confirm this model to within 4 percent on our ordered branches up to $\pmax=0.94$, about 10 percent near collapse and 11 percent for ten answers (Appendix~\ref{app:norm}). Exploration is \emph{on the path of the reward} if it passes through all three transformations, and \emph{off the path} otherwise. A term of the loss passes through the geometry and the preconditioner, and through the normalization if it is divided by the same standard deviation. A leak bypasses the learning rate and every preconditioner, and, under vanilla policy gradient, also the geometry.

\paragraph{Arms.}
We compare four placements of exploration, which we call arms U, R, B and C. Arm U leaves everything unnormalized, as in Dr.\ GRPO, and arm R divides the reward advantages but not the exploration term, as GRPO treats its KL penalty. Arm B divides both and drops groups with equal rewards, as the filter of DAPO does. Arm C adds a bonus to the reward before the division, as DARE adds its bonus for rare answers to the reward that GRPO normalizes \citep{du2026dare}.

\subsection{Collapse and three regimes}\label{sec:collapse-def}

An equilibrium is \emph{collapsed} if one bin holds at least $M_0=\frac38$ on ordered bins, a level between the branches of Theorem~\ref{thm:inloss-cap}, and \emph{committed} if one answer leads for unordered answers. Stable committed equilibria approach one answer as $c\to0$ (Proposition~\ref{prop:potts-uniform}(b)), and we speak of collapse for collapsed and committed equilibria alike. Other equilibria, such as the uniform policy and peaks as wide as the kernel, are \emph{broad}. An equilibrium is \emph{linearly stable} if the linearization of the flow, restricted to directions that change the policy, has its spectrum in the open left half-plane.

The \emph{lower boundary} is the smallest coefficient at which a broad equilibrium is stable, and the \emph{upper boundary} the largest at which a collapsed one is. They give three regimes. Below the lower boundary, only collapsed equilibria are stable, and above the upper boundary, only broad ones are. In the \emph{window} between them, which can be empty, both can be. A \emph{fold} is a coefficient at which a branch of equilibria turns back. For the Gaussian kernels of Theorem~\ref{thm:inloss-cap}, a stable equilibrium meets one with one unstable direction at the folds $c^*$ of the broad branch and $c_{\rm rel}$ of the collapsed branch. Stable ones exist just above $c^*$ and just below $c_{\rm rel}$, so the lower boundary is at most $c^*$, and the upper boundary is at least $c_{\rm rel}$. \emph{Prevention} keeps a broad start from collapsing, and \emph{release} undoes the collapse of a collapsed start. In the tabular flow, prevention thus needs at least the lower boundary, and release follows above the upper one. Under a bonus, collapse means that one answer takes all the probability, and other equilibria are interior. The experiments count a state as collapsed above a cutoff of its largest probability $\pmax$ between its broad and collapsed states, $0.2$ or $0.3$, and $0.95$ under a bonus, with two exceptions (Appendix~\ref{app:experiments}). A free-form problem is collapsed when one answer holds at least 95 percent of its samples with an answer.

\section{Regimes on the path of the reward}\label{sec:theory}\label{sec:inloss}

Proofs are in the appendix, and claims without proof are called Observations. Table~\ref{tab:status} in Appendix~\ref{app:proofs} gives the status of each claim and where it is proved or tested.

\subsection{Thresholds that belong to the objective}\label{sec:potential}

\begin{theorem}[Exploration in the loss has a potential]\label{thm:inloss-main}
Let $c>0$, let $D$ be symmetric, irreducible and stochastic with nonnegative entries, and let $F_c(p)=\frac13S_3-\frac14S_2^2-\frac c4\sum_{i,j}D_{ij}(p_i-p_j)^2$. Then the probability flow induced by \eqref{eq:inloss} is the gradient flow of $F_c$ on the open simplex in the Shahshahani metric, and NPG trained on the same objective is its Euclidean gradient flow. Every solution of either flow converges to a single equilibrium. The equilibria are the zeros of the logit velocity and are the same for both flows. An equilibrium is linearly stable for one flow if and only if it is for the other, and this holds if and only if $\nabla^2F_c$ is negative definite there.
\end{theorem}

\begin{lemma}[Preconditioners do not move equilibria]\label{lem:fixedpoints}
Let the logits $z(\theta)$ of a network follow $\theta'=A(\theta)\,(\partial_\theta z)^\top\mathcal V(z(\theta))$, with $A(\theta)$ symmetric positive definite, as for SGD and the stationary scaling of Adam. If $\partial_\theta z$ has full row rank, then $\theta$ is stationary if and only if $\mathcal V(z(\theta))=0$.
\end{lemma}

The lemma excludes momentum and Adam's dependence on sampled gradients (Appendix~\ref{app:updates}).

\begin{proposition}[Normalization]\label{prop:norm}
Let $\varsigma$ be a positive, continuously differentiable function on the open simplex that is bounded above, such as $\varsigma_G$ of \eqref{eq:sigmaG}.
\begin{enumerate}[label=(\alph*),leftmargin=*,itemsep=1pt]
\item If both terms are divided by $\varsigma$, $z'=\mathcal V/\varsigma$, the equilibria are those of \eqref{eq:inloss}, each with the same number of unstable directions, and every solution converges to an equilibrium.
\item If only the selection term is divided by $\varsigma$, $z'=p\odot(p-S_2)/\varsigma(p)+c\,(Dp-p)$, the equilibria at the coefficient $c$ are the equilibria of \eqref{eq:inloss} at the coefficient $c\,\varsigma(p)$.
\end{enumerate}
\end{proposition}

Hence, with exact gradients, the equilibria and folds of exploration in the loss do not move with the learning rate or with a preconditioner that depends on the parameters alone. Nor do they move with a normalizer of both terms that depends on the policy alone. A network need not descend $F_c$ and keeps the count of unstable directions only in the cases of Section~\ref{sec:regimes}.

\subsection{Exact regimes for unordered answers}\label{sec:unordered}

\begin{proposition}[Smoothing toward the uniform policy]\label{prop:potts-uniform}
Let $n\ge3$ and $D=\frac1n\mathbf1\mathbf1^\top$, so that \eqref{eq:inloss} reads $z'=p\odot(p-S_2)+c\,(\bar p-p)$. In either geometry, the following hold.
\begin{enumerate}[label=(\alph*),leftmargin=*,itemsep=1pt]
\item The uniform policy is linearly stable if $c>\frac1n$ and linearly unstable if $c<\frac1n$.
\item The equilibria with one entry $M$ larger than the $n-1$ others form the curve $c=\frac{n}{n-1}M(1-M)$, $\frac1n<M<1$. They exist if and only if $c\le\frac n{4(n-1)}$. They are linearly stable if $M>\frac12$ and have exactly one unstable direction if $M<\frac12$. Every other nonuniform equilibrium is linearly unstable.
\item If $c>\frac n{4(n-1)}$, every solution converges to $\bar p$. If $c<\frac1n$, the only linearly stable equilibria are the $n$ equilibria of (b) with $M>\frac12$, and almost every solution reaches one.
\end{enumerate}
\end{proposition}

\begin{corollary}[Collapse keeps the initial plurality]\label{cor:plurality}
In the setting of Proposition~\ref{prop:potts-uniform}, let $c<\frac1n$. Then almost every solution converges to the stable equilibrium of Proposition~\ref{prop:potts-uniform}(b) whose leading answer has the largest initial probability.
\end{corollary}

Below the lower boundary, the tabular flow with uniform smoothing thus commits to the initial plurality, as observed for training without labels \citep{he2026unsupervised}.

The majority reward of TTRL keeps the three regimes on policies with one leading answer and equal others. For any reward that depends only on group counts, symmetrically in the answers, these policies form an invariant set, where exact sums give the boundaries (Proposition~\ref{prop:symmetric}). For the majority reward with $n=10$ and $G=16$, they are $0.363$ and $0.766$ without normalization, against $0.0875$ and $0.2431$ for the leave-one-out agreement reward, which scales every threshold by $\frac{G-2}G$ (Appendix~\ref{app:majority-exact}). Dividing only the advantages multiplies the release fold by $2.07$ for the majority reward and by $6.55$ for the agreement reward, so a coefficient tuned for one reward does not transfer.

Inside the window, the outcome depends on the start, and on policies with one leading answer and equal others, a start with leading mass $M(0)<\frac12$ is kept from committing exactly when $c>\frac n{n-1}M(0)(1-M(0))$, and a start with $M(0)\ge\frac12$ only above the upper boundary (Proposition~\ref{prop:potts-invariant}). The coefficient that prevents commitment thus rises from $\frac1n$ at the uniform policy to $\frac n{4(n-1)}$ once the leader holds half of the probability, so prevention costs least early.

\subsection{Certified folds on ordered bins}\label{sec:twoscales}

On a ring, the uniform policy is stable only above $c_{\rm lin}\approx\frac{K}{2\pi^2w^2}$, about $1.44$ for $K=256$ and $w=3$, so the broad states that matter are peaks (Proposition~\ref{prop:inloss-uniform}).

\begin{theorem}[Computer-assisted folds]\label{thm:inloss-cap}
Let $D=D_w$ with $w\in\{1,\frac32,3,4,8\}$ on the rings and lines of 128 to 512 bins of Table~\ref{tab:inloss-cap}. Then \eqref{eq:inloss} has nondegenerate folds of site-symmetric equilibria, symmetric about one bin, at $c^*$ and at $c_{\rm rel}$, for example at $c^*=0.147749\ldots$ and $c_{\rm rel}=0.291945\ldots$ for $w=3$. At each fold, a linearly stable equilibrium meets one with exactly one unstable direction. Near the lower fold, these equilibria exist only for $c\ge c^*$, and the stable one is a broad peak. Near the upper fold, they exist only for $c\le c_{\rm rel}$, and the stable one is a spike.
\end{theorem}

The certificate is local and covers states symmetric about one bin (Appendix~\ref{app:guide-cap}). Those centered between two bins are unstable to translation in double precision (Table~\ref{tab:bond}). That broad data collapse below $c^*$ and reach the certified branch above it is numerical (Appendix~\ref{app:inloss-dial}).

\subsection{Boundaries and networks}\label{sec:regimes}

For the agreement reward on unordered answers, the boundaries $\frac1n$ and $\frac n{4(n-1)}$ are proved. For a Gaussian kernel, Theorem~\ref{thm:inloss-cap} shows that the lower boundary is at most $c^*$ and the upper boundary at least $c_{\rm rel}$, and numerically, both equal the folds. For the majority reward, exact sums give them on policies with one leading answer and equal others, as for the bonus, whose threshold near collapse comes from the sign of a drift.

With exploration in the loss and exact gradients, full-rank networks share the equilibria and folds (Lemma~\ref{lem:fixedpoints}). For unordered answers, they also share the stability of equilibria with at most two distinct entries (Proposition~\ref{prop:unordered-networks}). With the agreement reward above $\frac n{4(n-1)}$, every stationary point gives every state the uniform policy (Corollary~\ref{cor:upper}). On ordered bins, a preconditioner can destabilize a broad peak, so the lower boundary can depend on the optimizer (Observation~\ref{obs:ordered-precond}). None of this covers free-form answers.

\section{Exploration off the path of the reward}\label{sec:offpath}\label{sec:dichotomy}

\subsection{Bypassing the learning rate changes the unit}\label{sec:npg}

Take the smoothing $\Delta$ on a line of $K$ bins with reflecting ends. In the loss, with $D=\Id+\alpha\Delta$, its threshold is a coefficient $c$ of the objective. Applied as a leak, the smoothing leaves the uniform policy linearly stable for $\omega>\omega_{\rm lin}(K)=K^{-(q+1)}/\lambda_1^N$, with $\lambda_1^N=4\sin^2\frac\pi{2K}$, and unstable below (Theorem~\ref{thm:D5}). Every threshold of this flow lies in $\omega=\varepsilon/\eta$, so the mixing $\varepsilon$ that prevents collapse scales with the learning rate. Under NPG, the term of the loss is this leak with $\omega=\alpha c$, whose linear threshold is $c\approx52$ for $K=256$ and $\alpha=\frac12$ (Appendix~\ref{app:npg-dichotomy}).

\subsection{Bypassing the normalization changes the structure}\label{sec:statedep}\label{sec:vanilla}\label{sec:adam-outside}\label{sec:adam}

The normalization of GRPO rescales the reward by a factor that depends on the policy, as do the geometry of vanilla policy gradient and the preconditioner of Adam. Exploration that bypasses such a factor is rescaled relative to selection along a branch of equilibria.

\paragraph{The normalization.}\phantomsection\label{sec:norm-reward}
By Proposition~\ref{prop:norm}(b), dividing only the selection term by $\varsigma_G$ turns $c$ into $c\,\varsigma_G(p)$, which changes along a branch of equilibria. In the mean-field model, the folds of the site-symmetric branch then vanish for groups of 32 or more at $w=3$ and for every computed group at $w=\frac32$ (Observation~\ref{obs:unfold}). Normalizing only the reward thus removes the window for large groups, so prevention and release meet, though the narrow windows left for smaller groups are comparable to the error of the model. For unordered answers, the window survives but moves up (Appendix~\ref{app:norm-unordered}). A KL penalty or an entropy term in the loss of GRPO bypasses the division in the same way. Its effective coefficient thus changes with the policy, as \citet{vojnovic2025grpo} found for the KL penalty.

\paragraph{An answer-level bonus.}
The same factor sets the coefficient that a bonus for rare answers needs, depending on where it enters. A natural bonus gives sample $j$ the value $b_j=-\log(N_{a_j}/G)$, where $N_{a_j}$ counts the samples of its answer. Near collapse, a group with nonzero advantages has, to leading order, $G-1$ samples of the leading answer and one other. In arms C and U, the bonus with coefficient $c_{\rm b}$ gives the other sample an advantage of the sign of $c_{\rm b}\log(G-1)-1$. In arm R, with the population standard deviation of our runs, it gives the sign of $c_{\rm b}\log(G-1)-G/\sqrt{G-1}$. To leading order near collapse, the leading answer therefore gains mass below $c_{\rm b}^*=1/\log(G-1)$ in arms C and U and below a threshold $G/\sqrt{G-1}$ times higher in arm R, whatever the space of answers (Appendix~\ref{app:bonus}). This is the sign of a drift on policies with one leading answer, not a proof of linear stability. With the unbiased standard deviation of the TTRL code, the factor is $\sqrt G$. Before the division, one threshold $c_{\rm b}^*$ governs broad and collapsed policies alike, and a bonus just above it leaves both at an interior equilibrium. After the division, exact sums on policies with one leading answer show that a broad policy needs only the end of the branch of interior equilibria, which lies below the threshold of arm R near collapse, at $2.8$ to $3.2$ times $c_{\rm b}^*$ for ten answers and $2.3$ to $2.7$ times for free-form answers (Observation~\ref{obs:bonus-window}), so the normalization splits one threshold into two (Table~\ref{tab:bonus-main}).

\begin{table}[t]
\centering\scriptsize
\caption{Thresholds of the bonus for rare answers on policies with one leading answer, from exact sums with the population standard deviation. Before the division, one threshold serves broad and collapsed policies. After it, a broad policy needs the end of the branch of interior equilibria, for ten or for free-form answers, and a collapsed one the threshold near collapse (Appendix~\ref{app:bonus}).}
\label{tab:bonus-main}
\begin{tabular}{@{}lllll@{}}
\toprule
 & $G=8$ & $G=16$ & $G=32$ & $G=64$ \\
\midrule
before, $c_{\rm b}^*=1/\log(G-1)$ & $0.514$ & $0.369$ & $0.291$ & $0.241$ \\
after, broad, ten answers & $1.454$ & $1.074$ & $0.876$ & $0.780$ \\
after, broad, free-form answers & $1.375$ & $0.938$ & $0.705$ & $0.565$ \\
after, near collapse & $1.554$ & $1.526$ & $1.674$ & $1.946$ \\
\bottomrule
\end{tabular}
\end{table}

\section{Experiments}\label{sec:experiments}

Each test compares measured thresholds with the values that the theory predicts, and Appendix~\ref{app:experiments} gives the full settings and every run.

\subsection{A network testbed}\label{sec:testbed}\label{sec:discriminating}

A tanh network maps 32 states to logits over $K=256$ ordered bins and takes PPO steps on the leave-one-out agreement reward with $G=64$ samples per state (Appendix~\ref{app:testbed}). The threshold of the term $c\cdot\mathrm{CE}(D_wp_{\rm old},p)$ in the loss varied by a factor of at most $1.09$ over learning rates up to tenfold apart for SGD and AdamW and $w\in\{1,1.5,3\}$. As a separate SGD step, the same smoothing followed the unit law $\omega=\varepsilon/\eta$ to within 7 percent. For $w\ge1.5$, the thresholds lay 3 to 29 percent above the folds, more for AdamW, whose steps of fixed size let the policy escape from the broad equilibrium inside the window. A grid at $w=3$ found the window in five seeds (Figure~\ref{fig:regimes}a). With only the advantages normalized, the starts ended within $0.024$ in mean $\pmax$ at every $c$ in four seeds (Figure~\ref{fig:regimes}c), and dividing exploration too restored the window (Appendix~\ref{app:testbed-norm}).

\subsection{A language model with single-letter answers}\label{sec:llm}

In the letter test, Qwen2.5-1.5B-Instruct \citep{qwen25report} answers 64 ten-option questions of MMLU-Pro \citep{wang2024mmlupro} by a single letter, whose distribution we compute exactly. Each step samples 16 letters for each of 32 of the questions and takes PPO steps with AdamW on LoRA parameters \citep{hu2022lora} (Appendix~\ref{app:llm-test}). The questions share the adapter and mostly committed onto the same letter, so each run is one sample of one shared softmax. For the agreement reward, slow ramps released the collapsed start at the fold $0.2431$ or within one evaluation of it in four seeds (Figure~\ref{fig:regimes}b). At fixed coefficients, release came below the fold, which the parameters shared across questions may explain. For the majority reward, dividing only the advantages moved release by a factor of $1.89$ to $2.13$ between arms R and U, around the predicted $2.07$ and far from $6.55$ for the agreement reward.

\paragraph{An answer-level bonus across group sizes.}
With the majority reward and the bonus for $G$ from 8 to 64, broad starts ran in two seeds at $0.8$ and $1.18$ to $1.25$ times the threshold predicted for each arm (Appendix~\ref{app:letter-bonus}). Every run collapsed at the lower coefficient and none at the upper one. For $G\ge16$, arm R thus kept broad starts from collapsing at $0.50$ to $0.88$ times its threshold near collapse, as the end of its interior branch predicts. From the collapsed adapter at $G=16$ and $64$, arm R stayed collapsed where its broad start did not and released above its threshold near collapse (Figure~\ref{fig:regimes}d). Its release brackets fit the factor $G/\sqrt{G-1}$ but do not resolve it.

\subsection{Free-form test-time reinforcement learning}\label{sec:ttrl}

\begin{figure}[t]
\centering
\includegraphics[width=\linewidth]{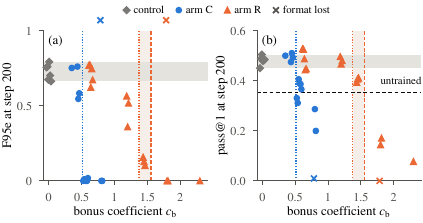}
\caption{Free-form TTRL with groups of 8, every run at step 200 against the coefficient of the bonus. (a) F95e, the share of covered problems collapsed onto one answer. (b) Pass@1, with the untrained model dashed. Vertical lines mark the threshold $0.514$ of arm C and, for arm R, the end $1.375$ of its interior branch (dotted) and its threshold $1.554$ near collapse (dashed), with the window shaded. The gray band spans the controls, and the arms are offset at $0.6$. Crosses mark the two runs that lost the answer format, drawn above the axis in (a).}
\label{fig:ttrl}
\end{figure}

Qwen2.5-Math-1.5B \citep{qwen25math} solves 128 MATH-500 problems \citep{hendrycks2021math,lightman2024verify} in free form, chosen with labels so that half of the initial pluralities are wrong (Appendix~\ref{app:ttrl}). On this subset, TTRL did not raise majority accuracy, so our numbers do not compare with published ones. Each step samples $G=8$ solutions for each of 16 problems, rewards the most frequent boxed answer and takes clipped policy-gradient steps without a critic, as in GRPO, with AdamW at $10^{-4}$ on LoRA parameters. Unlike the original TTRL \citep{zuo2025ttrl}, training has no KL term, starts from an adapter trained on the answer format and divides by the population standard deviation. Samples without an answer get the advantage $-0.1$ outside the group statistics. Every 20 steps, we sample 32 solutions per problem at temperature 1. F95e counts only problems whose leading answer held less than 90 percent at step 0. It is the share of collapsed problems among the covered ones, at least 90 percent of whose samples have an answer.

Without exploration, F95e reached $0.66$ to $0.79$ at step 200 in six seeds. At $c_{\rm b}=0.6$, or $1.17$ times its threshold $0.514$, arm C kept F95e at or below $0.065$ from step 100 on in all six seeds, with $0.000$ to $0.019$ at step 200. Arm R at the same coefficient, $0.44$ times the end $1.375$ of its interior branch, collapsed in its five seeds, with $0.62$ to $0.77$. In some seeds, fewer samples ended with a boxed answer as training went on, but 62 to 98 percent of eligible problems remained covered at step 200, and the contrast between the two placements held in every seed.

Varying the coefficient over three seeds placed the threshold of arm C in $(0.45,0.6]$, around the predicted $0.514$, and that of arm R in $(1.2,1.8]$ (Figure~\ref{fig:ttrl}, Appendix~\ref{app:sweep}). Their ratio lies between 2.0 and 4.0, around the predicted $2.7$ for broad starts.

Preventing collapse kept candidates among the samples. At $c_{\rm b}=0.6$, arm C kept pass@32 at $0.710$ against $0.561$ for the control. Its pass@1 at temperature 1 was $0.365$, above $0.354$ before training and below $0.481$ for the control, which concentrates the policy on its majority answer, and its solutions were longer. A process reward model then chose among 32 new samples per problem \citep{zhang2025prmlessons}, with best-of-32 accuracy $0.569$ with the bonus and $0.526$ without it, while the untrained model reached $0.603$. 

\subsection{Runs against the boundaries}\label{sec:reading}

Apart from the width $w=1$, where the broad state itself exceeds the collapse criterion, and testbed SGD runs whose steps became unstable at $c\ge10$, every broad start below a lower boundary collapsed in the testbed and the letter test, and no run stayed collapsed above an upper boundary (Table~\ref{tab:regimes}). The same held in 132 runs with new seeds (Appendix~\ref{app:seeds}). Release and collapse tracked boundaries that differ by a factor of $6.55$ between arms U and R and by $2.1$ between $G=8$ and $64$. Prevention needed a margin above the lower boundary, since inside the window, outcomes depended on the start, the seed, the learning rate and the optimizer.

\section{Which methods prevent collapse}\label{sec:meaning}

\begin{table}[t]
\centering\scriptsize
\caption{Common methods against collapse in the tabular flow, with the boundaries for the agreement reward and, for the bonus, the threshold of a broad policy under the majority reward on policies with one leading answer.}
\label{tab:remedies}
\begin{tabular}{@{}>{\raggedright\arraybackslash}p{0.22\linewidth}>{\raggedright\arraybackslash}p{0.73\linewidth}@{}}
\toprule
method & acts as exploration \\
\midrule
temperature & no, it rescales time (Proposition~\ref{prop:temperature}) \\
clipping, Adam & no, they move no equilibrium of exploration in the loss but can change outcomes in the window (Lemma~\ref{lem:fixedpoints}, Appendix~\ref{app:updates}) \\
normalization & no, it rescales exploration it does not divide (Proposition~\ref{prop:norm}) \\
KL penalty & partly, it pulls toward the reference policy and bypasses the division of GRPO (Section~\ref{sec:statedep}) \\
smoothing to $\bar p$ & yes, boundaries $\frac1n$ and $\frac n{4(n-1)}$ (Proposition~\ref{prop:potts-uniform}) \\
entropy bonus & yes, a Potts window under NPG, the committed branch stable only numerically (Theorem~\ref{thm:potts}) \\
bonus & yes, $c_{\rm b}^*$ before the division, $2.3$ to $3.2\,c_{\rm b}^*$ after it (Table~\ref{tab:bonus-main}) \\
leak & yes, a threshold in $\varepsilon/\eta$ (Theorem~\ref{thm:D5}) \\
\bottomrule
\end{tabular}
\end{table}

\paragraph{Placing exploration.}
Table~\ref{tab:remedies} sorts common methods by whether they act as exploration, and Appendix~\ref{app:guide} gives a recipe for setting exploration. Exploration stays on the path of the reward without a division, as in Dr.\ GRPO, or if the division reaches it too, as for a bonus inside the reward. Its equilibria and folds then do not move with the normalization, in the mean-field model, although sampled groups can still move them (Appendix~\ref{app:norm}). In a comparison that ignores the token level of the KL term and its nonuniform reference, the coefficient $0.001$ of TTRL lies far below the lower boundaries of smoothing for the majority reward (Appendix~\ref{app:ttrl}). For unordered answers, smoothing toward the uniform policy prevents collapse in the tabular flow only by pulling the policy to the uniform policy, which favors no answer. A bonus inside the reward just above $c_{\rm b}^*$ keeps a leading answer at an interior equilibrium, where free-form TTRL kept a mode share of $0.56$ to $0.67$ below the predicted $0.71$ to $0.88$. Larger coefficients lowered this share and pass@1 (Appendix~\ref{app:ttrl}).

\paragraph{What exploration cannot do.}
If the logit velocity of one softmax is locally Lipschitz and commutes with every permutation of the answers, the order of their probabilities is kept (Proposition~\ref{prop:order}). The rewards of Section~\ref{sec:setup} with smoothing toward $\bar p$, an entropy bonus or the bonus for rare answers satisfy this, with or without normalization. In the tabular flow, exploration therefore decides whether, when and how firmly a question commits, not which answer wins. Correcting a wrong majority needs a signal that breaks the symmetry, as when differential smoothing treats correct and incorrect answers differently \citep{gai2026differential} or a verifier rejects consistent but wrong answers \citep{pan2026coverrl}. Parameters shared across questions also break it, as collapsed questions of the letter test switched their leading letter, which the tabular flow leaves out. Methods that report gains in pass@1 over TTRL test where TTRL itself raises accuracy, unlike our subset, and several act outside our answer-symmetric terms \citep{evolrl2025,liu2025ettrl,wu2025spine,du2026dare,lin2026extinction,liao2026t3rl,wang2026serpo}, as Appendix~\ref{app:related-symmetry} discusses.

\section{Limitations and conclusion}\label{sec:conclusion}\label{sec:discussion}

\paragraph{Limitations.}
The theory treats one state at a time, but the questions of our language models shared parameters and committed together, a coupling that concurrent work models \citep{ding2026collective}. The thresholds of the majority reward and of the bonus hold on policies with one leading answer and equal others, the normalization is a mean-field model, and on ordered bins, only the folds are certified. We did not check full rank for our adapters or test a VLA model. Free-form TTRL used one model family, 128 problems selected with labels, groups of 8 for the vote and the update, 200 steps and no other methods against collapse as baselines. It departed from the original TTRL in its learning rate, normalization, handling of samples without an answer and evaluation temperature (Appendix~\ref{app:ttrl}). Its coefficient scan had three seeds and no collapsed starts.

\paragraph{Conclusion.}
In this work, we analyzed when exploration prevents the collapse of a policy trained on self-consistency rewards. In the tabular flow, the exploration coefficient has a lower boundary, below which only collapsed policies are stable, and an upper boundary, above which only policies spread over many answers are stable, and between the two, the initial policy decides the outcome. When exploration passes along the whole path of the reward, these boundaries are set by the training objective, so under exact gradients and in a mean-field model of the normalization, neither the learning rate nor the scaling of the optimizer moves them. When exploration bypasses a step, its threshold changes with the learning rate or with the policy, and the division of GRPO splits the threshold of a bonus for rare answers in two. Placed before the division, the bonus prevented collapse near its predicted threshold in a language model and in free-form TTRL and kept more correct answers among the samples, although it lowered pass@1 compared with TTRL without it. Exploration that treats all answers alike cannot correct a wrong majority, and in future work, we will combine it with signals that break the symmetry of the answers, such as verifiers.

\section*{Impact Statement}

This paper studies when exploration prevents the collapse of policies trained on self-consistency rewards and how its placement in the training algorithm changes the coefficient it needs. Its results can help practitioners keep more correct answers among the samples than training without exploration does, but they show neither how to correct a wrong majority nor how to repair a collapsed model. We do not see other specific societal consequences that need to be highlighted.

\bibliography{refs}
\bibliographystyle{icml2026}

\newpage
\appendix
\onecolumn
\setcounter{topnumber}{4}
\setcounter{bottomnumber}{2}
\setcounter{totalnumber}{6}
\renewcommand{\topfraction}{0.9}
\renewcommand{\bottomfraction}{0.6}
\renewcommand{\textfraction}{0.07}
\renewcommand{\floatpagefraction}{0.75}
\setlength{\textfloatsep}{7pt plus 2pt minus 2pt}
\setlength{\floatsep}{5pt plus 2pt minus 2pt}
\setlength{\intextsep}{5pt plus 2pt minus 2pt}

\section{Proofs of the main results}\label{app:proofs}

Every Lemma, Theorem, Proposition and Corollary of the main text is proved below, together with further results on ordered bins and networks. Auxiliary statements are proved where they are first needed. Throughout, $\mathbf1^\perp=\{\varphi\in\R^K\mid\sum_i\varphi_i=0\}$ is the tangent space of the simplex and $\bar p=\frac1K\mathbf1$ is the uniform policy, with $n$ in place of $K$ for unordered answers. Table~\ref{tab:status} gives the status of each claim.

\begin{table}[!htbp]
\centering\scriptsize
\renewcommand{\arraystretch}{1.15}
\caption{Status of the claims of Sections~\ref{sec:theory} to \ref{sec:meaning}. Proved means a theorem, proposition or corollary. Exact computation means exact sums over the counts of a group, without a proof of stability. Certified means a computer-assisted proof in interval arithmetic, and numerical means computed but not proved.}
\label{tab:status}
\begin{tabular}{@{}>{\raggedright\arraybackslash}p{0.36\linewidth}l>{\raggedright\arraybackslash}p{0.13\linewidth}>{\raggedright\arraybackslash}p{0.38\linewidth}@{}}
\toprule
claim & section & status & scope \\
\midrule
learning rate and preconditioner keep equilibria and folds & \S\ref{sec:potential} & proved & exact gradients, networks of full rank \\
dividing only selection rescales exploration by $\varsigma(p)$ & \S\ref{sec:potential}, \S\ref{sec:statedep} & proved & removal of the window numerical, mean-field model \\
window $\big(\frac1n,\frac n{4(n-1)}\big)$ for unordered answers & \S\ref{sec:unordered} & proved & both geometries, classical under NPG \\
collapse onto the initial plurality & \S\ref{sec:unordered} & proved & smoothing toward $\bar p$ \\
prevention costs least early & \S\ref{sec:unordered} & proved & one leading answer and equal others \\
boundaries of the majority reward & \S\ref{sec:unordered} & exact computation & one leading answer and equal others \\
folds on ordered bins & \S\ref{sec:twoscales} & certified & states symmetric about one bin, other states numerical \\
lower boundary at most $c^*$, upper at least $c_{\rm rel}$ & \S\ref{sec:regimes} & certified & equality with the folds numerical \\
upper boundary for networks & \S\ref{sec:regimes} & proved & smoothing toward $\bar p$, for $w=3$ on 256 bins above $0.3632$, numerical above $0.2919$ \\
linear threshold of a leak & \S\ref{sec:npg} & proved & line of $K$ bins \\
thresholds of the bonus & \S\ref{sec:statedep} & exact computation & one leading answer and equal others, near collapse the sign of a drift \\
order of the answers kept & \S\ref{sec:meaning} & proved & expected flow of one state \\
\bottomrule
\end{tabular}
\end{table}

\subsection{Policy flows and the lattice equations}\label{app:proofs-setup}

The flows \eqref{eq:lattice} and \eqref{eq:inloss} of Section~\ref{sec:setup} follow from the relation between logits and probabilities.

\begin{lemma}[Logits to probabilities]\label{lem:logit}
For any flow $z'$ of the logits, $\partial_t\log p_i=z_i'-\sum_jp_jz_j'$.
\end{lemma}

\begin{proof}
Differentiate $\log p_i=z_i-\log\sum_je^{z_j}$.
\end{proof}

With $z'=V$, Lemma~\ref{lem:logit} gives the selection part $p_i(V_i-\avg V_p)$ of \eqref{eq:lattice}. One step of the algorithm of Section~\ref{sec:exploration} changes $p$ by $\eta$ times this vector plus $\varepsilon\Delta p$ plus $O(\eta^2)$. Hence, with $\varepsilon=\omega\eta$ and $t=k\eta$, the iterates converge to the solution of \eqref{eq:lattice} on bounded time intervals as $\eta\to0$.

\begin{lemma}[Exploration through the geometry]\label{lem:explore-geometry}
Let $\tilde p$ be a probability vector held fixed during differentiation, and add $c\sum_j\tilde p_j\log p_j$ to the objective. Under NPG, this term contributes $\partial_tp=c(\tilde p-p)$ to the flow of the probabilities. Under vanilla policy gradient, it contributes $\partial_tp_i=c\,p_i\big((\tilde p_i-p_i)-\avg{\tilde p-p}_p\big)$, that is, $c(\tilde p-p)$ to the flow of the logits.
\end{lemma}

\begin{proof}
The Euclidean logit gradient of the term is $c(\tilde p-p)$, and Lemma~\ref{lem:logit} gives the second claim. For softmax policies, the natural gradient of an objective $\mathcal J(p)$ has logit velocity $\partial_{p_i}\mathcal J-\sum_jp_j\partial_{p_j}\mathcal J$ up to a constant shift, as for the reward in Section~\ref{sec:setup}. Here $\partial_{p_i}\mathcal J=c\tilde p_i/p_i$, so the velocity is $c(\tilde p_i/p_i-1)$ and its $p$-average vanishes. Lemma~\ref{lem:logit} then gives $\partial_tp_i=p_i\,c(\tilde p_i/p_i-1)=c(\tilde p_i-p_i)$.
\end{proof}

\subsection{Exploration in the loss}\label{app:inloss-proofs}\label{app:inloss-gradient}

The proofs of Theorem~\ref{thm:inloss-main} and Lemma~\ref{lem:fixedpoints} of Section~\ref{sec:potential} come first. Theorem~\ref{thm:inloss-cap} of Section~\ref{sec:twoscales} then locates the two folds of a Gaussian kernel, and bounds on the coefficients of peaked equilibria follow.

Throughout this subsection, $D$ is a symmetric stochastic matrix on the bins, so that $D_{ij}=D_{ji}\ge0$ and $\sum_jD_{ij}=1$. It is irreducible, that is, the graph with an edge between $i$ and $j$ whenever $D_{ij}>0$ is connected. We write $D_\star=\max_{i,j}D_{ij}$, $\mathcal P$ for the open simplex and $\bar{\mathcal P}$ for its closure. We fix $c>0$ and set
\begin{equation}
\mathcal V(p)=p\odot(p-S_2)+c\,(Dp-p),
\end{equation}
so that \eqref{eq:inloss} reads $z'=\mathcal V(p)$. We call this flow the vanilla flow of the objective. By Lemma~\ref{lem:explore-geometry}, NPG training of the same objective gives $\partial_tp=\mathcal V(p)$, since the reward part of NPG gives $p_i(p_i-S_2)$ and $\avg{p-S_2}_p=0$. This is \eqref{eq:lattice} with $q=0$ and with $c(D-\Id)$ in place of $\omega\Delta$, and we call it the NPG flow of the objective. The map $\mathcal V$ is defined on all of $\R^K$.

\begin{theorem}[A potential for exploration in the loss]\label{thm:inloss-gradient}
Let
\begin{equation}\label{eq:Fc}
F_c(p)=\frac13S_3-\frac14S_2^2-\frac c4\sum_{i,j}D_{ij}(p_i-p_j)^2 .
\end{equation}
Then $\nabla F_c=\mathcal V$ on $\R^K$, and $\sum_i\mathcal V_i(p)=S_2\big(1-\sum_ip_i\big)$.
\begin{enumerate}[label=(\alph*),leftmargin=*,itemsep=1pt]
\item The vanilla flow is
\begin{equation}
\partial_tp_i=p_i\big(\partial_iF_c-\avg{\nabla F_c}_p\big).
\end{equation}
This is the gradient flow of $F_c$ on $\mathcal P$ in the Shahshahani metric $\avg{\varphi,\tilde\varphi}_p=\sum_i\varphi_i\tilde\varphi_i/p_i$ \citep{shahshahani1979}, and, along solutions,
\begin{equation}
\frac d{dt}F_c=\sum_ip_i\big(\mathcal V_i-\avg{\mathcal V}_p\big)^2\ge0.
\end{equation}
\item The NPG flow is the Euclidean gradient flow $\partial_tp=\nabla F_c$. It preserves the plane $\sum_ip_i=1$, and $\frac d{dt}F_c=|\mathcal V|^2$.
\end{enumerate}
\end{theorem}

\begin{proof}
The derivatives of $\frac13S_3$ and $\frac14S_2^2$ are $p_i^2$ and $S_2p_i$. For the last term, the symmetry of $D$ gives
\begin{equation}
\partial_i\frac14\sum_{j,k}D_{jk}(p_j-p_k)^2=\sum_kD_{ik}(p_i-p_k)=p_i-(Dp)_i,
\end{equation}
since the rows of $D$ sum to one. Hence $\nabla F_c=\mathcal V$. The columns of $D$ also sum to one, so $\sum_i(Dp)_i=\sum_ip_i$ and $\sum_i\mathcal V_i=S_2-S_2\sum_ip_i$.

(a) By Lemma~\ref{lem:logit}, $\partial_tp_i=p_i(\mathcal V_i-\avg{\mathcal V}_p)=:f_i$. On the simplex, $\sum_if_i=0$, so $f\in\mathbf1^\perp$. For $\varphi\in\mathbf1^\perp$, one has $\avg{f,\varphi}_p=\sum_i(\mathcal V_i-\avg{\mathcal V}_p)\varphi_i=\nabla F_c\cdot\varphi$. Hence $f$ is the Shahshahani gradient of $F_c$, and $\frac d{dt}F_c=\sum_i\mathcal V_if_i=\sum_ip_i(\mathcal V_i-\avg{\mathcal V}_p)^2$, because $\sum_if_i=0$.

(b) The flow is $\partial_tp=\mathcal V=\nabla F_c$ by definition. On the plane $\sum_ip_i=1$, the entries of $\mathcal V$ sum to zero, so the plane is invariant. Moreover, $\frac d{dt}F_c=\nabla F_c\cdot\mathcal V=|\mathcal V|^2$.
\end{proof}

The last term of $F_c$ equals $-\frac c2\avg{p,(\Id-D)p}$, a Dirichlet energy of the policy on the graph with weights $D$. A smoothing term in the loss enters through the same geometry as the reward, Shahshahani under vanilla policy gradient and Euclidean under NPG. This is why the objective has a potential. Under vanilla policy gradient, a leak instead adds a Euclidean gradient to the Shahshahani gradient of selection, and the two do not combine. Under NPG, a leak with $\omega=\alpha c$ is the flow of part (b) for $D=\Id+\alpha\Delta$, so it has the potential $F_c$.

\begin{lemma}[Equilibria]\label{lem:inloss-equilibria}
(a) A point of $\mathcal P$ is an equilibrium of the vanilla flow if and only if $\mathcal V(p)=0$, and if and only if it is an equilibrium of the NPG flow.
(b) Every real solution $p\ne0$ of $\mathcal V(p)=0$ satisfies $\sum_ip_i=1$ and $p_i>0$ for all $i$.
\end{lemma}

\begin{proof}
(a) An interior point is an equilibrium of the vanilla flow exactly when $\mathcal V_i=C$ for all $i$ and some constant $C$. Since $\sum_i\mathcal V_i=0$ on the simplex, $C=0$.

(b) If $\mathcal V(p)=0$ and $p\ne0$, then $S_2>0$, and $\sum_i\mathcal V_i=S_2(1-\sum_ip_i)=0$ gives $\sum_ip_i=1$. The equation $\mathcal V_i=0$ can be written as
\begin{equation}
c\big((Dp)_i-p_i\big)=p_i(S_2-p_i).
\end{equation}
Let $p_{\min}=\min_ip_i$, and suppose that $p_{\min}\le0$. At every $i$ with $p_i=p_{\min}$, the right side equals $p_{\min}(S_2-p_{\min})\le0$. The left side is nonnegative, because $(Dp)_i$ is an average of entries that are at least $p_{\min}$. Both sides therefore vanish, so $p_j=p_{\min}$ whenever $D_{ij}>0$. By irreducibility, all entries equal $p_{\min}\le0$, which contradicts $\sum_ip_i=1$.
\end{proof}

The next two theorems prove the convergence and the stability statements of Theorem~\ref{thm:inloss-main}.

\begin{theorem}[Convergence]\label{thm:inloss-converge}
For every $c>0$, every solution of \eqref{eq:inloss} converges to a single equilibrium in $\mathcal P$. So does every solution of the NPG flow that starts in $\bar{\mathcal P}$.
\end{theorem}

\begin{proof}
The proof has five steps. Steps 1 to 4 treat the vanilla flow, and Step 5 treats the NPG flow.

\emph{Step 1 (the limit set consists of rest points).} The replicator vector field $f_i(p)=p_i(\mathcal V_i-\avg{\mathcal V}_p)$ is polynomial and leaves $\bar{\mathcal P}$ and each of its faces invariant. By the computation in Theorem~\ref{thm:inloss-gradient}(a), $F_c$ is nondecreasing along its solutions on all of $\bar{\mathcal P}$. Let $p(t)$ be a solution in $\mathcal P$, and let $\Omega$ be the set of its limit points as $t\to\infty$ for this extended flow \citep{hofbauer1998}. The set $\Omega$ is nonempty, compact, connected and invariant, and $F_c$ is constant on it. By the invariance principle of \citet{lasalle1960}, every $p^\infty\in\Omega$ is a rest point, that is, for all $i$,
\begin{equation}
p^\infty_i(\mathcal V_i(p^\infty)-\avg{\mathcal V}_{p^\infty})=0 .
\end{equation}

\emph{Step 2 (the limit set meets the interior).} Suppose that $\Omega$ lies in the boundary of $\bar{\mathcal P}$. Let $p^\infty\in\Omega$, and let $\mathcal J$ be its support, a proper subset of the bins. Then $\mathcal V_i(p^\infty)=\avg{\mathcal V}_{p^\infty}$ for $i\in\mathcal J$, and $\mathcal V_i(p^\infty)=c(Dp^\infty)_i\ge0$ for $i\notin\mathcal J$. Since $\sum_i\mathcal V_i(p^\infty)=0$,
\begin{equation}
|\mathcal J|\,\avg{\mathcal V}_{p^\infty}=-c\sum_{i\notin\mathcal J}(Dp^\infty)_i<0 .
\end{equation}
The sum is positive, because some $i\notin\mathcal J$ has $D_{ij}>0$ for some $j\in\mathcal J$. The function $p\mapsto\avg{\mathcal V}_p$ is continuous and negative on the compact set $\Omega$. Hence there is $\epsilon_1>0$ with $\avg{\mathcal V}_p\le-\epsilon_1$ on a neighborhood of $\Omega$, which contains $p(t)$ for all $t\ge t_0$. For $t\ge t_0$,
\begin{equation}
\frac d{dt}\sum_i\log p_i=\sum_i\big(\mathcal V_i-\avg{\mathcal V}_p\big)=-K\avg{\mathcal V}_p\ge K\epsilon_1 .
\end{equation}
This is impossible, because $\sum_i\log p_i\le-K\log K$ on the simplex. Hence $\Omega$ contains a point $p^\infty\in\mathcal P$, and $\mathcal V(p^\infty)=0$ by Lemma~\ref{lem:inloss-equilibria}.

\emph{Step 3 (the {\L}ojasiewicz bound).} Let $\mathcal M(p)=\mathrm{diag}(p)-pp^\top$, and let $\Pi$ be the orthogonal projection onto $\mathbf1^\perp$. For $\varphi\in\mathbf1^\perp$, one has $\varphi^\top\mathcal M(p)\varphi=\sum_ip_i\varphi_i^2-(\sum_ip_i\varphi_i)^2$, which is positive for $\varphi\ne0$ when $p\in\mathcal P$. Hence there are $\epsilon_0>0$ and $\lambda_\pm>0$ with $\lambda_-|\varphi|^2\le\varphi^\top\mathcal M(p)\varphi\le\lambda_+|\varphi|^2$ for $\varphi\in\mathbf1^\perp$ and $|p-p^\infty|\le\epsilon_0$. Since $\mathcal M\mathbf1=0$ and $\mathcal M$ maps $\R^K$ into $\mathbf1^\perp$, the flow is $\partial_tp=\mathcal M(p)\Pi\mathcal V(p)$, with $|\partial_tp|\le\lambda_+|\Pi\mathcal V|$. The vector $\Pi\mathcal V$ is the gradient of the restriction of $F_c$ to the plane $\sum_ip_i=1$. As $F_c$ is a polynomial, the inequality of \citet{lojasiewicz1963} gives $a\in(0,\frac12]$ and $C>0$ with
\begin{equation}\label{eq:A-lojasiewicz}
|F_c(p)-F_c(p^\infty)|^{1-a}\le C|\Pi\mathcal V(p)|
\end{equation}
on the plane for $|p-p^\infty|\le\epsilon_0$, after decreasing $\epsilon_0$.

\emph{Step 4 (finite length).} Along the solution, $F_c(p(t))$ increases to $F_c(p^\infty)$. While $|p(t)-p^\infty|\le\epsilon_0$ and $p(t)\ne p^\infty$, the bounds of Step 3 and \eqref{eq:A-lojasiewicz} give
\begin{equation}
-\frac d{dt}\big(F_c(p^\infty)-F_c(p)\big)^{a}=a\big(F_c(p^\infty)-F_c(p)\big)^{a-1}(\Pi\mathcal V)^\top\mathcal M\,\Pi\mathcal V\ge\frac{a\lambda_-}C\,|\Pi\mathcal V|\ge\frac{a\lambda_-}{C\lambda_+}\,|\partial_tp| .
\end{equation}
After a time $t_1$ at which the path lies in the ball, its length is therefore at most
\begin{equation}\label{eq:A-length}
\frac{C\lambda_+}{a\lambda_-}\big(F_c(p^\infty)-F_c(p(t_1))\big)^{a}
\end{equation}
for as long as it stays in the ball. Since $p^\infty\in\Omega$, there are times $t_1$ at which $|p(t_1)-p^\infty|$ and $F_c(p^\infty)-F_c(p(t_1))$ are arbitrarily small. After such a time, by \eqref{eq:A-length}, the path never leaves the ball, has finite length and converges. Its limit lies in $\Omega$, so $\Omega$ is a single point. The limit is therefore $p^\infty$.

\emph{Step 5 (the NPG flow).} The closed simplex is forward invariant, because $\partial_tp_i=c(Dp)_i\ge0$ where $p_i=0$, and $\sum_i\mathcal V_i=0$ on the plane. The flow is the Euclidean gradient flow of the polynomial $F_c$ restricted to the plane. By the {\L}ojasiewicz inequality, a bounded solution of such a flow converges to a single critical point. At a critical point, $\Pi\mathcal V=0$, hence $\mathcal V=0$ because $\sum_i\mathcal V_i=0$. The limit therefore lies in $\mathcal P$ by Lemma~\ref{lem:inloss-equilibria}.
\end{proof}

\begin{theorem}[Stability does not depend on the geometry]\label{thm:inloss-index}
Let $p$ be an equilibrium, and let
\begin{equation}
H=\nabla^2F_c(p)=\mathrm{diag}(2p-S_2-c)-2pp^\top+cD .
\end{equation}
Then $H\mathbf1=-S_2\mathbf1$ and $H\mathbf1^\perp\subset\mathbf1^\perp$. The linearizations at $p$ of the vanilla flow and of the NPG flow, restricted to $\mathbf1^\perp$, have real eigenvalues. Their numbers of positive, zero and negative eigenvalues equal those of $H$ on $\mathbf1^\perp$. In particular, $p$ is linearly stable for one flow if and only if it is linearly stable for the other, if and only if $H$ is negative definite on $\R^K$. The number of unstable directions is the number of positive eigenvalues of $H$.
\end{theorem}

\begin{proof}
Differentiating $\sum_i\mathcal V_i=S_2(1-\sum_ip_i)$ with respect to $p_j$ gives $\sum_iH_{ij}=2p_j(1-\sum_ip_i)-S_2=-S_2$ on the simplex. As $H$ is symmetric, $H\mathbf1=-S_2\mathbf1$ and $H\mathbf1^\perp\subset\mathbf1^\perp$. The NPG flow $\partial_tp=\mathcal V$ has the linearization $H$. The vanilla flow is $\partial_tp=\mathcal M(p)\mathcal V(p)$, with $\mathcal M$ as in the proof of Theorem~\ref{thm:inloss-converge}. Since $\mathcal V(p)=0$, its linearization is $\mathcal MH$, which maps $\mathbf1^\perp$ into $\mathbf1^\perp$. Let $\mathcal M_\perp$ and $H_\perp$ be the restrictions of $\mathcal M$ and $H$ to $\mathbf1^\perp$. The map $\mathcal M_\perp$ is symmetric and positive definite, so $\mathcal M_\perp H_\perp$ is similar to $\mathcal M_\perp^{1/2}H_\perp\mathcal M_\perp^{1/2}$, which is congruent to $H_\perp$. Sylvester's law of inertia gives the claim, and the remaining eigenvalue of $H$ on $\R^K$ is $-S_2<0$.
\end{proof}

\begin{proof}[Proof of Theorem~\ref{thm:inloss-main}]
The gradient structures are Theorem~\ref{thm:inloss-gradient}, and the convergence is Theorem~\ref{thm:inloss-converge}. The equilibria are described by Lemma~\ref{lem:inloss-equilibria}, and the stability statement is Theorem~\ref{thm:inloss-index}.
\end{proof}

For a general preconditioner, the equilibria are still the zeros of $\mathcal V$ (Lemma~\ref{lem:fixedpoints}), but Theorem~\ref{thm:inloss-index} covers their stability only for the two geometries above.

\begin{proof}[Proof of Lemma~\ref{lem:fixedpoints}]
Since $A(\theta)$ is invertible, $A(\theta)(\partial_\theta z)^\top\mathcal V=0$ if and only if $(\partial_\theta z)^\top\mathcal V=0$. Since $\partial_\theta z$ has full row rank, its transpose is injective, so $(\partial_\theta z)^\top\mathcal V=0$ if and only if $\mathcal V=0$. The argument uses nothing about $\mathcal V$, so it holds for every logit velocity.
\end{proof}

\paragraph{Computer-assisted folds.}
Theorem~\ref{thm:inloss-cap} is proved by certified interval computations, for which we first fix the kernels and the symmetric equilibria. On a ring of $K$ bins, the Gaussian kernel is $D_{ij}=g((j-i)\bmod K)$, with $g(d)=e^{-d^2/(2w^2)}/\sum_{d'}e^{-d'^2/(2w^2)}$ for the wrapped distances $-\frac K2\le d,d'<\frac K2$. On a line of an odd number $K$ of bins, numbered $0,\dots,K-1$, we truncate the profile to $|d|\le\lceil12w\rceil+5$ and normalize it to unit sum. We then reflect it at the two ends, $D_{ij}=\sum_{m=-2}^2\big(g(i-j+2mK)+g(i+j+1+2mK)\big)$. Both kernels are symmetric, stochastic and irreducible. On the ring, a policy is called site-symmetric if it is invariant under the reflection $j\mapsto-j$. On the line, we use the reflection about the central bin. In both cases, the symmetric vectors form a subspace that is invariant under $\mathcal V$ and under $H$. The \emph{peak} of a symmetric equilibrium is its entry at the center of the reflection.

\begin{table}[!htbp]
\centering\footnotesize
\caption{Certified folds of the symmetric equilibria of \eqref{eq:inloss} for the Gaussian kernel of width $w$ bins (Theorem~\ref{thm:inloss-cap-full}). The three columns from $c^*$ on describe the lower fold, and the three columns from $c_{\rm rel}$ on describe the upper fold. The digits shown are common to all listed geometries. Each column $c''$ is the second derivative of $c$ along the curve of equilibria, parametrized by the peak.}
\label{tab:inloss-cap}
\begin{adjustbox}{max width=\linewidth}
\begin{tabular}{llllllll}
\toprule
$w$ & geometries & $c^*$ & peak & $c''$ & $c_{\rm rel}$ & peak & $c''$ \\
\midrule
$1$ & rings $128$, $256$, line $129$ & $0.44720\,02914\,15856\,50285\ldots$ & $0.32684$ & $2.2801$ & $0.44935\,78532\,52217\,07478\ldots$ & $0.41223$ & $-1.4065$ \\
$\frac32$ & rings $128$, $256$, line $129$ & $0.29638\,99980\,46015\,17003\ldots$ & $0.20285$ & $12.004$ & $0.35252\,03123\,50551\,95385\ldots$ & $0.47045$ & $-2.3362$ \\
$3$ & rings $256$, $512$, line $257$ & $0.14774\,97416\,56434\,62888\ldots$ & $0.10096$ & $32.730$ & $0.29194\,51620\,98195\,36979\ldots$ & $0.49017$ & $-2.2097$ \\
$4$ & ring $256$ & $0.11077\,26365\,77790\,38623\ldots$ & $0.07583$ & $44.842$ & $0.28011\,11991\,63851\,72680\ldots$ & $0.49321$ & $-2.1579$ \\
$8$ & ring $512$ & $0.05536\,87032\,55579\,18355\ldots$ & $0.03805$ & $87.170$ & $0.26415\,12622\,44164\,08640\ldots$ & $0.49697$ & $-2.0785$ \\
\bottomrule
\end{tabular}
\end{adjustbox}
\end{table}

\begin{theorem}[Theorem~\ref{thm:inloss-cap} in full]\label{thm:inloss-cap-full}
Let $D=D_w$ on one of the geometries of Table~\ref{tab:inloss-cap}, untruncated on rings and truncated as above on lines.
\begin{enumerate}[label=(\alph*),leftmargin=*,itemsep=1pt]
\item There is a symmetric equilibrium $p^*$ at $c=c^*$ with the listed peak. At $p^*$, the matrix $H$ has a simple eigenvalue $0$ with a symmetric eigenvector $\psi$, normalized by $\psi=1$ at the peak, and $K-1$ negative eigenvalues. One of them is the eigenvalue $-S_2$ of the direction $\mathbf1$, which does not change the policy. Along the curve of equilibria $p^*+s\psi+O(s^2)$ through $p^*$, one has $c=c^*+\frac12c''s^2+O(s^3)$ with $c''>0$, and the eigenvalue of $H$ through zero has the sign of $s$. Hence, for $c$ slightly above $c^*$, there are exactly two equilibria near $p^*$. One is linearly stable and has the lower peak, and the other has exactly one unstable direction. For $c$ slightly below $c^*$, there is none.
\item The same holds at $c=c_{\rm rel}$, with $c''<0$ and with an eigenvalue through zero of the sign of $-s$. Hence, for $c$ slightly below $c_{\rm rel}$, there are exactly two equilibria near the fold. One is linearly stable and has the higher peak, and the other has exactly one unstable direction. For $c$ slightly above $c_{\rm rel}$, there is none.
\item For $w=\frac32$ on the ring of $128$ bins and $c=0.30$, $0.31$, $0.33$ and $0.35$, there are site-symmetric equilibria with $0$, $1$ and $0$ unstable directions. Their peaks are $0.1803$, $0.2304$ and $0.6784$, then $0.1613$, $0.2647$ and $0.6578$, then $0.1409$, $0.3267$ and $0.6073$, and finally $0.1273$, $0.4237$ and $0.5166$.
\end{enumerate}
The enclosures of $c^*$ and $c_{\rm rel}$ have radius below $10^{-39}$, and their values for different geometries with the same $w$ differ by less than $10^{-27}$.
\end{theorem}

\begin{proof}[Proof of Theorems~\ref{thm:inloss-cap} and \ref{thm:inloss-cap-full}]
Theorem~\ref{thm:inloss-cap} consists of parts (a) and (b) for every width of Table~\ref{tab:inloss-cap}, and Steps 1 to 3 below apply to every geometry listed there. The proof has four steps. Steps 1 to 3 enclose a fold, count the signs of the eigenvalues of $H$ there and derive its shape. Step 4 treats part (c).

\emph{Step 1 (enclosure of the fold).} All quantities are evaluated in ball arithmetic with $256$ bits \citep{johansson2017arb}, including the entries of $D$. On the symmetric subspace, we use the extended system
\begin{equation}
\mathcal V(p,c)=0,\qquad H(p,c)\psi=0,\qquad \tilde\psi^\top\psi=1
\end{equation}
of \citet{moore1980}, where $\tilde\psi$ is the unit vector of the peak bin. A numerical solution is refined by Newton's method to a residual below $10^{-76}$. The Krawczyk operator \citep{krawczyk1969,rump2010} of the box of radius $10^{-40}$ around it maps the box into its interior, with contraction ratio below $5\times10^{-12}$. Hence the extended system has exactly one solution in the box, and its Jacobian is invertible there. The solution is a probability vector with positive entries by Lemma~\ref{lem:inloss-equilibria}, even where the enclosure of a small entry contains zero.

\emph{Step 2 (inertia).} The matrix $H$ is block diagonal in a basis of symmetric and antisymmetric vectors. On the ring, this basis consists of $e_0$, $e_{K/2}$, $e_j+e_{K-j}$ and $e_j-e_{K-j}$ for $1\le j<K/2$. An interval Cholesky factorization certifies that the antisymmetric block is negative definite. It also certifies that the symmetric block is negative definite after the row and column of the peak are removed. By interlacing, the symmetric block has at most one nonnegative eigenvalue, and, at the fold, it has the eigenvalue $0$ of $\psi$.

\emph{Step 3 (a quadratic turning point).} Let
\begin{align}
a&=\psi^\top\partial_p^2\mathcal V[\psi,\psi],\qquad \partial_p^2\mathcal V[\psi,\psi]=2\psi\odot\psi-4(p\cdot\psi)\psi-2|\psi|^2p,\nonumber\\
b&=\psi^\top\partial_c\mathcal V,\qquad \partial_c\mathcal V=Dp-p.
\end{align}
The enclosures give $a>0$ and $b<0$ at every lower fold and $a<0$ and $b<0$ at every upper fold. Since $\ker H$ is spanned by $\psi$ and $H$ is symmetric, the range of $H$ is $\psi^\perp$. As $b\ne0$, the equilibria near $(p^*,c^*)$ form a smooth curve through it, and, as $a\ne0$, the fold is a quadratic turning point. Along the curve, $c(s)=c^*-\frac a{2b}s^2+O(s^3)$, so $c''=-a/b$. The eigenvalue of $H$ through zero has the derivative $a/|\psi|^2$ at $s=0$ \citep{crandall1973}. The other $K-1$ eigenvalues stay negative near the fold. With the signs of $a$ and $b$, this gives the statements on the two sides of the folds.

\emph{Step 4 (part (c)).} Part (c) uses the Krawczyk test for $\mathcal V(p,c)=0$ on the site-symmetric subspace at fixed rational $c$ and the same certificates. For the equilibria of index one, the symmetric block with one row and column removed is certified negative definite, and the Rayleigh quotient of a numerical eigenvector is certified positive. Each fold takes at most $11$ seconds.
\end{proof}

For $w=8$ on the ring of $512$ bins, numerical continuation from the broad state can reach the fold of another site-symmetric branch, at $c=0.0554091$, where $H$ has positive directions. We therefore start the continuation toward the lower fold for $w=8$ from a state rescaled from the end of the broad branch of $w=3$, and the inertia test of the certificate distinguishes the two branches. That the certified fold, and not the other one, ends the branch continued from the uniform policy is numerical.

\paragraph{A bound on peaked equilibria.}
The following bound limits the coefficients at which a single bin can carry a large probability.

\begin{proposition}[A bound on peaked equilibria]\label{prop:release}
Let $D$ be symmetric, stochastic and irreducible. At every equilibrium and every bin $i$ with $p_i=M>D_\star$, one has
\begin{equation}
c\le M^2(1-M)/(M-D_\star).
\end{equation}
Consequently, if $D_\star<\frac12$ and $c>\frac1{4(1-2D_\star)}$, no equilibrium has a bin with probability at least $\frac12$. For fixed $M_0\in(0,\frac12]$, no equilibrium has $\pmax\ge M_0$ once $c>\frac14+\frac12D_\star+O(D_\star^2)$.
\end{proposition}

\begin{proof}
Write $\mathcal V_i=0$ as $c\big(M-(Dp)_i\big)=M(M-S_2)$. Since $(Dp)_i\le D_\star\sum_jp_j=D_\star$ and $S_2\ge M^2$, we get $c(M-D_\star)\le M^2(1-M)$. For $f(M)=M^2(1-M)/(M-D_\star)$ and $M\ge\frac12$, one has $\frac d{dM}\log f=\frac2M-\frac1{1-M}-\frac1{M-D_\star}\le\frac1M-\frac1{1-M}\le0$. Hence the maximum of $f$ on $[\frac12,1)$ is $f(\frac12)=\frac1{4(1-2D_\star)}$. On $[M_0,1]$, $f(M)=M(1-M)+D_\star(1-M)+O(D_\star^2)$ uniformly, and the maximum of $M(1-M)+D_\star(1-M)$ is $\frac14+\frac12D_\star+\frac14D_\star^2$.
\end{proof}

For $w=3$ on a ring of 256 bins, $D_\star=0.13298$, and the bound decreases in $M$ on $[\frac38,1)$. It therefore excludes every equilibrium with $\pmax\ge\frac38$ once $c>0.3632$ and with $\pmax\ge\frac12$ once $c>0.3406$, against $c_{\rm rel}=0.2919$. For $w=8$, it excludes every equilibrium with a bin above $\frac12$ once $c>1/(4(1-2D_\star))=0.2777$, against $c_{\rm rel}=0.2642$. In numerical continuation for $w$ from $0.97$ to $32$, the spike branches keep their peaks above $0.39$ up to the fold, so for $w\ge\frac32$, where $D_\star<0.27$, the proposition also bounds their folds.

\subsection{Normalization}\label{app:proofs-norm}

Part (a) of Proposition~\ref{prop:norm} carries Theorem~\ref{thm:inloss-main} and Corollary~\ref{cor:upper} over to a common normalizer of both terms, and part (b) is the starting point of Section~\ref{sec:norm-reward}.

\begin{proof}[Proof of Proposition~\ref{prop:norm}]
(a) Since $\varsigma>0$, $\mathcal V/\varsigma$ vanishes exactly where $\mathcal V$ does, so the equilibria agree. At an equilibrium $p^*$, the derivative of $\mathcal V/\varsigma$ is $\partial_p\mathcal V(p^*)/\varsigma(p^*)$, because the term with the derivative of $1/\varsigma$ is multiplied by $\mathcal V(p^*)=0$. The same holds for the flows of the probabilities, $\partial_tp=\mathcal M(p)\mathcal V/\varsigma$ under vanilla policy gradient and $\partial_tp=\mathcal V/\varsigma$ under NPG. Their linearizations at $p^*$ are those of Theorem~\ref{thm:inloss-index} divided by $\varsigma(p^*)$, and a positive factor does not change the number of eigenvalues of each sign.

For the convergence, let $\tilde p$ solve the unnormalized flow from the same initial policy in the open simplex, and let $t_{\rm raw}(t)$ solve $\frac{d}{dt}t_{\rm raw}=1/\varsigma(\tilde p(t_{\rm raw}))$ with $t_{\rm raw}(0)=0$. By the chain rule and the uniqueness of solutions, $p(t)=\tilde p(t_{\rm raw}(t))$, as in Theorem~\ref{thm:std}. Since $\varsigma\le C$ for a constant $C$, $t_{\rm raw}(t)\ge t/C\to\infty$, so $p(t)$ has the limit of $\tilde p$, which exists by Theorem~\ref{thm:inloss-main}.

(b) Since $\varsigma(p)>0$, the equation $p\odot(p-S_2)/\varsigma(p)+c(Dp-p)=0$ is equivalent to $p\odot(p-S_2)+c\,\varsigma(p)(Dp-p)=0$, which is $\mathcal V(p)=0$ at the coefficient $c\,\varsigma(p)$. The entries of the normalized logit velocity sum to zero on the simplex, so, as in Lemma~\ref{lem:inloss-equilibria}(a), its equilibria in the open simplex are its zeros.
\end{proof}

Part (b) leaves stability open. With $\mathcal V_{\tilde c}$ the velocity of \eqref{eq:inloss} at the coefficient $\tilde c$, the normalized velocity is $\mathcal V_{c\varsigma(p)}(p)/\varsigma(p)$. At an equilibrium $p^*$, with $\varsigma=\varsigma(p^*)$, its derivative is
\begin{equation}
\frac1\varsigma\,\partial_p\mathcal V_{c\varsigma}(p^*)+\frac c\varsigma\,(Dp^*-p^*)\,\nabla\varsigma(p^*)^\top .
\end{equation}
The rank-one term can change the number of unstable directions away from the folds of \eqref{eq:inloss}, as in Appendix~\ref{app:norm-unfold}.

\subsection{Unordered answers}\label{app:proofs-unordered}

Propositions~\ref{prop:potts-uniform}, \ref{prop:potts-invariant}, \ref{prop:symmetric} and \ref{prop:order} and Corollaries~\ref{cor:plurality} and \ref{cor:upper} are proved below. Networks also inherit the unstable directions of the tabular flow (Proposition~\ref{prop:unordered-networks}) and, for the agreement reward, its convergence (Proposition~\ref{prop:net-converge}).

For $n=2$, both thresholds of Proposition~\ref{prop:potts-uniform} equal $\frac12$, and the window is empty.

Proposition~\ref{prop:potts-uniform} has the following counterpart on the policies with one leading answer.

\begin{proposition}[One leading answer]\label{prop:potts-invariant}
In the setting of Proposition~\ref{prop:potts-uniform}, in either geometry, the policies with one entry $M$ and $n-1$ equal entries form an invariant set. On it, a solution from $M(0)\in(\frac1n,\frac12)$ converges to $\bar p$ if $c>\frac n{n-1}M(0)(1-M(0))$ and to the stable equilibrium of Proposition~\ref{prop:potts-uniform}(b) if $c<\frac n{n-1}M(0)(1-M(0))$. From $M(0)\ge\frac12$, it converges to $\bar p$ if $c>\frac n{4(n-1)}$ and to that equilibrium if $c<\frac n{4(n-1)}$.
\end{proposition}

The proof of Proposition~\ref{prop:potts-uniform}(c) uses the following standard fact.

\begin{lemma}[Unstable equilibria attract a null set]\label{lem:nullset}
Let the flow of a continuously differentiable vector field on the open simplex have solutions for all $t\ge0$. Let its linearization at an equilibrium $p^*$, restricted to $\mathbf1^\perp$, have an eigenvalue with positive real part. Then the initial policies whose solutions converge to $p^*$ form a set of Lebesgue measure zero.
\end{lemma}

\begin{proof}
By the center-stable manifold theorem \citep{shub1987}, the points of a neighborhood $U$ of $p^*$ whose forward orbits stay in $U$ lie on a $C^1$ manifold of dimension less than that of the simplex. A solution converging to $p^*$ stays in $U$ after some integer time $m$. Its initial policy therefore lies in the preimage of this manifold under the flow map at time $m$. Each such map is a $C^1$ injection with invertible derivative, so these preimages are null sets, and so is their union over $m$.
\end{proof}

\begin{proof}[Proof of Propositions~\ref{prop:potts-uniform} and \ref{prop:potts-invariant}]
All parts rest on the quadratic and the Hessian form below, and the proof of part (b) has three steps. Here $D=\frac1n\mathbf1\mathbf1^\top$ is symmetric, stochastic and irreducible, so Theorem~\ref{thm:inloss-main} and the results of Appendix~\ref{app:inloss-proofs} apply in both geometries. Since $Dp=\bar p$ on the simplex, Lemma~\ref{lem:inloss-equilibria} shows that an equilibrium satisfies, for every $i$,
\begin{equation}
p_i^2-(S_2+c)p_i+\frac cn=0 .
\end{equation}
Every entry is therefore one of the two roots of this quadratic. We write $M$ for the larger root and $y$ for the smaller one, so that $M+y=S_2+c$ when they differ. For $\varphi\in\mathbf1^\perp$, $D\varphi=0$, so the Hessian of Theorem~\ref{thm:inloss-index} satisfies
\begin{equation}
\varphi^\top H\varphi=\sum_i(2p_i-S_2-c)\varphi_i^2-2(p\cdot\varphi)^2 .
\end{equation}

(a) At $\bar p$, $2p_i-S_2-c=\frac1n-c$ and $p\cdot\varphi=0$, so $H=(\frac1n-c)\Id$ on $\mathbf1^\perp$, while $H\mathbf1=-S_2\mathbf1$. Theorem~\ref{thm:inloss-index} gives the claim.

(b) \emph{Step 1 (several leading entries).} Let an equilibrium take the value $M$ at $m$ entries and $y<M$ at the others, with $1\le m\le n-1$. Then $2M-S_2-c=M-y$ and $2y-S_2-c=y-M$. If $m\ge2$, choose two entries $i\ne j$ with value $M$, and let $\varphi=e_i-e_j$. Then $p\cdot\varphi=0$ and $\varphi^\top H\varphi=2(M-y)>0$, so $H$ has a positive eigenvalue, and the equilibrium is unstable by Theorem~\ref{thm:inloss-index}.

\emph{Step 2 (the curve of one leading entry).} Let $m=1$, with $M$ at the entry $1$ and $y=(1-M)/(n-1)$ at the others. The condition $M+y=S_2+c$ with $S_2=M^2+(1-M)^2/(n-1)$ gives
\begin{equation}
c=M+\frac{1-M}{n-1}-M^2-\frac{(1-M)^2}{n-1}=M(1-M)+\frac{M(1-M)}{n-1}=\frac n{n-1}M(1-M),
\end{equation}
and $y<M$ means $M>\frac1n$. Conversely, for such $M$ and $c$, let $y=(1-M)/(n-1)$. Then $M+y=S_2+c$ by the same computation. The equations $p_i^2-(S_2+c)p_i+\frac cn=0$ at $M$ and at $y$ differ by $(M-y)(M+y-S_2-c)=0$, and the sum of all $n$ of them vanishes on the simplex. Hence both hold, and the state is an equilibrium. The function $c_{\rm br}(M)=\frac n{n-1}M(1-M)$ increases on $(\frac1n,\frac12)$ from $\frac1n$ to $\frac n{4(n-1)}$ and decreases on $(\frac12,1)$ to $0$, which gives the range of $c$.

\emph{Step 3 (stability on the curve).} For the stability at $m=1$, the vectors $\varphi\in\mathbf1^\perp$ with $\varphi_1=0$ form a subspace $\mathcal W$ of dimension $n-2$. On $\mathcal W$, $p\cdot\varphi=y\sum_i\varphi_i=0$ and $(H\varphi)_i=(y-M)\varphi_i$, so $\mathcal W$ is an eigenspace of $H$ with the negative eigenvalue $y-M$. The orthogonal complement of $\mathcal W$ in $\mathbf1^\perp$ is spanned by $\psi=(n-1,-1,\dots,-1)$, which is therefore also an eigenvector, because $H$ is symmetric and preserves $\mathbf1^\perp$. With $p\cdot\psi=(n-1)(M-y)$,
\begin{align}
\psi^\top H\psi&=(n-1)^2(M-y)+(n-1)(y-M)-2(n-1)^2(M-y)^2\nonumber\\
&=(n-1)(M-y)\big(n-2-2(n-1)(M-y)\big).
\end{align}
Since $(n-1)(M-y)=nM-1$, the last factor equals $n(1-2M)$. The eigenvalue of $\psi$ is therefore negative for $M>\frac12$ and positive for $M<\frac12$, and Theorem~\ref{thm:inloss-index} gives the stability statements.

(c) Every nonuniform equilibrium has $1\le m\le n-1$ entries at the larger root. By (b), those with $m=1$ exist only for $c\le\frac n{4(n-1)}$. For general $m$, write $M=y+d$ with $d>0$. The mass condition gives $y=(1-md)/n$, so $S_2=\frac1n+md^2(1-\frac mn)$, and $M+y=S_2+c$ gives
\begin{equation}
nc=1+(n-2m)d-m(n-m)d^2\le1+\frac{(n-2m)^2}{4m(n-m)}=\frac{n^2}{4m(n-m)}\le\frac{n^2}{4(n-1)},
\end{equation}
since $m(n-m)\ge n-1$ for $1\le m\le n-1$. Hence, for $c>\frac n{4(n-1)}$, $\bar p$ is the only equilibrium, and every solution converges to it by Theorem~\ref{thm:inloss-main}. If $c<\frac1n$, $\bar p$ is unstable by (a), and equilibria with $m\ge2$ are unstable by (b). Those with $m=1$ have $c_{\rm br}(M)=c<\frac1n$, hence $M>\frac12$, and are stable. Every solution converges to an equilibrium by Theorem~\ref{thm:inloss-main}. The equilibria are finitely many, since each choice of the $m$ entries at the larger root allows at most two values of $d$. Every equilibrium other than the $n$ stable ones has a positive eigenvalue of $H$ on $\mathbf1^\perp$, by (a) and Step 1, and hence of the linearization of either flow (Theorem~\ref{thm:inloss-index}). Both flows are polynomial and keep the open simplex forward invariant, so Lemma~\ref{lem:nullset} applies, and the initial policies whose solutions converge to one of them form a null set. Hence almost every solution converges to one of the $n$ stable equilibria.

Proposition~\ref{prop:potts-invariant}. The flow commutes with permutations of the answers, so the set of policies whose entries other than the first are equal is invariant. On it, write $p_1=M$ and $p_j=y=(1-M)/(n-1)$ for $j\ge2$. By Lemma~\ref{lem:logit} and the computation in (b),
\begin{equation}
\partial_t\log(M/y)=\mathcal V_1-\mathcal V_j=(M-y)(M+y-S_2-c)=(M-y)\big(c_{\rm br}(M)-c\big).
\end{equation}
Under NPG, Lemma~\ref{lem:logit} gives instead $\partial_t\log(M/y)=(M-y)\big(1-c/(nMy)\big)$, and $nMy=c_{\rm br}(M)$, so both geometries give the same signs. Since $M/y$ increases with $M$, the scalar function $M(t)$ increases where $c_{\rm br}(M)>c$ and decreases where $c_{\rm br}(M)<c$, for $M>\frac1n$. A bounded monotone solution of this scalar equation converges to the first zero of $c_{\rm br}(M)-c$, or to $M=\frac1n$, in its direction of motion.

If $M(0)<\frac12$ and $c>c_{\rm br}(M(0))$, then $c_{\rm br}<c$ on $(\frac1n,M(0)]$, so $M$ decreases to $\frac1n$, which is $\bar p$. If $M(0)<\frac12$ and $c<c_{\rm br}(M(0))$, then $M$ increases, and $c_{\rm br}>c$ until the zero of $c_{\rm br}-c$ beyond $\frac12$, which is the stable equilibrium of (b). If $M(0)\ge\frac12$, the same argument gives convergence to that equilibrium for $c<\frac n{4(n-1)}=\max c_{\rm br}$ and to $\bar p$ for $c>\frac n{4(n-1)}$.
\end{proof}

Section~\ref{sec:unordered} uses the following extension of Proposition~\ref{prop:potts-uniform} to rewards that depend only on the counts of a group.

\begin{proposition}[Symmetric rewards]\label{prop:symmetric}
Let the reward of a sample depend only on its answer and on the counts of the answers in its group, symmetrically under permutations of the answers. Let the advantages be the rewards minus their group mean, possibly divided by a positive function of the rewards of the group. For $p=(M,y,\dots,y)$ with $y=\frac{1-M}{n-1}$, let $\mathcal A(M)$ be the expected advantage of a sample of answer $1$. Under smoothing toward $\bar p$ with $c>0$, these policies form an invariant set of vanilla policy gradient and of NPG. On it, for $\frac1n<M<1$, the mass $M$ increases if $c<c_{\rm eq}(M)$ and decreases if $c>c_{\rm eq}(M)$, where $c_{\rm eq}(M)=M\mathcal A(M)/(M-\frac1n)$.
\end{proposition}

\begin{proof}[Proof of Proposition~\ref{prop:symmetric}]
The proof writes both flows with the expected advantages and then restricts them to policies with one leading answer. Write $\mathcal A_i(p)$ for the expected advantage of a sample of answer $i$. In every group, the advantages sum to zero, also after the division by a positive factor of the group. Under vanilla policy gradient, the term $-p$ of the gradient of $\log p_{a_j}$ therefore cancels, and the expected update of the logits by selection is $\frac1G\E\sum_j\hat{\mathcal A}_je_{a_j}$, where $\hat{\mathcal A}_j$ is the advantage of sample $j$. Since the samples are exchangeable, its entry $i$ is $p_i\mathcal A_i$. With the smoothing term, the logit velocity under vanilla policy gradient and the flow of the probabilities under NPG are
\begin{equation}
z_i'=p_i\mathcal A_i+c(\frac1n-p_i),\qquad \partial_tp_i=p_i(\mathcal A_i-\avg{\mathcal A}_p)+c(\frac1n-p_i),
\end{equation}
where $\avg{\mathcal A}_p=0$ by the same cancellation.

Let $\pi$ be a permutation of the answers. Since the reward depends only on the counts and treats the answers symmetrically, $\mathcal A_{\pi(i)}(\pi p)=\mathcal A_i(p)$. Both flows therefore commute with permutations, and the policies whose entries other than the first are equal form an invariant set. On it, write $p_1=M$, $p_j=y=\frac{1-M}{n-1}$ and $\mathcal A_1=\mathcal A(M)$. Then $\sum_ip_i\mathcal A_i=0$ gives $y\mathcal A_j=-M\mathcal A(M)/(n-1)$ for $j\ge2$, and $M-y=\frac{nM-1}{n-1}$. Under vanilla policy gradient,
\begin{equation}
\partial_t\log\frac My=z_1'-z_j'=\frac n{n-1}M\mathcal A(M)-c\,(M-y)=\frac n{n-1}\Big(M\mathcal A(M)-c\,\big(M-\tfrac1n\big)\Big).
\end{equation}
Under NPG, $\mathcal A_1-\mathcal A_j=\mathcal A(M)/(1-M)$ and $\frac{1/n-M}M-\frac{1/n-y}y=\frac1n\big(\frac1M-\frac1y\big)=-\frac{M-1/n}{M(1-M)}$, so
\begin{equation}
\partial_t\log\frac My=\frac1{M(1-M)}\Big(M\mathcal A(M)-c\,\big(M-\tfrac1n\big)\Big).
\end{equation}
Since $M/y$ increases with $M$, the mass $M$ increases in both geometries where $M\mathcal A(M)>c(M-\frac1n)$, and it decreases where the inequality is reversed. For $M>\frac1n$, these conditions are $c<c_{\rm eq}(M)$ and $c>c_{\rm eq}(M)$, respectively. The same argument applies if the smoothing term of each group is multiplied by a nonnegative factor of the group with positive mean. The expected smoothing term is then multiplied by this mean, which divides $c_{\rm eq}$ by it.
\end{proof}

Section~\ref{sec:regimes} uses the following consequence for networks.

\begin{corollary}[No collapsed stationary point above the upper boundary]\label{cor:upper}
Let a network with one policy per state satisfy the assumptions of Lemma~\ref{lem:fixedpoints}, with the agreement reward and $D=\frac1n\mathbf1\mathbf1^\top$ for $n\ge3$. If $c>\frac n{4(n-1)}$, every stationary point gives every state the uniform policy, also with both terms divided by one normalizer.
\end{corollary}

\begin{proof}[Proof of Corollary~\ref{cor:upper}]
Write the logits of all states as one vector $z(\theta)$, and let $\mathcal V$ act on the logits of each state. By Lemma~\ref{lem:fixedpoints}, $\theta$ is stationary if and only if $\mathcal V$ vanishes at the logits of every state. The policy of every state is then an equilibrium of \eqref{eq:inloss}. By Proposition~\ref{prop:potts-uniform}(c), every solution converges to $\bar p$ for $c>\frac n{4(n-1)}$. An equilibrium is a constant solution, so $\bar p$ is the only equilibrium, and every state has the uniform policy. If both terms are divided by a positive normalizer, the velocity keeps its zeros, so the claim still holds (Proposition~\ref{prop:norm}(a)).
\end{proof}

On ordered bins, Proposition~\ref{prop:release} gives the same conclusion above an explicit bound. For $w=3$ on the ring of 256 bins, every stationary point of such a network gives every state $\pmax<\frac38$ once $c>0.3632$, against the fold $c_{\rm rel}=0.2919$ of the tabular flow.

Section~\ref{sec:meaning} uses the following symmetry.

\begin{proposition}[Exploration that treats answers alike keeps their order]\label{prop:order}
Let the logit velocity of one softmax over answers be locally Lipschitz and commute with every permutation of the answers. If $p_i(0)>p_j(0)$, then $p_i(t)>p_j(t)$ on the interval of existence, and if $p_i(0)=p_j(0)$, then $p_i(t)=p_j(t)$ there.
\end{proposition}

The rewards of Section~\ref{sec:setup} with smoothing toward $\bar p$, an entropy bonus or the bonus for rare answers satisfy these assumptions, with or without normalization. The statement concerns the deterministic flow of one state, and sampling noise or shared parameters can reorder the answers.

\begin{proof}[Proof of Proposition~\ref{prop:order}]
Write the flow as $z'=F(z)$. The rewards and exploration terms listed before the proof give a locally Lipschitz $F$, since the expected advantages of groups of finite size are polynomials in $p$ and the exploration terms are smooth in $z$. Let $\tau$ swap the answers $i$ and $j$. Since $F$ commutes with $\tau$, the curve $\tau z(t)$ solves the same equation as $z(t)$. If $p_i(t_0)=p_j(t_0)$ at some time $t_0$, then $z_i(t_0)=z_j(t_0)$, so $\tau z(t_0)=z(t_0)$, and uniqueness gives $\tau z(t)=z(t)$, that is, $p_i(t)=p_j(t)$, on the whole interval of existence. Hence $p_i-p_j$ either vanishes identically there or never vanishes, and by continuity, it keeps the sign it has at $t=0$.
\end{proof}

\begin{proof}[Proof of Corollary~\ref{cor:plurality}]
By Proposition~\ref{prop:potts-uniform}(c), almost every solution converges to one of the $n$ stable equilibria, each with one entry $M>\frac12$ and $n-1$ entries below $\frac12$. Initial policies with two equal largest entries form a null set. Let $p(0)$ have a single largest entry $p_i(0)$. Both flows keep the open simplex invariant, so the logits exist for all $t\ge0$, and under NPG, we take the logit velocity $\mathcal V_i(p)/p_i$. In either geometry, the logit velocity is smooth on the open simplex and commutes with every permutation of the answers, so Proposition~\ref{prop:order} gives $p_i(t)>p_j(t)$ for all $j\ne i$ and $t\ge0$. The limit therefore has $p_i\ge p_j$ for every $j$. Its largest entry is unique, so it is the entry $i$.
\end{proof}

\begin{proposition}[Stability of networks for unordered answers]\label{prop:unordered-networks}
Let $n\ge3$, and let the reward and the advantages satisfy the assumptions of Proposition~\ref{prop:symmetric}. Let $\mathcal A_i(p)$ be the expected advantage of a sample of answer $i$, and write
\begin{equation}\label{eq:A-sym-velocity}
\mathcal V_i(z)=p_i\mathcal A_i(p)+c\Big(\frac1n-p_i\Big)
\end{equation}
for the logit velocity of vanilla policy gradient under smoothing toward $\bar p$. For the leave-one-out agreement reward, it is \eqref{eq:inloss} with selection multiplied by $\frac{G-2}G$. Let $p^*$ be an equilibrium whose entries take at most two values, such as the uniform policy or a policy $(M,y,\dots,y)$, and let $z^*$ be its logits.
\begin{enumerate}[label=(\alph*),leftmargin=*,itemsep=1pt]
\item The Jacobian $\mathcal J=\partial_z\mathcal V(z^*)$ is symmetric.
\item Let a network with one policy per state satisfy the assumptions of Lemma~\ref{lem:fixedpoints}. Let $\theta^*$ be a stationary point at which the policy of every state $s$ is such an equilibrium, with Jacobian $\mathcal J_s$. Then, for every symmetric positive definite $A(\theta^*)$, the linearization of the network at $\theta^*$ has real eigenvalues. It has $\sum_sn_+(\mathcal J_s)$ positive and $\sum_sn_-(\mathcal J_s)$ negative eigenvalues, where $n_+$ and $n_-$ count the positive and negative eigenvalues of a symmetric matrix. In particular, it has no positive eigenvalue if and only if no $\mathcal J_s$ has one.
\item For the agreement reward without normalization, every equilibrium has entries of at most two values. A full-rank network therefore has a positive eigenvalue at a stationary point if and only if the policy of some state is a linearly unstable equilibrium of the tabular flow.
\end{enumerate}
\end{proposition}

\begin{proof}
Since the reward treats the answers alike, $\mathcal V(\pi z)=\pi\mathcal V(z)$ for every permutation $\pi$ of the answers. Let $\Pi$ be the group of permutations that fix $p^*$. A permutation in $\Pi$ maps $z^*$ to $z^*$ up to a shift of all logits, which does not change the policy. Differentiating at $z^*$ therefore shows that $\mathcal J$ commutes with every $\pi\in\Pi$. Moreover, $\mathcal J\mathbf1=0$, and $\mathcal J$ maps into $\mathbf1^\perp$ because the entries of $\mathcal V$ sum to zero.

(a) If all entries of $p^*$ are equal, $\Pi$ is the full symmetric group, which acts irreducibly on $\mathbf1^\perp$. By Schur's lemma, $\mathcal J$ is a multiple of the orthogonal projection onto $\mathbf1^\perp$. Otherwise, let $\mathcal I_1$ be the set of the $m$ answers with the larger value and $\mathcal I_2$ the set of the other $n-m$, with $1\le m\le n-1$. Then $\Pi$ consists of the permutations that preserve $\mathcal I_1$ and $\mathcal I_2$, and
\begin{equation}\label{eq:A-schur-split}
\mathbf1^\perp=\mathrm{span}(\psi)\oplus W_1\oplus W_2,\qquad \psi=(n-m)\,\mathbf1_{\mathcal I_1}-m\,\mathbf1_{\mathcal I_2}.
\end{equation}
Here $\mathbf1_{\mathcal I_k}$ is the indicator vector of $\mathcal I_k$, and $W_k$ is the space of vectors supported on $\mathcal I_k$ whose entries sum to zero. The three summands are invariant under $\Pi$ and pairwise orthogonal. The nonzero ones are absolutely irreducible and mutually inequivalent. Indeed, $\Pi$ acts trivially on $\psi$ and acts on each $W_k$ by the standard representation of the symmetric group of $\mathcal I_k$, which stays irreducible over the complex numbers. Schur's lemma therefore gives
\begin{equation}\label{eq:A-schur-J}
\mathcal J=a\,P_\psi+b_1P_{W_1}+b_2P_{W_2}
\end{equation}
with the orthogonal projections $P_\psi$, $P_{W_1}$ and $P_{W_2}$ onto the three summands and reals $a$, $b_1$ and $b_2$. Hence $\mathcal J$ is symmetric.

(b) Write $J = \partial_\theta z(\theta^*)$ for the logits of all states and $\mathcal{J} = \mathrm{diag}(\mathcal{J}_s)$. Since $\mathcal V$ vanishes at every state, the linearization of $\theta' = A(\theta)\,(\partial_\theta z)^\top \mathcal V(z(\theta))$ at $\theta^*$ is $L = A J^\top \mathcal{J} J$. It is similar to $A^{1/2} J^\top \mathcal{J} J A^{1/2}$, which is symmetric by (a) and congruent to $J^\top \mathcal{J} J$. Since $J$ has full row rank, Sylvester's law of inertia gives $n_\pm(J^\top \mathcal{J} J) = n_\pm(\mathcal{J}) = \sum_s n_\pm(\mathcal{J}_s)$.

(c) By the proof of Proposition~\ref{prop:potts-uniform}, every entry of an equilibrium is one of the two roots of the quadratic $p_i^2-(S_2+c)p_i+\frac cn=0$, with $c$ multiplied by $\frac G{G-2}$ for the leave-one-out reward. The matrix $\mathcal J_s$ has the nonzero eigenvalues of the linearization of the tabular flow at the policy of state $s$, so part (b) gives the claim.
\end{proof}

\begin{proposition}[Networks converge to equilibria of the tabular flow]\label{prop:net-converge}
Let $n\ge3$, let the reward be the agreement reward without normalization, and let $D=\frac1n\mathbf1\mathbf1^\top$. Let a network with one policy per state and twice continuously differentiable logits $z(\theta)$ follow
\begin{equation}\label{eq:A-net-flow}
\theta'=A(\theta)\,(\partial_\theta z)^\top\mathcal V\big(z(\theta)\big),
\end{equation}
with $A(\theta)$ symmetric positive definite and locally Lipschitz. If $\theta(t)$ stays bounded and $\partial_\theta z$ has full row rank at every limit point of $\theta(t)$, then the policy of every state converges to an equilibrium of \eqref{eq:inloss}. If $c>\frac n{4(n-1)}$, it is $\bar p$ in every state.
\end{proposition}

\begin{proof}
For this kernel, $Dp=\bar p$, and the logit velocity of each state is a gradient,
\begin{equation}\label{eq:A-net-potential}
\mathcal V=\nabla_z\Phi,\qquad \Phi(z)=\frac12S_2+\frac cn\sum_{i=1}^n\log p_i,
\end{equation}
because $\partial_{z_i}S_2=2p_i(p_i-S_2)$ and $\sum_j\partial_{z_i}\log p_j=1-np_i$. Let $\Phi_{\rm net}(\theta)$ be the sum of $\Phi(z_s(\theta))$ over the states $s$, and write $J=\partial_\theta z$ for the Jacobian of the logits of all states. Along \eqref{eq:A-net-flow},
\begin{equation}\label{eq:A-net-lyapunov}
\frac{d}{dt}\Phi_{\rm net}\big(\theta(t)\big)=\mathcal V^\top J\,A(\theta)\,J^\top\mathcal V\ \ge\ 0,
\end{equation}
with equality only where $J^\top\mathcal V=0$. On the bounded trajectory, the logits are bounded, so every $p_i$ stays away from zero and $\Phi_{\rm net}$ is bounded. The set $\Omega$ of limit points of $\theta(t)$ is nonempty, compact, connected and invariant, and $\Phi_{\rm net}$ is constant on it. By the invariance principle of \citet{lasalle1960}, $J^\top\mathcal V=0$ on $\Omega$, and the full row rank of $J$ gives $\mathcal V=0$ there. At every point of $\Omega$, the policy of every state is therefore an equilibrium of \eqref{eq:inloss}. For this kernel, these equilibria are finitely many. Let a nonuniform equilibrium have its larger value at $m$ given entries. By the proof of Proposition~\ref{prop:potts-uniform}(c), the gap $d$ between its two values is a root of the nondegenerate quadratic $nc=1+(n-2m)d-m(n-m)d^2$, so each choice of these entries gives at most two equilibria. The policy of a state depends continuously on $\theta$, so its limit points form a connected subset of this finite set, that is, one equilibrium. If $c>\frac n{4(n-1)}$, Proposition~\ref{prop:potts-uniform}(c) leaves only $\bar p$.
\end{proof}

\begin{remark}
The proposition covers SGD and any preconditioner that depends only on the parameters, such as a stationary scaling of Adam. It does not cover the dependence of Adam on past gradients, sampled rewards, or the epochs and clipping of PPO. For the leave-one-out reward, $\frac12S_2$ is replaced by $\frac{G-2}{2G}S_2$, which multiplies the threshold by $\frac{G-2}G$. With this replacement, the sum $-\Phi_{\rm net}$ is the landscape of Appendix~\ref{app:llm}. Full row rank needs at least as many parameters as logits, so it cannot hold for free-form answers, whose number is unbounded, and we claim no transfer to networks there. We did not check it for the low-rank adapters of Appendix~\ref{app:llm}.
\end{remark}

\begin{observation}[Ordered bins]
\label{obs:ordered-precond}
With $H$ as in Theorem~\ref{thm:inloss-index} and $\mathcal M$ as in its proof, $[H,\mathcal M]=c\,[D,\mathcal M]$, which does not vanish for the Gaussian kernel, so the logit Jacobian $H\mathcal M$ of \eqref{eq:inloss} is not symmetric. Numerically, for $w = 3$ on the ring of $256$ bins and $c$ from $0.152$ to $0.285$, the symmetric part of $H\mathcal M$ has a positive eigenvalue between $3 \times 10^{-5}$ and $1.4 \times 10^{-4}$ at the stable broad equilibrium. With its eigenvector $f$, the preconditioner $ff^\top+10^{-3}\,\Id$ makes this equilibrium linearly unstable. At the stable spikes we computed, the symmetric part has no positive eigenvalue above round-off. By Lemma~\ref{lem:fixedpoints}, a network shares the folds of Theorem~\ref{thm:inloss-cap} for every preconditioner, while the stability of a broad peak inside the window is a property of the tabular flow.
\end{observation}

\subsection{The linear threshold of a leak}\label{app:npg-dichotomy}

The linear threshold of a leak in Section~\ref{sec:npg} rests on the spectrum of the graph Laplacian. On a line of $K$ bins with reflecting ends, numbered $1,\dots,K$, the matrix $-\Delta$ has the eigenvalues $\lambda_k^N=4\sin^2\frac{\pi k}{2K}$ and the eigenvectors $\varphi_k(i)=\cos\frac{\pi k(i-1/2)}K$ for $k=0,\dots,K-1$.

\begin{theorem}[Linear threshold of the uniform policy]\label{thm:D5}
On a line of $K$ bins with reflecting ends, the linearization of \eqref{eq:lattice} at $\bar p$ on $\mathbf1^\perp$ has the eigenvalues $K^{-(q+1)}-\omega\lambda_k^N$ for $k=1,\dots,K-1$. The uniform policy is unstable for $\omega<\omega_{\rm lin}(K)$ and linearly stable for $\omega>\omega_{\rm lin}(K)$, where
\begin{equation}\label{eq:omegalin}
\omega_{\rm lin}(K)=\frac{K^{-(q+1)}}{4\sin^2\frac\pi{2K}} .
\end{equation}
In units of the action range, $\sigma=\omega K^{q-1}$, the threshold is $\sigma=1/(K^2\lambda_1^N)$, which tends to $1/\pi^2$ as $K\to\infty$ for both geometries.
\end{theorem}

\begin{proof}[Proof of Theorem~\ref{thm:D5}]
For a zero-mean perturbation $f$ of $\bar p$, the derivatives of $S_2$ and $V_i$ at $\bar p$ in the direction $f$ are $\frac2K\sum_if_i=0$ and $K^{-q}f_i$. Since $V_j=0$ at $\bar p$, the derivative of $\avg{V}_p$ in this direction vanishes. The linearized reaction is $K^{-(q+1)}f_i$, and the diffusion has the Neumann spectrum $-\omega\lambda_k^N$.
\end{proof}

Under NPG, the smoothing term in the loss with $D=\Id+\alpha\Delta$ and $0<\alpha\le\frac12$ gives the flow $\partial_tp=p\odot(p-S_2)+\alpha c\,\Delta p$ (Lemma~\ref{lem:explore-geometry}), which is \eqref{eq:lattice} with $q=0$ and $\omega=\alpha c$. Its uniform policy is therefore linearly stable exactly for $c>\omega_{\rm lin}(K)/\alpha$ with $q=0$, and by Theorem~\ref{thm:inloss-index}, so is that of vanilla policy gradient on the same objective. For $K=256$ and $\alpha=\frac12$, this threshold is $c=51.9$.

\section{Further results on exploration in the loss}\label{app:inloss-more}

The potential, the convergence and the certified folds of the lattice equation \eqref{eq:inloss} are proved in Appendix~\ref{app:inloss-proofs}, and the results below complement them.

\subsection{The uniform policy, lines and states between bins}\label{app:inloss-thresholds}

Section~\ref{sec:twoscales} uses three further results, the threshold above which the uniform policy is stable, the folds on lines with reflecting ends and the stability of states centered between two bins.

From now on, $D=D_w$ is the Gaussian kernel of the testbed. On a ring of $K$ bins, $D_{ij}=g((j-i)\bmod K)$ with $g(d)=e^{-d^2/(2w^2)}/\sum_{d'}e^{-d'^2/(2w^2)}$ for the wrapped distances $-\frac K2\le d,d'<\frac K2$. Its entries are positive, so the results of Appendix~\ref{app:inloss-proofs} apply. A policy symmetric under the reflection $j\mapsto-j$ is called site-symmetric. The site-symmetric vectors form a subspace that is invariant under $\mathcal V$ and under $H$.

\begin{proposition}[Uniform threshold of exploration in the loss]\label{prop:inloss-uniform}
On the ring of $K$ bins, let $D$ be the periodization of a symmetric probability kernel $g$ on $\Z$ with variance $w_g^2>0$ and finite fourth moment $\avg{d^4}_g=\sum_dd^4g(d)$. Let $\hat g_k$ be the eigenvalues of $D$, and suppose that $\hat g_1\ge\hat g_k$ for $k=1,\dots,K-1$. The uniform policy is an equilibrium of \eqref{eq:inloss}. It is linearly unstable for $c<c_{\rm lin}(K)$ and linearly stable for $c>c_{\rm lin}(K)$, where
\begin{equation}\label{eq:clin}
c_{\rm lin}(K)=\frac{K^{-1}}{1-\hat g_1}=\frac{K}{2\pi^2w_g^2}\big(1+O(K^{-2})\big).
\end{equation}
For the Gaussian kernel, $\hat g_1\ge\hat g_k$ holds, and $w_g^2$ differs from $w^2$ by a relative error below $10^{-6}$ for $w\ge1$.
\end{proposition}

\begin{proof}
The proof has four steps.

\emph{Step 1 (Linearization).} By Lemma~\ref{lem:logit}, $\partial_tp_i=p_i\big(\mathcal V_i-\avg{\mathcal V}_p\big)$. Since $D$ is symmetric and stochastic, $D\bar p=\bar p$. Further, $\bar p_i-S_2(\bar p)=0$, so $\mathcal V(\bar p)=0$ and $\bar p$ is an equilibrium. Let $p=\bar p+f$ with $\sum_if_i=0$. Then $S_2=\frac1K+\sum_if_i^2$, so to first order $\mathcal V=\frac1Kf+c(D-\Id)f$, whose entries sum to zero because $D$ preserves sums. Hence $\avg{\mathcal V}_p$ is of second order, and the linearization at $\bar p$ is
\begin{equation}\label{eq:B-linear}
\partial_tf=\frac1K\Big(\frac1Kf+c(D-\Id)f\Big)+O(|f|^2).
\end{equation}

\emph{Step 2 (Fourier modes).} The circulant matrix $D$ is diagonalized by the Fourier modes. On the mode $k\ne0$, the linearization \eqref{eq:B-linear} has the eigenvalue
\begin{equation}
\frac1K\Big(\frac1K-c(1-\hat g_k)\Big).
\end{equation}
Since $1-\hat g_k\ge1-\hat g_1$, all these eigenvalues are negative if and only if $c(1-\hat g_1)>\frac1K$, and some is positive if $c(1-\hat g_1)<\frac1K$.

\emph{Step 3 (Asymptotics).} One has
\begin{equation}
1-\hat g_1=\sum_dg(d)\Big(1-\cos\frac{2\pi d}K\Big),
\end{equation}
and $0\le\frac{y^2}2-(1-\cos y)\le\frac{y^4}{24}$ gives
\begin{equation}
1-\hat g_1=\frac{2\pi^2w_g^2}{K^2}\big(1-\epsilon_K\big),\qquad 0\le\epsilon_K\le\frac{\pi^2\avg{d^4}_g}{3w_g^2K^2},
\end{equation}
which yields \eqref{eq:clin}.

\emph{Step 4 (The Gaussian kernel).} For the Gaussian kernel, the Poisson summation formula shows that $\hat g_k$ is proportional to $\sum_{l\in\Z}e^{-w^2(2\pi k/K-2\pi l)^2/2}$, a wrapped normal density evaluated at the angle $2\pi k/K$. This density decreases on $[0,\pi]$, so $\hat g_1\ge\hat g_k$. The same formula shows that the variance of the discrete Gaussian differs from $w^2$ by an exponentially small relative error, below $10^{-6}$ for $w\ge1$.
\end{proof}

For $w=1.5$, $c_{\rm lin}(K)$ is $2.886$, $5.766$, $11.529$ and $23.057$ for $K=128$, $256$, $512$ and $1024$. This threshold grows with $K$, since a kernel of fixed width in bins acts only at its own scale. In units of the action range, the linear threshold of a leak instead has the finite limit $1/\pi^2$ on a line (Theorem~\ref{thm:D5}).

\paragraph{A line with reflecting ends.}\phantomsection\label{app:inloss-line}
On a line of $K$ bins, the kernel is reflected at the ends, $D^{\rm line}_{ij}=\sum_{l\in\Z}\big(g(i-j+2lK)+g(i+j+1+2lK)\big)$ for bins $0,\dots,K-1$. The ring kernel of $2K$ bins acts on even extensions as $D^{\rm line}$. A policy on the line is therefore an equilibrium at $c$ if and only if its even extension, with half of the mass in each copy, is one on the ring at $\frac c2$. For a policy symmetric about the center of the line, $D^{\rm line}p$ equals $Dp$ for the ring of $K$ bins, up to the truncation of the kernels of the certificates. The interior thresholds on a line are therefore those of that ring, as Theorem~\ref{thm:inloss-cap-full} confirms to $10^{-27}$.

\paragraph{States centered between two bins on a ring.}
We continued the states centered between two bins on a ring in double precision, in the subspace symmetric about their center. We then computed the spectrum of $H$ on the antisymmetric subspace, which contains the translations (Table~\ref{tab:bond}). The lower fold of these states lies just below $c^*$ for $w=3$, $4$ and $8$. Near both of their folds and along the branch between them, the antisymmetric block has a positive eigenvalue, so these states are unstable to translation and do not lower the boundaries on a ring. For $w=1$ and $\frac32$, their branch has no fold, and this eigenvalue is positive wherever double precision resolves it. At the certified folds, the antisymmetric spectrum of the states centered on a bin is negative. These computations are not certified, and for wide peaks, the translation eigenvalue falls below the resolution of double precision.

\begin{table}[!htbp]
\centering\footnotesize
\caption{Folds of the states centered between two bins on rings, in double precision, with the largest eigenvalue of $H$ on the antisymmetric subspace at each fold. The column $c^*$ repeats the certified lower fold of the states centered on a bin.}
\label{tab:bond}
\begin{tabular}{lllllll}
\toprule
$w$ & ring & $c^*$ & lower fold (peak) & eigenvalue & upper fold (peak) & eigenvalue \\
\midrule
$1$, $\frac32$ & $128$ & $0.44720$, $0.29639$ & none & & none & \\
$3$ & $256$ & $0.14775$ & $0.14666$ ($0.1028$) & $5.3\times10^{-3}$ & $0.17437$ ($0.2356$) & $0.166$ \\
$4$ & $256$ & $0.11077$ & $0.11040$ ($0.0762$) & $1.9\times10^{-3}$ & $0.15891$ ($0.2413$) & $0.193$ \\
$8$ & $512$ & $0.05537$ & $0.05534$ ($0.0380$) & $1.9\times10^{-4}$ & $0.13999$ ($0.2466$) & $0.225$ \\
\bottomrule
\end{tabular}
\end{table}

\subsection{Direct simulations of the window}\label{app:inloss-dial}

Long simulations of \eqref{eq:inloss} agree with the certified folds, and Appendix~\ref{app:testbed} compares the broad peaks with the network testbed. For $w=1.5$, a policy started from broad data collapses below $c^*$ into a spike one or two bins wide, and above this threshold, it settles into a broad peak.

Long-time bisection from broad data gives the brackets $[0.2962,0.2965]$ for $w=1.5$ and $[0.1477,0.1483]$ for $w=3$ around the folds $0.29639$ and $0.14775$. The approach is slow near the threshold. At $w=1.5$ and $c=0.285$, $0.29$ and $0.295$, the policy still has $\pmax$ between $0.17$ and $0.19$ after $2\times10^4$ time units and collapses by $5\times10^4$, $5\times10^4$ and $2\times10^5$. For $w=1$, the bisection gave $0.511$, because the broad state itself has $\pmax>0.2$ up to this coefficient.

\paragraph{The hysteresis window.}
For $w=1.5$, a spike of mass $0.99$ remains a spike up to $c=0.350$ and dissolves at $c=0.355$ for $K=128$ and $256$. Broad data stop collapsing at $0.2963$, in agreement with the certified window $[0.29639,0.35252]$. Between $c=0.30$ and $0.35$, the spike has $\pmax$ between $0.68$ and $0.52$ and the peak between $0.180$ and $0.127$. Theorem~\ref{thm:inloss-cap-full}(c) certifies both as stable at $c=0.30$, $0.31$, $0.33$ and $0.35$. This window is narrower than that of a leak with expected gradients, whose ratio is $1.59$ to $1.71$ (Appendix~\ref{sec:hysteresis}). For $w\gtrsim2.5$, it is wider. At $w=3$ and $K=256$, broad data collapse at $c=0.144$ and $0.146$ and stay broad at $0.150$ and $0.152$. A spike of mass $0.99$ persists at $c=0.288$ and $0.290$ and dissolves at $0.294$ and $0.296$ into a broad peak.

\subsection{Unordered answers}\label{app:finite}\label{sec:finite}

Answers of a language model have no natural order, so exploration cannot use neighbors. Proposition~\ref{prop:potts-uniform} gives the thresholds of smoothing toward the uniform policy. On unordered answers, sampling temperature only rescales time, whereas an entropy bonus acts as exploration with a window that Section~\ref{sec:related} relates to the Potts model.

\begin{proposition}[Temperature is a change of time]\label{prop:temperature}
Suppose that samples are drawn from $\mathrm{softmax}(z/\tau)$ and that log-probabilities are computed at the same temperature $\tau$. Then $\tilde z=z/\tau$ follows the vanilla flow at $\tau=1$ with time multiplied by $\tau^{-2}$, and the NPG flow of $\tilde z$ does not depend on $\tau$.
\end{proposition}

\begin{proof}
Let $\mathcal J$ be the objective and $p=\mathrm{softmax}(\tilde z)$ with $\tilde z=z/\tau$. The gradient of $\mathcal J$ with respect to $z$ is $\frac1\tau\nabla_{\tilde z}\mathcal J$. Gradient ascent $z'=\nabla_z\mathcal J$ therefore gives $\tilde z'=\frac1{\tau^2}\nabla_{\tilde z}\mathcal J(\tilde z)$. The Fisher matrix in $z$ is $\mathcal I_z=\mathcal I_{\tilde z}/\tau^2$, so the natural gradient step is $z'=\mathcal I_z^{+}\nabla_z\mathcal J=\tau\,\mathcal I_{\tilde z}^{+}\nabla_{\tilde z}\mathcal J$, and $\tilde z'=\mathcal I_{\tilde z}^{+}\nabla_{\tilde z}\mathcal J$ does not depend on $\tau$.
\end{proof}

Temperature thus changes the speed of collapse, not its outcome. Adam is invariant to the scale of the gradient up to its constant $\epsilon_{\rm A}$. Under Adam, $\tau^{-1}$ thus cancels in the step on $z$, so $\tilde z$ follows the dynamics at $\tau=1$ with learning rate $\eta/\tau$.

In 64 tabular conditions ($\tau\in\{0.6,1,1.2,2\}$, SGD and Adam, matched and mismatched temperatures, two initial policies and two rewards), 62 of the 64 runs collapsed onto the same mode, and only the time changed. For example, the time until the policy $\mathrm{softmax}(z)$ at temperature one has $1-p_{\rm mode}<10^{-3}$ fell from 950 to 300 to 100 steps for $\tau=1$, $1.2$ and $2$.

\paragraph{The entropy bonus is a Potts model.}\label{app:potts}
An entropy bonus with coefficient $c_{\rm H}$ changes NPG to
\begin{equation}
z_i'=(R_i-\avg{R}_p)-c_{\rm H}(\log p_i-\avg{\log p}_p).
\end{equation}
Let $\mathrm{LM}(a,b)=(a-b)/(\log a-\log b)$ be the logarithmic mean, and let $[P]$ equal $1$ if $P$ holds and $0$ otherwise.

\begin{theorem}[Equilibria under an entropy bonus]\label{thm:potts}
Let $n\ge3$.
\begin{enumerate}[label=(\alph*),leftmargin=*,itemsep=1pt]
\item For the agreement reward, the equilibria satisfy $p_i\propto e^{p_i/c_{\rm H}}$, the mean-field Potts equation \citep{wu1982potts}, whose solutions are the symmetric logit quantal response equilibria of the coordination game with rationality $1/c_{\rm H}$ \citep{mckelvey1995qre}. The uniform policy is linearly stable if and only if $c_{\rm H}>1/n$. Equilibria of the form $(M,p_{\rm r},\dots,p_{\rm r})$ with $M>p_{\rm r}=(1-M)/(n-1)$ exist exactly when $c_{\rm H}=\mathrm{LM}(M,p_{\rm r})$. Hence they exist for some $M$ if and only if
\begin{equation}\label{eq:betas}
c_{\rm H}\le c_{\rm H}^{\rm s}(n)=\max_{M\in(1/n,1)}\mathrm{LM}\Big(M,\frac{1-M}{n-1}\Big),
\end{equation}
and $c_{\rm H}^{\rm s}(n)\log n\to1$ as $n\to\infty$.
\item For the majority reward in the limit of large groups, $R_i=[i=\arg\max_jp_j]$, the unique equilibrium whose most likely answer is $1$ has $p_1=e^{1/c_{\rm H}}/(e^{1/c_{\rm H}}+n-1)$, with all other entries equal.
\end{enumerate}
\end{theorem}

For the agreement reward, the uniform state and the collapsed branch coexist for $1/n<c_{\rm H}<c_{\rm H}^{\rm s}(n)$, a window whose ratio of endpoints grows like $n/\log n$. Here $c_{\rm H}^{\rm s}(n)=0.311$, $0.193$, $0.144$ and $0.117$ for $n=4$, $16$, $64$ and $256$, so that $c_{\rm H}^{\rm s}(n)\log n$ equals only $0.43$, $0.54$, $0.60$ and $0.65$, and the limit is a poor approximation for moderate $n$.

In tabular simulations, each transition from a uniform or a peaked start lies in a grid bracket with $1/n$ or $c_{\rm H}^{\rm s}(n)$ at one end, for example $(0.250,0.288)$ and $(0.311,0.340)$ for $n=4$. The mode probabilities on the collapsed branch equal the theoretical values to three digits, except at the fold, where they differ by at most $0.002$. For the large-group majority reward, the measured equilibria equal part (b) to three digits, without hysteresis.

With groups of $16$ samples, the expected advantage of the majority reward is a polynomial in $p$. On policies with one leading answer and equal others, the entropy bonus with $n=10$ then has a window from $0.363$ to $0.546$ (Appendix~\ref{app:majority-exact}).

\begin{proof}[Proof of Theorem~\ref{thm:potts}]
The proof has five steps. Steps 1 to 4 prove part (a), and Step 5 proves part (b).

\emph{Step 1 (The Potts equation).} Equilibria satisfy
\begin{equation}
(p_i-S_2)-c_{\rm H}\log p_i=\mathrm{const},
\end{equation}
so $p_i-c_{\rm H}\log p_i$ does not depend on $i$, which is equivalent to $p_i\propto e^{p_i/c_{\rm H}}$.

\emph{Step 2 (Stability of the uniform policy).} At $\bar p=\frac1n\mathbf1$, let $f$ be a zero-mean perturbation of $z$. To first order, it changes $p_i$ by $f_i/n$, leaves $S_2$ unchanged and changes $\log p_i$ by $n$ times the change of $p_i$. Hence $f'=(\frac1n-c_{\rm H})f$, which is stable if and only if $c_{\rm H}>\frac1n$.

\emph{Step 3 (One leading answer).} For $p=(M,p_{\rm r},\dots,p_{\rm r})$ with $M>p_{\rm r}$, the equilibrium condition is
\begin{equation}
M-c_{\rm H}\log M=p_{\rm r}-c_{\rm H}\log p_{\rm r},
\end{equation}
that is, $c_{\rm H}=\mathrm{LM}(M,p_{\rm r})$. The function $\mathrm{LM}_n(M)=\mathrm{LM}(M,\frac{1-M}{n-1})$ is continuous on $(\frac1n,1)$, with $\mathrm{LM}_n(M)\to\frac1n$ as $M\downarrow\frac1n$ and $\mathrm{LM}_n(M)\to0$ as $M\uparrow1$. Its range is therefore $(0,c_{\rm H}^{\rm s}(n)]$.

\emph{Step 4 (Asymptotics).} Since $\mathrm{LM}(M,p_{\rm r})\le\frac{M+p_{\rm r}}2\le M$, one has $\mathrm{LM}_n(M)<1/\log n$ for $M<1/\log n$. For $M\ge1/\log n$, $\mathrm{LM}(M,p_{\rm r})\le M/\log(M/p_{\rm r})\le1/\log(M/p_{\rm r})$ and $M/p_{\rm r}\ge M(n-1)\ge(n-1)/\log n$, so $\mathrm{LM}_n(M)\le1/(\log(n-1)-\log\log n)$. Conversely, $M=1-1/\log n$ gives $\mathrm{LM}_n(M)\ge(1-2/\log n)/\log(n\log n)$. Both bounds are $(1+o(1))/\log n$.

\emph{Step 5 (The majority reward).} In the limit of large groups, the advantages are $1-p_1$ and $-p_1$, so the equilibria satisfy
\begin{equation}
(1-p_1)-c_{\rm H}\log p_1=-p_1-c_{\rm H}\log p_j
\end{equation}
for every $j\ge2$. Hence $p_j=p_1e^{-1/c_{\rm H}}$, and normalization gives the formula, unique given the most likely answer.
\end{proof}

\section{Normalization and symmetric rewards}\label{app:norm}
The standard deviation \eqref{eq:sigmaG} of Appendix~\ref{app:norm-mf} underlies Section~\ref{sec:exploration}, Proposition~\ref{prop:norm} and Sections~\ref{sec:unordered} and \ref{sec:norm-reward}. The exact thresholds of Appendices~\ref{app:majority-exact} and \ref{app:bonus} are the predicted values with which Appendices~\ref{app:majority-ramp} and \ref{app:ttrl} compare the runs.

\subsection{The mean-field standard deviation}\label{app:norm-mf}

Draw a group of $G$ samples $a_1,\dots,a_G$ from $p$, and give sample $j$ the leave-one-out agreement reward
\begin{equation}
r_j=\frac1{G-1}\sum_{k\ne j}[a_k=a_j],
\end{equation}
where $[a_k=a_j]$ is $1$ if the samples agree and $0$ otherwise. With the mean $\bar r$ of the group as baseline, the logit update is $\frac1G\sum_j(r_j-\bar r)e_{a_j}$, because the term $-p$ of the gradient of $\log p_{a_j}$ cancels when the advantages sum to zero. Its mean is $\frac{G-2}G\,p\odot(p-S_2)$. Indeed, given $a_j=i$, the mean of $r_j$ is $p_i$. For $k\ne j$, the mean of $r_k$ is $(p_i+(G-2)S_2)/(G-1)$, because sample $k$ agrees with sample $j$ with probability $p_i$ and with each of the other $G-2$ samples with probability $S_2$. Hence the mean of $\bar r$ is $(2p_i+(G-2)S_2)/G$, and the mean advantage of a sample in bin $i$ is
\begin{equation}
\frac{G-2}G(p_i-S_2) .
\end{equation}

Given $a_j=i$, the reward $r_j$ is binomial with mean $p_i$ and variance $p_i(1-p_i)/(G-1)$. Over the draw of $a_j$, its variance is therefore
\begin{align}
\sum_ip_i(p_i-S_2)^2+\sum_ip_i\frac{p_i(1-p_i)}{G-1}&=\big(S_3-S_2^2\big)+\frac{S_2-S_3}{G-1}\nonumber\\
&=\varsigma_G(p)^2 .
\end{align}

GRPO divides the advantages by the unbiased standard deviation $s$ of the rewards of the group, plus $10^{-6}$. This estimates $\varsigma_G(p)$, but it is correlated with the advantages. We therefore measure the normalized update directly. Let $\mathcal F=p\odot(p-S_2)$, and define $\varsigma_{\rm eff}$ by projecting the mean normalized update on $\mathcal F$, so that its component along $\mathcal F$ is
\begin{equation}
\frac{G-2}G\,\mathcal F/\varsigma_{\rm eff} .
\end{equation}

\begin{table}[!htbp]
\centering\footnotesize
\caption{Monte Carlo check of the normalized update at states along the curves of equilibria of Appendix~\ref{app:norm-unfold}, with $G=64$ and $20000$ groups per state. The states include some near the upper fold for $w=3$ and the lower fold for $w=\frac32$. The cosine is between the mean normalized update and $\mathcal F$. The last column is the fraction of groups whose rewards are all equal.}
\label{tab:norm-mc}
\begin{adjustbox}{max width=\linewidth}
\begin{tabular}{lllllll}
\toprule
kernel & $\pmax$ & $\varsigma_G$ & $\varsigma_{\rm eff}/\varsigma_G$ & cosine & groups with $s=0$ \\
\midrule
$w=3$ & $0.024$ & $0.0172$ & $0.983$ & $0.999$ & $0$ \\
 & $0.129$ & $0.0464$ & $1.013$ & $0.998$ & $0$ \\
 & $0.328$ & $0.1405$ & $0.965$ & $1.000$ & $0$ \\
 & $0.382$ & $0.1715$ & $0.964$ & $1.000$ & $0$ \\
 & $0.494$ & $0.2332$ & $0.967$ & $1.000$ & $0$ \\
 & $0.804$ & $0.3153$ & $0.980$ & $1.000$ & $0$ \\
 & $0.932$ & $0.2344$ & $1.009$ & $1.000$ & $0.012$ \\
 & $0.972$ & $0.1626$ & $1.096$ & $1.000$ & $0.158$ \\
$w=\frac32$ & $0.048$ & $0.0261$ & $0.982$ & $0.999$ & $0$ \\
 & $0.203$ & $0.0734$ & $0.991$ & $0.999$ & $0$ \\
 & $0.449$ & $0.1914$ & $0.967$ & $1.000$ & $0$ \\
 & $0.929$ & $0.2367$ & $1.004$ & $1.000$ & $0.010$ \\
 & $0.970$ & $0.1656$ & $1.091$ & $1.000$ & $0.144$ \\
\bottomrule
\end{tabular}
\end{adjustbox}
\end{table}

The mean normalized update is parallel to $\mathcal F$, with cosines of at least $0.998$. Along both curves, $\varsigma_{\rm eff}$ lies between $0.964$ and $1.013$ times $\varsigma_G$ as long as $\pmax\le0.94$ (Table~\ref{tab:norm-mc}). At $\pmax=0.97$, it is about 10 percent larger, because 14 to 16 percent of the groups then have equal rewards and contribute nothing. The unnormalized update agrees with $\frac{G-2}G\,\mathcal F$ to within $3.2$ percent, the size of the Monte Carlo error.

\subsection{The folds unfold}\label{app:norm-unfold}

\begin{observation}[Normalizing only the reward removes the folds for large groups]\label{obs:unfold}
Let $D=D_w$ on a ring of $256$ bins, and consider $z'=\frac{G-2}G\,p\odot(p-S_2)/\varsigma_G(p)+c\,(Dp-p)$ for $G$ from $4$ to $256$. For $w=3$, the curve of site-symmetric equilibria through the broad state has folds only for $G\le29$, with window ratios of at most $1.147$. For $G\ge32$, and for $w=\frac32$ at every computed $G$, it has no fold. For $w=3$, the coefficient at which the equilibrium has $\pmax=0.2$ grows from $1.29$ at $G=8$ to $2.97$ at $G=256$ (Table~\ref{tab:norm-unfold}).
\end{observation}

We followed the curve of site-symmetric equilibria of \eqref{eq:inloss} on a ring of $256$ bins by pseudo-arclength continuation, with residuals below $10^{-13}$, from the broad state at $c=1.2/w$ through both folds to the spike branch. Its folds are $0.14775$ and $0.29194$ for $w=3$ and $0.29639$ and $0.35252$ for $w=\frac32$, as in Theorem~\ref{thm:inloss-cap}. By Proposition~\ref{prop:norm}(b), a point $p$ of this curve at the coefficient $c_{\rm lat}$ is an equilibrium of the flow of Observation~\ref{obs:unfold} at the coefficient
\begin{equation}\label{eq:C-map}
c=\frac{G-2}G\,\frac{c_{\rm lat}}{\varsigma_G(p)} .
\end{equation}
Table~\ref{tab:norm-unfold} gives the windows of the normalized curve and the coefficient where the equilibrium has $\pmax=0.2$, as in the testbed.

\begin{table}[!htbp]
\centering\footnotesize
\caption{Bistable windows and the coefficient at $\pmax=0.2$ on the curve of equilibria with only the reward normalized, ring of $256$ bins. A window is the range of $c$ between the two folds of the normalized curve, with its ratio in parentheses, and none means that the curve is monotone. Groups of 25 to 29 also have a narrow window. Empty cells were not computed. Without normalization, the windows are $[0.143,0.283]$ for $w=3$ and $[0.287,0.342]$ for $w=\frac32$ in the units of $G=64$.}
\label{tab:norm-unfold}
\begin{adjustbox}{max width=\linewidth}
\begin{tabular}{lllll}
\toprule
$G$ & window, $w=3$ & $c$ at $\pmax=0.2$, $w=3$ & window, $w=\frac32$ & $c$ at $\pmax=0.2$, $w=\frac32$ \\
\midrule
$4$ & $[0.541,0.621]$ ($1.147$) & & none & \\
$8$ & $[1.201,1.292]$ ($1.076$) & $1.29$ & none & $1.67$ \\
$16$ & $[1.929,1.972]$ ($1.022$) & $1.89$ & none & $2.56$ \\
$24$ & $[2.367,2.377]$ ($1.004$) & & none & \\
$32$ & none & $2.36$ & none & $3.35$ \\
$64$ & none & $2.68$ & none & $3.95$ \\
$128$ & none & $2.87$ & none & $4.35$ \\
$256$ & none & $2.97$ & none & $4.59$ \\
\bottomrule
\end{tabular}
\end{adjustbox}
\end{table}

By \eqref{eq:C-map}, a fold of the normalized curve is a point where
\begin{equation}
\log c_{\rm lat}-\log\varsigma_G
\end{equation}
is stationary along the curve. At the folds of the unnormalized curve, $\log c_{\rm lat}$ is stationary, while $\varsigma_G$ increases as the peak sharpens. Along the middle branch, $c_{\rm lat}$ grows by a factor of $1.98$ for $w=3$ and $1.19$ for $w=\frac32$. For $G=64$, $\varsigma_G$ grows from $0.0416$ to $0.2316$ and from $0.0734$ to $0.2025$, by factors of $5.6$ and $2.8$. For $G=64$, the normalized curve is therefore monotone, and only for $G\le29$ and $w=3$ does a narrow window remain.

Integrated for $w=3$ and $G=64$ from a broad Gaussian of width $0.1$ of the range and a spike of mass $0.99$, the normalized flow reached the same state from both. This state has $\pmax=0.580$, $0.403$, $0.293$, $0.221$ and $0.165$ for $c=1$, $1.5$, $2$, $2.5$ and $3$.

\subsection{Normalizing both terms with sampled groups}\label{app:norm-both}

The folds of Section~\ref{sec:twoscales} need a correction for sampled standard deviations when both terms are divided.

When the exploration term is divided by the same $s$ and groups with $s=0$ are dropped, as in the testbed and in arm B of the language-model test, the expected update is
\begin{equation}
\frac{G-2}G\,\mathcal F/\varsigma_{\rm eff}+(c/\varsigma_{\rm h})(Dp-p)\quad\text{with}\quad\varsigma_{\rm h}=1/\E\big[[s>0]/(s+10^{-6})\big] .
\end{equation}
As in Proposition~\ref{prop:norm}(b), its equilibria at the coefficient $c$ are those of \eqref{eq:inloss} at $c_{\rm lat}=c\,\varsigma_{\rm eff}/(\frac{G-2}G\varsigma_{\rm h})$. In the units of the testbed, the folds therefore move from $\frac{G-2}G\,c_{\rm lat}$ to
\begin{equation}
\frac{G-2}G\,c_{\rm lat}\,\varrho ,
\end{equation}
with $\varrho=\varsigma_{\rm h}/\varsigma_{\rm eff}$ at the fold.

For $w=3$ and $G=64$, Monte Carlo estimates with $30000$ groups per state give $\varrho$ between $0.83$ and $0.93$ near the lower fold and between $0.95$ and $0.995$ near the upper fold. With $\varrho$ at the two fold states of the unnormalized flow, we estimate the folds at $0.131$ and $0.276$, against $0.143$ and $0.283$ without normalization, with the ratio $2.11$ instead of $1.98$. These are estimates, not bounds. The folds of the sampled flow are the turning points of $c_{\rm lat}\varrho$ along the curve, and we did not follow $\varrho$ along it, so whether both folds persist is not established. If they do and $\varrho$ stays in the range above, the lower fold lies between about $0.119$ and $0.133$ and the upper one between $0.269$ and $0.281$.

\subsection{Unordered answers}\label{app:norm-unordered}

For $n$ unordered answers with smoothing toward $\bar p$, normalizing only the reward gives, in mean field, $z'=\frac{G-2}G\,p\odot(p-S_2)/\varsigma_G(p)+c\,(\bar p-p)$. At $\bar p$, one has $\varsigma_G(\bar p)^2=(\frac1n-\frac1{n^2})/(G-1)$, and since $p\odot(p-S_2)$ vanishes at $\bar p$, the linearization of the normalized term is that of the unnormalized one divided by $\varsigma_G(\bar p)$. The uniform policy is therefore linearly stable if and only if
\begin{equation}
c>\frac{G-2}G\big/(n\,\varsigma_G(\bar p)) ,
\end{equation}
which is $1.130$ for $n=10$ and $G=16$. By Proposition~\ref{prop:norm}(b), the states with one entry $M$ and $n-1$ equal entries are equilibria at
\begin{equation}
c=\frac{G-2}G\,\frac n{n-1}\,M(1-M)/\varsigma_G ,
\end{equation}
with $\varsigma_G$ evaluated at that state. For $n=10$ and $G=16$, this coefficient rises from $1.130$ at $M=\frac1n$ to its maximum $1.698$ at $M=0.210$ and then decreases, to $1.525$ at $M=0.3$. Its decreasing part has $\pmax=0.949$, $0.721$, $0.547$, $0.414$, $0.342$ and $0.273$ at $c=0.22$, $0.6$, $0.9$, $1.2$, $1.4$ and $1.6$.

A window therefore exists in mean field, but the basin of the uniform policy is narrow. From one draw of logits equal to $0.3$ times independent standard normal variables, the normalized mean-field flow reached the decreasing branch for every $c\le1.4$ of our grid and the uniform policy for $c\ge1.6$. A start closer to uniform returns to it for $c$ just above $1.13$. The free logit table of the language-model test, which samples groups, shows no hysteresis in arm R (Appendix~\ref{app:llm-test}).

The sampled standard deviation departs more from $\varsigma_G$ for ten answers and $G=16$ than on the ordered curves of Table~\ref{tab:norm-mc}. Monte Carlo estimates with $400000$ groups per state, for states with one entry $M$ and nine equal entries, put $\varsigma_{\rm eff}/\varsigma_G$ between $0.89$ and $1.10$. Exact sums over the counts of a group with the unbiased standard deviation of the code give the stability threshold $1.1005$ and the release fold $1.5926$ for arm R. For arm B, they give $0.0690$ and $0.2194$ (Appendix~\ref{app:majority-exact}).

\subsection{Exact thresholds for symmetric rewards}\label{app:majority-exact}

Sections~\ref{sec:unordered} and \ref{sec:llm} use the thresholds of Proposition~\ref{prop:symmetric} computed exactly for finite groups.

For $p=(M,y,\dots,y)$ with $y=\frac{1-M}{n-1}$, the expected advantage $\mathcal A(M)$ of Proposition~\ref{prop:symmetric} is a finite sum over the count vectors of a group, which are multinomial with parameters $G$ and $p$. For $n=10$ and $G=16$, we enumerated all $2042975$ of them. The agreement reward is leave-one-out. The majority reward gives $1$ to the samples of the most frequent answer and $0$ to the others, with ties broken at random.

Whichever tied answer wins, the realized rewards of a group have the mean $f$, the count of the winning answer divided by $G$. Their standard deviation is that of values $0$ and $1$ with this mean. Arm U subtracts the group mean, and arm R divides the advantages by $s+10^{-6}$, with $s$ the unbiased standard deviation of the realized rewards, as our implementation does. The expected advantage of a sample of a tied answer is therefore its probability of winning the tie minus $f$, divided in arm R by $s+10^{-6}$. A standard deviation of the expected rewards, with $1/T$ for each sample of $T$ tied answers, would instead give $0.950$ and $1.661$ for arm R. Arm B also divides the smoothing term of each group by $s+10^{-6}$ and drops groups with $s=0$, so its $c_{\rm eq}$ is divided by the mean of $[s>0]/(s+10^{-6})$. The thresholds of Table~\ref{tab:symmetric} are
\begin{equation}
\text{stability threshold}=\lim_{M\downarrow\frac1n}c_{\rm eq}(M),\qquad\text{release fold}=\max_{\frac1n<M<1}c_{\rm eq}(M).
\end{equation}

For $G=64$, a dynamic program over the multiplicities of the counts of the other answers gives the expected advantage of a leading sample and agrees with the enumeration at $G=16$. Monte Carlo estimates with $4\times10^5$ groups, from the code of the language-model test, agree with the enumeration to about $7\times10^{-4}$ per answer, with and without normalization.

\begin{table}[!htbp]
\centering\footnotesize
\caption{Thresholds of smoothing toward the uniform policy for $n=10$ answers (Proposition~\ref{prop:symmetric}), exact sums over the counts of a group. Each cell gives the stability threshold of the uniform policy, the limit of $c_{\rm eq}$ at $M=\frac1n$, and the release fold, the maximum of $c_{\rm eq}$. Arm U divides nothing. Arm R divides the advantages by the unbiased standard deviation of the group, and arm B also divides the smoothing term and drops groups with equal rewards. The empty cell was not computed.}
\label{tab:symmetric}
\begin{adjustbox}{max width=\linewidth}
\begin{tabular}{llll}
\toprule
$G$ & arm & agreement & majority \\
\midrule
$16$ & U & $0.0875$, $0.2431$ & $0.3632$, $0.7661$ \\
$16$ & R & $1.1005$, $1.5926$ & $0.8063$, $1.5866$ \\
$16$ & B & $0.0690$, $0.2194$ & $0.3481$, $0.7374$ \\
$64$ & U & $0.0969$, $0.2691$ & $0.6233$, $1.2902$ \\
$64$ & R & & $1.6338$, $3.1411$ \\
\bottomrule
\end{tabular}
\end{adjustbox}
\end{table}

For the majority reward with $G=16$, the folds lie at $M=0.295$, $0.280$ and $0.308$ in arms U, R and B, against $M=0.5$, $0.247$ and $0.590$ for the agreement reward. On the stable branch beyond the fold, the leading answer of arm U has $\pmax=0.813$, $0.724$, $0.640$, $0.560$, $0.483$ and $0.400$ at $c=0.2$, $0.3$, $0.4$, $0.5$, $0.6$ and $0.7$. In arm R, it has $0.860$, $0.757$, $0.654$, $0.562$, $0.481$ and $0.403$ at $c=0.4$, $0.6$, $0.8$, $1.0$, $1.2$ and $1.4$.

By the symmetry among the trailing answers, all directions that separate them share one eigenvalue. On the stable side of every fold for $G=16$, we evaluated it at three points, and it was negative at all of them. For the majority reward in arm U, it is $-0.067$, $-0.060$ and $-0.043$ at $M=0.366$, $0.471$ and $0.612$. These computations do not show that solutions off the invariant set reach it.

With an entropy bonus in place of the smoothing, the same enumeration gives, for the majority reward in arm U, the stability threshold $0.3632$ and the release fold $0.5456$ at $M=0.265$. The window of Theorem~\ref{thm:potts}(a) therefore has a counterpart for the majority reward with finite groups.

\subsection{An answer-level bonus}\label{app:bonus}

The thresholds of the bonus in Section~\ref{sec:norm-reward} and the predictions of the TTRL test in Appendix~\ref{app:ttrl} are derived below.

For free-form answers, a sample $j$ with a boxed answer receives the bonus $b_j=-\log(N_{a_j}/G_{\rm v})$. Here $G_{\rm v}$ is the number of samples of the group that have an answer, equal to $G$ when every sample has one, and $N_{a_j}$ is the number with the same answer. The group mean $\bar b$ is then the plug-in entropy of the answers. Arm C adds $c_{\rm b}\,b$ to the majority reward and normalizes the sum as in GRPO, with the population standard deviation that our runs use. Arm R normalizes the reward advantages and adds $c_{\rm b}(b-\bar b)$ afterward. Arm U adds $c_{\rm b}(b-\bar b)$ to the reward advantages without any division, and arm B divides this sum by the standard deviation of the rewards. Arms U and B have the same threshold near collapse as arm C.

\paragraph{The threshold near collapse.}
Near collapse, a group that contributes to the update has, to leading order, $G-1$ samples of the leading answer and one other. The samples of the leading answer have the majority reward $1$ and the bonus $\log\frac G{G-1}$, and the other sample has $0$ and $\log G$. In arm C, the group takes two values. Normalizing a group in which one sample takes one value and $G-1$ samples the other gives the single sample the advantage $\sqrt{G-1}$ times the sign of the difference between its value and the other. This difference is
\begin{equation}
c_{\rm b}\log G-1-c_{\rm b}\log\frac G{G-1}=c_{\rm b}\log(G-1)-1,
\end{equation}
so the other answer gains if and only if
\begin{equation}
c_{\rm b}>c_{\rm b}^*=1/\log(G-1) .
\end{equation}

In arm R, the normalized reward advantage of the other sample is $-\sqrt{G-1}$, and its bonus advantage is $c_{\rm b}\frac{G-1}G\log(G-1)$. It therefore gains if and only if
\begin{equation}
c_{\rm b}\log(G-1)>G/\sqrt{G-1},
\end{equation}
so the threshold of arm R is $G/\sqrt{G-1}$ times $c_{\rm b}^*$, whatever the number of answers. For $G=8$, $c_{\rm b}^*=0.514$, and the threshold of arm R is $1.554$. When the advantages use only the samples with an answer, a group with one or two samples without an answer has the threshold $0.558$ or $0.621$ in arm C. With the unbiased standard deviation in place of the population one, the same calculation gives the factor $\sqrt G$ instead of $G/\sqrt{G-1}$. The official TTRL code divides by the unbiased standard deviation, which is $\sqrt{G/(G-1)}$ times the population one. This multiplies every threshold of arm R by $\sqrt{(G-1)/G}$. For free-form answers and $G=8$, it moves the threshold near collapse from $1.554$ to $1.454$ and the end of the branch from $1.375$ to $1.286$. Arms C, U and B keep their thresholds, since they divide whole advantages or nothing.

\paragraph{Interior equilibria and windows.}
In arm C, a group in which the reward plus the bonus takes only two values has normalized advantages that do not depend on the gap between them. The bonus then acts only through their order. With two answers, or when all answers other than the leading one are distinct, arm C below $c_{\rm b}^*$ therefore has exactly the expected flow of $c_{\rm b}=0$. On policies with one leading answer and equal others, the coefficient at which such a policy is an equilibrium decreases in arms C, B and U beyond its maximum over $M$. It tends to $c_{\rm b}^*$ as $M\to1$. Between $c_{\rm b}^*$ and this maximum, the policy therefore stays at an interior equilibrium with $M<1$, and there is no hysteresis at $c_{\rm b}^*$.

\paragraph{Free-form answers in arm C.}
For free-form answers, $n=\infty$, the samples other than those of the leading answer are all distinct. A group in which the leading answer appears $k\ge2$ times then takes two values in arm C, and normalization gives its leading samples together an advantage proportional to $\sqrt{k(G-k)}$, with the sign of $1-c_{\rm b}\log k$. Groups with $k\le1$ or $k=G$ contribute nothing on average, so the velocity of the leading answer has the sign of
\begin{equation}
\sum_{k=2}^{G-1}\binom Gk M^k(1-M)^{G-k}\sqrt{k(G-k)}\,\operatorname{sign}(1-c_{\rm b}\log k) .
\end{equation}
The sign factor changes once in $k$, from positive to negative, and binomial weights are totally positive of order two in $k$ and $M$, so this sum changes sign at most once in $M$ \citep{karlin1968total}. Near $M=0$, the term $k=2$ dominates, and near $M=1$, the term $k=G-1$. For every $G$, arm C therefore has exactly one interior equilibrium among policies with one leading answer when $c_{\rm b}^*<c_{\rm b}<1/\log2\approx1.443$, and it is stable. This upper end is $2.8$ to $6.0$ times $c_{\rm b}^*$ for $G$ from 8 to 64. Above it, even two agreeing samples lose to distinct ones, and the policy spreads over distinct answers. With $n=10$, the uniform policy becomes stable at $c_{\rm lin}$, which is at least $1.77\,c_{\rm b}^*$ in Table~\ref{tab:bonus-window}. In both cases, a bonus before the normalization between $c_{\rm b}^*$ and $1.7\,c_{\rm b}^*$ leaves the interior equilibrium as the only stable state on these policies.

Arm R behaves differently. Its coefficient $c_{\rm eq}$ falls beyond its maximum to a minimum at the end of the branch of interior equilibria and then rises to the threshold near collapse as $M\to1$. Between this end and the threshold near collapse, collapse and an interior equilibrium are both stable. For $G=8$, the branch ends at $1.521$, $1.454$ and $1.375$ for $n=5$, $10$ and $\infty$, just below the threshold $1.554$. The coefficient scan of Appendix~\ref{app:sweep} ran arm R at $1.45$ inside this narrow window for problems with more than ten answers. For larger groups, the window widens.

\begin{observation}[A window for the bonus after the normalization]\label{obs:bonus-window}
On policies with one leading answer and equal others, exact enumeration with $n=10$ answers and the population standard deviation gives the following for the bonus added after the normalization. The branch of interior equilibria ends at a coefficient between $2.8$ and $3.2$ times $c_{\rm b}^*$ for $G$ from $8$ to $64$, while collapse stays stable on this set up to $G/\sqrt{G-1}$ times $c_{\rm b}^*$ (Table~\ref{tab:bonus-window}). Between the two, collapse and an interior equilibrium are both stable on this set, and the window widens with $G$. For $G\ge16$, the uniform policy is also stable in its upper part, above $c_{\rm lin}$. For free-form answers, $n=\infty$, the branch ends between $2.3$ and $2.7$ times $c_{\rm b}^*$. With the bonus before the normalization, $c_{\rm eq}$ decreases from its maximum down to $c_{\rm b}^*$ at collapse, so collapse and an interior equilibrium are never both stable on this set.
\end{observation}

\begin{table}[!htbp]
\centering\footnotesize
\caption{Thresholds of the answer-level bonus on policies with one leading answer and equal others, from exact sums over the counts of a group with the population standard deviation and $n=10$ unless stated. Before the normalization (arm C), collapse is released and prevented at the same coefficient $c_{\rm b}^*=1/\log(G-1)$. After it (arm R), a broad policy is kept from collapsing above the end of the branch of interior equilibria, with its leading mass $M$ there in parentheses. A collapsed policy is released only above the threshold near collapse. In both arms, the uniform policy is linearly stable above $c_{\rm lin}$, and $c_{\max}$ is the largest coefficient with an interior equilibrium. Rows marked unbiased use the unbiased standard deviation of the TTRL code. The last two rows divide the two thresholds of arm R by $c_{\rm b}^*$. For free-form answers, the interior branch of arm C ends at $1/\log2$ for every $G$.}
\label{tab:bonus-window}
\begin{tabular}{lllll}
\toprule
 & $G=8$ & $G=16$ & $G=32$ & $G=64$ \\
\midrule
$c_{\rm b}^*=1/\log(G-1)$ & $0.514$ & $0.369$ & $0.291$ & $0.241$ \\
arm C, $c_{\rm lin}$ & $0.910$ & $0.676$ & $0.635$ & $0.714$ \\
arm C, $c_{\max}$ & $0.910$ & $0.837$ & $0.915$ & $1.121$ \\
arm C, $c_{\max}$, $n=\infty$ & $1.443$ & $1.443$ & $1.443$ & $1.443$ \\
\midrule
arm R, end of branch ($M$ there) & $1.454$ ($0.81$) & $1.074$ ($0.73$) & $0.876$ ($0.73$) & $0.780$ ($0.75$) \\
arm R, end of branch, $n=\infty$ & $1.375$ & $0.938$ & $0.705$ & $0.565$ \\
arm R, near collapse & $1.554$ & $1.526$ & $1.674$ & $1.946$ \\
arm R, near collapse, unbiased & $1.454$ & $1.477$ & $1.647$ & $1.931$ \\
arm R, end of branch, $n=\infty$, unbiased & $1.286$ & $0.908$ & $0.694$ & $0.561$ \\
arm R, $c_{\rm lin}$ & $1.858$ & $1.497$ & $1.519$ & $1.829$ \\
arm R, $c_{\max}$ & $1.892$ & $1.805$ & $2.104$ & $2.758$ \\
\midrule
arm R, end over $c_{\rm b}^*$ & $2.83$ & $2.91$ & $3.01$ & $3.23$ \\
arm R, near collapse over $c_{\rm b}^*$ & $3.02$ & $4.13$ & $5.75$ & $8.06$ \\
\bottomrule
\end{tabular}
\end{table}

The letter test of Appendix~\ref{app:letter-bonus} probes both thresholds of the two arms for $G$ from $8$ to $64$.

\paragraph{The official setting.}
The original TTRL trains on groups of 32 and divides by the unbiased standard deviation. A bonus added to it would have, by these sums, the threshold $c_{\rm b}^*=0.291$ in arm C and $1.647$ near collapse in arm R. The original votes over 64 samples, and with the vote over the training group instead, the branch of arm R would end at $0.694$ for free-form answers. The official handling of samples without an answer breaks the symmetry that these values assume, as the next paragraph shows, so they are untested predictions for that setting.

\paragraph{Samples without an answer.}
If samples without an answer receive the reward $0$ and enter the group statistics, as in the official TTRL code, a group of $G-1$ samples of the leading answer and one such sample pushes that answer up. The push is as hard as that of one other answer in arm C and harder in arm R. These samples break the symmetry of Proposition~\ref{prop:symmetric}, so the direction of the flow depends on the geometry. Under vanilla policy gradient, collapse is then stable on policies with one leading answer at every $c_{\rm b}$ whenever the fraction of samples without an answer, the invalid fraction, is positive and held fixed. For arm C at $c_{\rm b}=0.6$ and $n=\infty$, the boundary of the collapsed region lies at $p_{\rm mode}=0.99$, $0.96$ and $0.91$ for invalid fractions of $0.01$, $0.03$ and $0.05$. From an invalid fraction of $0.057$ on, no interior equilibrium remains. Under NPG at $c_{\rm b}=0.6$, collapse stays unstable for invalid fractions below $0.357$. The reward $0$ also drives the invalid fraction down, so a broad start with an invalid fraction of up to $0.2$ reaches the interior equilibrium in both flows. In the vanilla flow, a collapsed start inside the region is still released, but from $p_{\rm mode}=0.99$ and an invalid fraction of $0.01$, it takes about twice as long as with these samples left out.

If the advantages are computed from the samples with an answer only, the threshold holds group by group with $G_{\rm v}$ in place of $G$. With the advantage $0$ for samples without an answer, both geometries have the same equilibria at a fixed invalid fraction, and from $0.246$ on, collapse is again stable at $c_{\rm b}=0.6$. Under vanilla policy gradient, the invalid fraction grows as the leading answer is pushed down. A fixed advantage of $-0.1$ for these samples drives the invalid fraction down slowly, to about $0.005$ at time $2000$ of the vanilla flow. By then, every start we tried, with $p_{\rm mode}$ from $0.6$ to $0.97$ and invalid fractions from $0.03$ to $0.10$, had reached the interior equilibrium, $0.717$ for $n=\infty$ and $0.861$ for $n=2$ at $c_{\rm b}=0.6$. Since the invalid fraction decays much more slowly than the policy collapses without the bonus, the test of Appendix~\ref{app:ttrl} counts only problems with at least 90 percent of samples with an answer.

\paragraph{Predicted states.}
We take $G=8$, a fixed invalid fraction from $0$ to $0.1$ and the advantage $0$ for samples without an answer. The interior equilibrium of arm C at $c_{\rm b}=0.6$ then has $p_{\rm mode}$ among samples with an answer between $0.856$ and $0.878$ for $n=2$, between $0.774$ and $0.835$ for $n=3$ and between $0.708$ and $0.808$ for $n=\infty$. With the advantage $-0.1$, the invalid fraction decays, and the equilibria lie near the lower ends. Arm R at $c_{\rm b}=2.3$ has $p_{\rm mode}$ between $0.702$ and $0.713$ for $n=2$, no leading branch for $n=3$ and a low leading branch near $0.22$ for $n=\infty$. At $c_{\rm b}=0$, in arm C at $c_{\rm b}=0.35$ and in arm R at $c_{\rm b}=0.6$, collapse is stable on these policies for every $n$. Arm R at $c_{\rm b}=0.6$ moves the leading answer at about 57 to 68 percent of the speed of $c_{\rm b}=0$ for $M$ from $0.7$ to $0.9$ and $n=3$ and $\infty$.

\section{Optimizers and other update rules}\label{app:updates}\label{app:networks}\label{sec:networks}

The main text uses SGD, the fixed-step update $z\leftarrow z+\eta V$, and Adam. Appendix~\ref{app:updates-def} collects our results for four further update rules.

\subsection{Other update rules}\label{app:updates-def}

These rules change how the logit velocity becomes a step.

\begin{itemize}[leftmargin=*,itemsep=1pt]
\item Normalization by the standard deviation $\sd_p(R)$ of the reward under the policy, as in GRPO \citep{shao2024deepseekmath}, turns the logit velocity into $V/\sd_p(R)$ (Theorem~\ref{thm:std} and Appendix~\ref{app:norm}).
\item The sign update moves every logit by $\eta\,\sign(V_i)$. Its flow
\begin{equation}
z_i'=\sign(V_i)
\end{equation}
is the small-step limit of Adam when its constant $\epsilon_{\rm A}$ is negligible, because its averages then track the current gradient.
\item Heavy-ball momentum replaces the step by an exponential average of past steps, a filter that does not move fixed points.
\item Clipping caps the logit velocity, $|V_i|\le C_V$, a tabular analogue of the clipped objective of PPO \citep{schulman2017ppo}, with the cap $\epsilon_{\rm clip}=C_V\eta$ per step.
\end{itemize}
The sign update, Adam and clipping bound the speed. The sign update, momentum and clipping do not move the zeros of the logit velocity. Normalization by the standard deviation moves no zero without exploration but rescales exploration relative to selection (Proposition~\ref{prop:norm}(b)). Table~\ref{tab:capped-adam} and Appendix~\ref{app:prereg-hyst} test the sign update and momentum.

\paragraph{Normalization by the standard deviation.}\label{app:std}

\begin{theorem}[Normalization by the standard deviation is a change of time]\label{thm:std}
Let $z_{\rm raw}(t_{\rm raw})$ solve $z'=\mathcal F(z)$, and let $\varsigma>0$ be continuous. The solution of $z'=\mathcal F(z)/\varsigma(z)$ is $z(t)=z_{\rm raw}(t_{\rm raw}(t))$, where
\begin{equation}
t(t_{\rm raw})=\int_0^{t_{\rm raw}}\varsigma(z_{\rm raw}(t'))\,dt' .
\end{equation}
The two flows have the same trajectories and limits. If the raw flow concentrates as $t_{\rm raw}\to\infty$, the normalized flow concentrates in finite learning time if and only if $\int_0^\infty\varsigma\,dt_{\rm raw}<\infty$.
\end{theorem}

\begin{proof}
With $dt_{\rm raw}/dt=1/\varsigma(z_{\rm raw}(t_{\rm raw}))$, one has $\frac{d}{dt}z_{\rm raw}(t_{\rm raw}(t))=\mathcal F/\varsigma$, and uniqueness of solutions identifies the two curves. The learning time needed to traverse the whole raw trajectory is $\int_0^\infty\varsigma\,dt_{\rm raw}$.
\end{proof}

Theorem~\ref{thm:std} concerns the flow without exploration. With exploration, the normalization divides only the selection term, so it shifts the balance between selection and exploration (Proposition~\ref{prop:norm}(b)). Near the uniform policy, $\sd_p(R)$ of the agreement reward vanishes, while the normalized advantage stays of order one. In the hysteresis test of a leak, no exploration up to $\varepsilon=0.5$ prevented collapse under this normalization (Appendix~\ref{app:prereg-hyst}). With groups of $G$ samples, $\varsigma_G$ of \eqref{eq:sigmaG} stays positive at the uniform policy, and the threshold of a term in the loss is finite and grows with $G$ (Section~\ref{sec:norm-reward}).

\section{Experimental details}\label{app:experiments}\label{app:prereg}

The predicted thresholds of the experiments in Section~\ref{sec:experiments} come from the theory without fitted parameters. A state counts as collapsed when its $\pmax$ exceeds $0.2$ in the testbed, $0.3$ for the agreement reward and $0.2$ for the majority reward in the language model, and $0.95$ under a bonus. These levels lie between the broad and the collapsed states, except at $w=1$ in the testbed, where the broad state itself exceeds $0.2$, and for $0.3$ in arm R of the agreement reward (Appendix~\ref{app:seeds}). Each run used a single GPU.

\begin{table}[!htbp]
\centering
\caption{Outcomes of the runs relative to the boundaries, with the new seeds of Appendix~\ref{app:seeds} in parentheses. Testbed thresholds are SGD and AdamW seed means, or ranges in the second row. Broad means that at most half of the states or questions collapsed, and the bonus rows of the letter test used collapsed starts for $G=16$ and $64$. The free-form row uses F95e at step 200.}
\label{tab:regimes}
\scriptsize
\begin{tabular}{@{}>{\raggedright\arraybackslash}p{0.29\textwidth}>{\raggedright\arraybackslash}p{0.16\textwidth}>{\raggedright\arraybackslash}p{0.27\textwidth}>{\raggedright\arraybackslash}p{0.20\textwidth}@{}}
\toprule
setting, seeds and boundaries & below the lower & inside the window & above the upper \\
\midrule
testbed, $w=3$, seeds~0,~1 (2~to~4), $[0.143,0.283]$ & collapsed at $0.14$ & prevented from $0.16$, $0.20$, released from $0.28$ & broad at $0.30$, $0.32$ \\
testbed, both normalized, seed~0 and partly~1 (2~to~4), $[0.131,0.276]$ estimated & collapsed at $0.10$, $0.12$ & prevented from $0.14$ to $0.18$, released at $0.26$ by AdamW in seeds 0, 1 and~3 & broad at $0.28$, $0.30$ \\
letter test, agreement, arm U, seeds~0 to~2 (3,~4), $[0.0875,0.2431]$, arm B, seed~0, $[0.069,0.219]$ & committed at $0.06$ & depended on start, seed and learning rate, ramps of seeds 0 to 3 released at or just below the fold & broad at $0.26$, $0.30$ \\
letter test, agreement, arm R, seeds~0 (3,~4), $[1.10,1.59]$ & committed at $0.22$ to $0.9$ & at $1.2$, broad start committed in 2 of 3 seeds and collapsed start released, both broad at $1.4$ & broad at $1.6$ to $2.4$ \\
letter test, majority ramps, arm U, seeds~0,~1 (2,~3), $[0.363,0.766]$ & committed at the ends & committed at $0.48$ to $0.60$, released at $0.70$ to $0.74$ & none committed \\
letter test, majority ramps, arm R, seeds~0,~1 (2,~3), $[0.806,1.587]$ & committed at the ends & loop closed in seed~0, open in seeds~1~to~3 & one brief excursion in seed 0 \\
letter test, bonus, arm C, $G=8$ to $64$, seeds~0,~1, both boundaries $c_{\rm b}^*$ & collapsed at $0.8c_{\rm b}^*$ & none & interior at $1.25c_{\rm b}^*$ from both starts \\
letter test, bonus, arm R, as above, $[0.780,1.946]$ for $G=64$ & collapsed at $0.8$ times the lower & interior at $1.25$ times the lower for $G\ge16$, collapsed starts stayed & collapsed starts released, broad start interior for $G=8$ \\
free-form TTRL, seeds~1 to~6, arm C, $0.514$, arm R, $[1.375,1.554]$ & C collapsed at $0.35$, $0.45$, R at $0.6$, and at $1.2$ in 2 of 3 seeds & R prevented at $1.45$ on the window set & C prevented at $0.6$, $0.8$, R at $1.8$, $2.3$, except two runs that lost the format \\
\bottomrule
\end{tabular}
\end{table}

\subsection{The ordered-action network testbed}\label{app:testbed}

In the threshold scan of Section~\ref{sec:testbed}, the threshold of the smoothing term was nearly independent of the learning rate in the loss and proportional to it as a separate SGD step.

\paragraph{Model and training.}
Thirty-two states with random inputs in $\R^{32}$ are mapped by a tanh network with two hidden layers of width 256 to $K=256$ logits. Behavior cloning toward Gaussian targets builds a broad start (mean $\pmax=0.017$) and a peaked start (mean $\pmax=0.50$). The collapsed start, shared by every optimizer, comes from training the broad start with AdamW without exploration until the mean $\pmax$ exceeds $0.95$, about 50 steps. Each RL step samples $G=64$ actions per state and rewards each by the fraction of the other 63 samples in its bin,
\begin{equation}\label{eq:F-loo}
r_j=\frac1{G-1}\sum_{k\ne j}[a_k=a_j].
\end{equation}
Here $a_j$ is the bin of sample $j$, and $[a_k=a_j]$ equals $1$ if the two samples share a bin and $0$ otherwise. The step then subtracts the per-state mean and runs four epochs of the clipped surrogate with clipping $0.2$. The exploration term is $c\cdot\mathrm{CE}(D_wp_{\rm old},p)$, with $D_w$ mirrored at reflecting ends or wrapped around on a ring, and $w=1.5$ in the main and fine grids. The optimizers are AdamW at $3\times10^{-4}$ with $\beta_2\in\{0.999,0.99\}$ and no weight decay, and SGD at $0.3$, for 6000 steps with evaluations every 100 steps. The collapse threshold of a grid is the smallest $c$ at which the fraction of states with $\pmax>0.2$, averaged over seeds, is below one half.

\paragraph{Main and fine grids.}
The main grid has 288 runs over optimizers, boundary conditions, starts, eight values of $c$ and two seeds, and four runs with noise on executed actions. The fine grid $c\in\{0.3,0.4,0.5,0.6,0.7,0.85,1\}$ has three seeds and 168 runs.

SGD with $c\ge10$ collapsed again as its step became unstable, and these runs are excluded. The prevention and release thresholds were both $c=1$ in every condition of the main grid and in $(0.3,0.4]$ on the fine grid, a spacing too coarse to resolve the window $[0.287,0.342]$ of the objective at $w=1.5$. Noise on executed actions is not exploration, since the reward depends on the policy only through its convolution with the noise, so a spike stays optimal. With noise of 4 and 16 bins, 100 and 72 percent of the peaked states stayed peaked.

\paragraph{Threshold scan.}
Each chain checks that the policy collapses at $c=0.005$ and does not collapse at $10$ for AdamW and $2$ for SGD. It then bisects $\log c$ eight times to a relative width of $2.4$ to $3$ percent, with a number of steps inversely proportional to the learning rate, and $\hat c$ is the geometric mean over two seeds. In the separate-step condition, each epoch takes the optimizer step on the policy loss and then a plain SGD step with learning rate $0.3$ on the smoothing loss.

\paragraph{Tabular predictor.}
With $G=64$, the expected logit gradient of the testbed is that of \eqref{eq:inloss} with selection multiplied by $\frac{62}{64}$. Every threshold of the lattice equation is therefore multiplied by $\frac{31}{32}$ in the units of the testbed. The finite-horizon prediction iterates this equation for $2\times10^4$ time units from a Gaussian of width $0.1$ on a ring of 256 bins and bisects with the criterion $\pmax>0.2$. The stationary prediction runs it to stationarity (Appendix~\ref{app:numerics-inloss}). Since the network has the equilibria of this equation (Lemma~\ref{lem:fixedpoints}), agreement between the network and these predictions tests whether the transition is a fold (Table~\ref{tab:c1b-full}).

\begin{table}[!htbp]
\centering\scriptsize
\caption{Collapse threshold $\hat c$ of the smoothing term in the network testbed. The tabular rows solve the lattice equation \eqref{eq:inloss} of the objective with no free parameter, in the units of the testbed, where the leave-one-out reward of 64 samples multiplies selection by $62/64$. The fold is the certified threshold of Theorem~\ref{thm:inloss-cap} (Table~\ref{tab:inloss-cap}) in these units. The bisection rows use the criterion $\pmax>0.2$ of the network, either to stationarity or for $2\times10^4$ time units. At $w=1$, the broad state itself meets this criterion up to $c=0.495$.}
\label{tab:c1b-full}
\begin{adjustbox}{max width=\linewidth}
\begin{tabular}{lllll}
\toprule
exploration & optimizer and learning rates & $w=1$ & $w=1.5$ & $w=3$ \\
\midrule
in the loss & AdamW, $10^{-4}$, $3\times10^{-4}$, $10^{-3}$ & $0.491$, $0.499$, $0.514$ & $0.315$, $0.319$, $0.315$ & $0.169$, $0.184$, $0.176$ \\
in the loss & SGD, $0.1$, $0.3$, $1$ & $0.491$, $0.497$, failed & $0.300$, $0.304$, $0.297$ & $0.151$, $0.158$, $0.158$ \\
fold of the lattice equation & none & $0.4332$ & $0.2871$ & $0.1431$ \\
tabular bisection, stationary & none & $0.495$ & $0.2871$ & $0.1434$ \\
tabular bisection, $2\times10^4$ time units & none & $0.476$ & $0.273$ & $0.135$ \\
\midrule
separate SGD step & SGD, $0.1$, $0.3$, $1$ & & $0.095$, $0.287$, $0.892$ & \\
separate SGD step & AdamW, $10^{-4}$, $3\times10^{-4}$, $10^{-3}$ & & $0.073$, $0.324$, $1.03$ & \\
\bottomrule
\end{tabular}
\end{adjustbox}
\end{table}

\paragraph{Results of the scan.}
Of 48 chains, SGD at learning rate 1 with $w=1$ in the loss gave no valid bracket, and with the smoothing outside AdamW, only one seed per learning rate gave one. With the smoothing term in the loss, $\hat c$ varies by a factor of $1.01$ to $1.09$ over a tenfold range of learning rates, threefold for SGD with $w=1$. The product $\hat cw$ is nearly constant, $0.45$ to $0.55$, as $c^*w$ is for the folds of Table~\ref{tab:inloss-cap}, while $\hat cw^2$ varies by a factor of $2.8$ to $3.3$. As a separate SGD step, $\hat c/\mathrm{lr}$ is constant to within 7 percent, and outside AdamW, $\hat c$ changes by a factor of 14. At the two ends of the final bracket, the mean $\pmax$ is $0.21$ to $0.22$ and $0.19$ to $0.20$ in the loss for $w=1$. For $w\in\{1.5,3\}$, it is $0.32$ to $0.84$ and $0.08$ to $0.38$. The tabular prediction gives a similar jump, $0.733$ and $0.868$ at $0.95$ times and $0.170$ and $0.084$ at $1.05$ times its threshold.

Averaged over learning rates, AdamW lies 2, 5 and 13 percent above SGD for $w=1$, $1.5$ and $3$, an excess that comes from escape from the broad state (Appendix~\ref{app:testbed-c1x}). The stationary lattice equation predicts $0.495$, $0.2871$ and $0.1434$, the folds of the objective for $w\ge1.5$, while for $w=1$, the broad state itself crosses the criterion above a fold at $0.4332$. This prediction lies between 9 percent below and 1 percent above the SGD values and between 22 percent below and 1 percent above the AdamW values. Stopped at $2\times10^4$ time units, it lies 3 to 27 percent below the network.

\subsubsection{Hysteresis and escape from the broad state}\label{app:testbed-c1x}

The window of the objective appears in the network at $w=3$ (Figure~\ref{fig:regimes}a), since both starts kept their state for $c$ from $0.20$ to $0.26$, and AdamW escaped from the broad equilibrium inside it.

\paragraph{Design.}
Runs use a ring, 32 states, 64 samples, PPO and 6000 steps unless stated, with AdamW at $3\times10^{-4}$ and SGD at $0.3$. The hysteresis grid uses $c\in\{0.14,0.16,\dots,0.32\}$ at $w=3$ and $c\in\{0.28,0.29,\dots,0.36,0.38\}$ at $w=1.5$, with both starts, both optimizers and two seeds. The escape test starts every state at the broad equilibrium at $w=3$, with one seed, and repeats the test on a free logit table. A threshold is the smallest $c$ of the grid at and above which the mean fraction $f(c)$ of collapsed states stays at most one half,
\begin{equation}\label{eq:F-thr}
\min\{c\mid f(c')\le\tfrac12\text{ for every }c'\ge c\text{ of the grid}\}.
\end{equation}
\paragraph{Hysteresis.}
Table~\ref{tab:c1x-hyst} gives the grid at $w=3$. For $c$ from $0.20$ to $0.26$, broad starts stayed broad and collapsed starts stayed collapsed under both optimizers. The prevention thresholds are $0.16$ for SGD and $0.20$ for AdamW, and the release threshold is $0.28$ for both, against the folds $0.143$ and $0.283$, with ratios $1.75$ and $1.40$ against $1.98$. The final states lie on the two branches of the lattice equation. At $c=0.26$, AdamW released 28 and 22 percent of the collapsed states of the two seeds, at times between 100 and 5600 steps. SGD lost up to 56 percent of the collapsed states of one seed within the first 1000 steps at every $c$ of the window. These losses are likely an overshoot of the first steps from the sharp collapsed start. SGD lost none of the other seed for $c$ from $0.18$ to $0.24$. Three new seeds at $w=3$ gave the same seed means of the thresholds (Appendix~\ref{app:seeds}).

At $w=1.5$, the window $[0.287,0.342]$ is not wider than these shifts, and the second seed alone reverses the order of the optimizers, so the grid does not resolve the window. Every state collapsed at $c=0.28$, below the lower fold, and from $c=0.35$ on, above the upper fold, at most 2 percent did.

\begin{table}[!htbp]
\centering\scriptsize
\caption{Hysteresis grid at $w=3$. Mean over two seeds of the final fraction of collapsed states, with the mean $\pmax$ in parentheses. The last column gives $\pmax$ of the broad and spike equilibria of the lattice equation in the units of the testbed.}
\label{tab:c1x-hyst}
\begin{adjustbox}{max width=\linewidth}
\begin{tabular}{llllll}
\toprule
$c$ & AdamW, broad start & AdamW, collapsed start & SGD, broad start & SGD, collapsed start & lattice equation \\
\midrule
$0.14$ & $1.00$ ($0.849$) & $1.00$ ($0.855$) & $1.00$ ($0.843$) & $1.00$ ($0.831$) & none, $0.853$ \\
$0.16$ & $1.00$ ($0.822$) & $1.00$ ($0.827$) & $0.22$ ($0.237$) & $0.84$ ($0.694$) & $0.070$, $0.827$ \\
$0.18$ & $0.62$ ($0.522$) & $1.00$ ($0.795$) & $0.00$ ($0.062$) & $0.91$ ($0.722$) & $0.058$, $0.798$ \\
$0.20$ & $0.05$ ($0.088$) & $1.00$ ($0.763$) & $0.00$ ($0.053$) & $0.83$ ($0.626$) & $0.050$, $0.767$ \\
$0.22$ & $0.00$ ($0.050$) & $1.00$ ($0.729$) & $0.00$ ($0.046$) & $0.83$ ($0.599$) & $0.045$, $0.731$ \\
$0.24$ & $0.00$ ($0.043$) & $1.00$ ($0.685$) & $0.00$ ($0.041$) & $0.72$ ($0.499$) & $0.040$, $0.689$ \\
$0.26$ & $0.00$ ($0.040$) & $0.75$ ($0.487$) & $0.00$ ($0.038$) & $0.62$ ($0.402$) & $0.037$, $0.636$ \\
$0.28$ & $0.00$ ($0.036$) & $0.00$ ($0.036$) & $0.00$ ($0.034$) & $0.00$ ($0.035$) & $0.034$, $0.541$ \\
$0.30$ & $0.00$ ($0.033$) & $0.00$ ($0.033$) & $0.00$ ($0.032$) & $0.00$ ($0.032$) & $0.031$, none \\
$0.32$ & $0.00$ ($0.031$) & $0.00$ ($0.031$) & $0.00$ ($0.029$) & $0.00$ ($0.030$) & $0.029$, none \\
\bottomrule
\end{tabular}
\end{adjustbox}
\end{table}

\paragraph{Escape from the broad state.}
Table~\ref{tab:c1x-escape} gives the escape test. AdamW leaves the broad equilibrium inside the window, while SGD at $0.3$ stays from $c=0.16$ on. On a free logit table, Adam escapes in the same range, with a boundary that rises with its learning rate. With 256 samples instead of 64, the boundary of the network moves down from about $0.185$ to $0.172$. SGD also escapes when its steps are large, collapsing 66 percent of the states of the table at $c=0.16$ with learning rate $64$ and none at $8$. At $256$, its steps are too large for the flow to describe, since spikes no longer settle on the branch of the lattice equation. No condition escaped at $c=0.25$. Escaped states settle within $0.025$ of the spike branch in the mean, except for SGD at $256$. %

\begin{table}[!htbp]
\centering\scriptsize
\caption{Escape from the broad equilibrium at $w=3$. Final fraction of collapsed states, one seed, with $c$ in the columns. The folds of the objective are $0.143$ and $0.283$ in the units of the testbed. Empty cells were not run.}
\label{tab:c1x-escape}
\begin{adjustbox}{max width=\linewidth}
\begin{tabular}{lllllllll}
\toprule
condition & $0.15$ & $0.16$ & $0.17$ & $0.18$ & $0.19$ & $0.20$ & $0.22$ & $0.25$ \\
\midrule
network, AdamW $3\times10^{-4}$, 64 samples & $1.00$ & $1.00$ & $1.00$ & $0.81$ & $0.25$ & $0.09$ & $0.00$ & $0.00$ \\
network, AdamW $3\times10^{-4}$, 256 samples & $1.00$ & $0.94$ & $0.59$ & $0.06$ & $0.00$ & $0.00$ & $0.00$ & $0.00$ \\
network, AdamW $10^{-4}$, 18000 steps & $1.00$ & $1.00$ & $0.56$ & $0.28$ & $0.03$ & $0.03$ & $0.00$ & $0.00$ \\
network, AdamW $10^{-3}$ & & & $1.00$ & & $0.25$ & & $0.00$ & \\
network, SGD $0.3$ & $1.00$ & $0.00$ & $0.00$ & $0.00$ & $0.00$ & $0.00$ & $0.00$ & $0.00$ \\
table, Adam $0.003$ & $0.44$ & $0.00$ & $0.00$ & $0.00$ & $0.00$ & $0.00$ & $0.00$ & $0.00$ \\
table, Adam $0.01$ & $1.00$ & $1.00$ & $0.38$ & $0.12$ & $0.00$ & $0.00$ & $0.00$ & $0.00$ \\
table, Adam $0.01$, 256 samples & & $0.78$ & & $0.00$ & & & & \\
table, Adam $0.03$ & $1.00$ & $1.00$ & $1.00$ & $1.00$ & $1.00$ & $0.84$ & $0.12$ & $0.00$ \\
table, SGD $8$ & $0.00$ & $0.00$ & $0.00$ & $0.00$ & $0.00$ & $0.00$ & $0.00$ & $0.00$ \\
table, SGD $64$ & & $0.66$ & & $0.00$ & & $0.00$ & & \\
table, SGD $256$ & & $1.00$ & & $0.34$ & & $0.00$ & & \\
\bottomrule
\end{tabular}
\end{adjustbox}
\end{table}

\paragraph{Why Adam escapes.}\looseness=-1
At the broad equilibrium at $c=0.22$, the sampled gradient is mostly noise, with a median ratio of $0.04$ between the absolute mean and the standard deviation of a parameter. SGD steps shrink with this noise, while Adam keeps steps of the size of its learning rate. In the grid runs that stayed broad at $c=0.22$, the standard deviation of $\pmax$ over time was $0.0033$ to $0.0046$ under AdamW and $0.0015$ to $0.0016$ under SGD. The largest $\pmax$ under AdamW was $0.124$, about half the saddle value $0.246$. Escapes come early, and under AdamW from the broad start with seed 0, a run of 24000 steps raised the final fraction only from $0.78$ to $0.84$ at $c=0.18$ and from $0.03$ to $0.06$ at $c=0.20$.

\subsection{Normalization by the group standard deviation}\label{app:testbed-norm}

Dividing only the advantages left both starts at nearly the same state, which removes the window (Section~\ref{sec:testbed}, Figure~\ref{fig:regimes}c), and dividing both terms restored the window.

\paragraph{Design.}
These runs use the setup of Appendix~\ref{app:testbed-c1x}, with seed 0 unless stated. With only the advantages normalized, the advantages of each state are divided by the unbiased standard deviation of its rewards plus $10^{-6}$, as in GRPO. With both terms normalized, the smoothing term is divided by the same quantity, and states whose rewards are all equal are dropped from both terms. SGD uses $0.02$ and $0.01$ in the two conditions, and a free logit table under Adam at $0.01$, started broad, tests the dependence on the group size.

\begin{table}[!htbp]
\centering\scriptsize
\caption{Network with only the advantages normalized, $w=3$. Final fraction of collapsed states ($\pmax>0.2$), with the mean $\pmax$ in parentheses. The last column is the normalized mean-field flow, from both starts.}
\label{tab:c1n-reward}
\begin{adjustbox}{max width=\linewidth}
\begin{tabular}{llllll}
\toprule
$c$ & AdamW, broad & AdamW, collapsed & SGD, broad & SGD, collapsed & mean field \\
\midrule
$1.0$ & $1.00$ ($0.594$) & $1.00$ ($0.592$) & $1.00$ ($0.594$) & $1.00$ ($0.593$) & $0.580$ \\
$1.5$ & $1.00$ ($0.422$) & $1.00$ ($0.415$) & $1.00$ ($0.419$) & $1.00$ ($0.418$) & $0.403$ \\
$2.0$ & $1.00$ ($0.303$) & $1.00$ ($0.304$) & $1.00$ ($0.311$) & $1.00$ ($0.306$) & $0.293$ \\
$2.5$ & $0.78$ ($0.213$) & $0.81$ ($0.229$) & $0.72$ ($0.213$) & $0.97$ ($0.228$) & $0.221$ \\
$3.0$ & $0.00$ ($0.143$) & $0.03$ ($0.140$) & $0.00$ ($0.127$) & $0.00$ ($0.135$) & $0.165$ \\
$4.0$ & $0.00$ ($0.096$) & $0.00$ ($0.089$) & $0.00$ ($0.089$) & $0.00$ ($0.090$) & $0.083$ \\
$6.0$ & $0.00$ ($0.063$) & $0.00$ ($0.064$) & $0.00$ ($0.062$) & $0.00$ ($0.063$) & $0.059$ \\
\bottomrule
\end{tabular}
\end{adjustbox}
\end{table}

\begin{table}[!htbp]
\centering\scriptsize
\caption{Network with both terms normalized, $w=3$, seed 0. Final fraction of collapsed states, with the mean $\pmax$ in parentheses. Empty cells were not run. A second seed under AdamW gave $0.62$ ($0.542$) for the broad start at $c=0.16$ and $0.28$ ($0.207$) for the collapsed start at $c=0.26$. The broad and spike equilibria of the lattice equation are in the last column of Table~\ref{tab:c1x-hyst}.}
\label{tab:c1n-both}
\begin{adjustbox}{max width=\linewidth}
\begin{tabular}{lllll}
\toprule
$c$ & AdamW, broad & AdamW, collapsed & SGD, broad & SGD, collapsed \\
\midrule
$0.10$ & $1.00$ ($0.891$) & $1.00$ ($0.898$) & & \\
$0.12$ & $1.00$ ($0.865$) & $1.00$ ($0.877$) & $1.00$ ($0.873$) & $1.00$ ($0.875$) \\
$0.14$ & $1.00$ ($0.848$) & $1.00$ ($0.851$) & $0.00$ ($0.086$) & \\
$0.16$ & $0.62$ ($0.535$) & $1.00$ ($0.824$) & $0.00$ ($0.066$) & $1.00$ ($0.824$) \\
$0.18$ & $0.06$ ($0.103$) & $1.00$ ($0.790$) & & \\
$0.20$ & $0.03$ ($0.075$) & $1.00$ ($0.758$) & $0.00$ ($0.049$) & $1.00$ ($0.765$) \\
$0.22$ & $0.00$ ($0.044$) & $1.00$ ($0.725$) & & \\
$0.24$ & $0.00$ ($0.040$) & $1.00$ ($0.680$) & $0.00$ ($0.039$) & $1.00$ ($0.688$) \\
$0.26$ & $0.00$ ($0.037$) & $0.50$ ($0.337$) & & $1.00$ ($0.632$) \\
$0.28$ & $0.00$ ($0.033$) & $0.00$ ($0.033$) & $0.00$ ($0.032$) & $0.00$ ($0.033$) \\
$0.30$ & $0.00$ ($0.031$) & $0.00$ ($0.031$) & & \\
\bottomrule
\end{tabular}
\end{adjustbox}
\end{table}

\paragraph{Results.}
With only the advantages normalized, the two starts end within $0.016$ of each other in mean $\pmax$ at every $c$ for both optimizers. The fraction of collapsed states passes one half between $c=2.5$ and $3.0$ (Table~\ref{tab:c1n-reward}). The mean $\pmax$ decreases smoothly and lies within $0.02$ of the mean-field flow except at $c=3.0$, where the network ends at $0.127$ to $0.143$ against $0.165$. At $w=\frac32$ under AdamW, the two starts agree within $0.005$ at every $c$ from $2$ to $6$ and lie within $0.012$ of the mean-field flow. Over the AdamW learning rates $10^{-4}$, $3\times10^{-4}$ and $10^{-3}$, with $18000$ steps at the smallest, the mean $\pmax$ from the broad start moved by at most $0.02$ at $c=1.5$, $2.5$ and $4.0$. On the logit table, larger groups end more concentrated, with mean $\pmax$ $0.134$, $0.305$ and $0.316$ at $c=2$ for groups of 16, 64 and 256. The fraction collapsed passes one half between $c=1.5$ and $2.0$ for $G=16$, between $2.5$ and $3.0$ for $G=64$ and near $3.0$ for $G=256$.

With both terms normalized, the window returns (Table~\ref{tab:c1n-both}), with prevention and release thresholds $0.18$ and $0.26$ under AdamW, averaged over the seeds run at each $c$, and $0.14$ and $0.28$ under SGD. The folds of the sampled correction are $0.131$ and $0.276$ (Appendix~\ref{app:norm-both}). The final states lie on the two branches of the lattice equation, for example $0.044$ against $0.045$ on the broad one and $0.725$ against $0.731$ on the spike at $c=0.22$. As without normalization, AdamW escapes from the broad state above the lower fold. Seeds 2 to 4 also showed the window (Appendix~\ref{app:seeds}).

\subsection{The language-model test}\label{app:llm-test}\label{app:llm}

In a language model with single-letter answers (Section~\ref{sec:llm}), ramps released within one evaluation of the fold, and inside the window, the outcome depended on the seed and the learning rate.

\paragraph{Model and data.}
The model is Qwen2.5-1.5B-Instruct \citep{qwen25report} with LoRA adapters \citep{hu2022lora} of rank 32 and alpha 64 on all linear layers except the output head. The questions come from MMLU-Pro \citep{wang2024mmlupro}, with ten options and prompts of at most 400 tokens that end right before the answer letter. They are drawn from the 2639 of 2665 questions on which the ten letters carry at least $0.8$ of the next-token mass. We select 16 in each of four strata of the largest letter probability, $[0.1,0.3)$, $[0.3,0.5)$, $[0.5,0.7)$ and $[0.7,1)$, and $p$ is the softmax of the ten letter logits there.

\paragraph{Training.}
Each step draws 32 questions and samples $G=16$ letters from $p$ for each, with the leave-one-out agreement reward \eqref{eq:F-loo}, now over letters, and advantages relative to the mean of the group. The loss is the PPO clipped surrogate with clipping $0.2$ over four epochs plus $c\cdot\mathrm{CE}(\bar p,p)$, with AdamW at $10^{-4}$ without weight decay and gradient clipping at $1$. Arm U normalizes nothing, and arm R divides the advantages of each question by the unbiased standard deviation $s$ of its rewards plus $10^{-6}$. Arm B also divides the exploration term by $s+10^{-6}$ and drops questions with $s=0$ from both terms. The two start adapters are built once without normalization and shared by all arms. The collapsed adapter is trained with $c=0$ to a mean $\pmax$ of $0.998$, and the broad adapter with $c=1$ to $0.106$, each within 10 steps. Each run takes 300 steps, with $p$ evaluated for all 64 questions every 20 steps.

\paragraph{Analysis and theory.}
A question counts as collapsed if its $\pmax$ exceeds $0.3$ at the last evaluation. The threshold of a grid is the smallest $c$ such that, at this $c$ and every larger $c$ of the grid, at most half of the questions are collapsed, as in \eqref{eq:F-thr}. It is measured from the broad start for prevention and from the collapsed start for release. The grid is $c\in\{0.06,0.10,\dots,0.30\}$ in steps of $0.04$ for arms U and B and $c\in\{0.22,0.6,0.9,1.2,1.4,1.6,1.8,2.0,2.4\}$ for arm R. A learning-rate check repeats arm U at $c\in\{0.14,0.18\}$ with $3\times10^{-5}$ and $3\times10^{-4}$. With $n=10$ and $G=16$, the mean-field stability threshold and release fold are $0.0875$ and $0.2431$ in arms U and B and $1.130$ and $1.698$ in arm R (Appendix~\ref{app:norm-unordered}). Exact sums over the counts of a group, which Table~\ref{tab:regimes} uses, give $1.1005$ and $1.5926$ in arm R and $0.0690$ and $0.2194$ in arm B (Table~\ref{tab:symmetric}). From a leading letter of initial mass $M(0)<\frac12$, prevention needs $\frac78\cdot\frac{10}9M(0)(1-M(0))$, for example $0.124$ for $M(0)=0.15$.

\paragraph{Results of the grid.}
Tables~\ref{tab:llm-UB} and \ref{tab:llm-R} give the final states. In arms U and B, the prevention and release thresholds were $0.14$ and $0.18$, and the grid did not resolve the sampled correction of arm B (Appendix~\ref{app:norm-unordered}). In arm R, they were $1.4$ and $1.2$. At $c=1.2$, the broad start ended at $0.414$, on the branch of one leading answer, while the collapsed start was released, and neither run had settled. In arm B, a question whose 16 samples agree has no update, because both terms are dropped. At $c=0.06$, where 75 and 82 percent of the groups of the last 50 steps had equal rewards, both starts ended near $0.99$, above the spike value $0.934$ of arm U.

\begin{table}[!htbp]
\centering\scriptsize
\caption{Grid of arms U and B, learning rate $10^{-4}$, seed 0. Final fraction of collapsed questions ($\pmax>0.3$), with the mean $\pmax$ in parentheses. The last column gives $\pmax$ of the stable spike of Proposition~\ref{prop:potts-uniform} with the factor $\frac78$.}
\label{tab:llm-UB}
\begin{adjustbox}{max width=\linewidth}
\begin{tabular}{llllll}
\toprule
$c$ & U, broad & U, collapsed & B, broad & B, collapsed & spike \\
\midrule
$0.06$ & $1.00$ ($0.934$) & $1.00$ ($0.940$) & $1.00$ ($0.988$) & $1.00$ ($0.991$) & $0.934$ \\
$0.10$ & $1.00$ ($0.888$) & $1.00$ ($0.882$) & $1.00$ ($0.906$) & $1.00$ ($0.880$) & $0.884$ \\
$0.14$ & $0.00$ ($0.144$) & $1.00$ ($0.786$) & $0.02$ ($0.139$) & $0.52$ ($0.485$) & $0.826$ \\
$0.18$ & $0.00$ ($0.113$) & $0.12$ ($0.199$) & $0.00$ ($0.120$) & $0.00$ ($0.107$) & $0.755$ \\
$0.22$ & $0.00$ ($0.111$) & $0.00$ ($0.110$) & $0.00$ ($0.111$) & $0.00$ ($0.107$) & $0.654$ \\
$0.26$ & $0.00$ ($0.109$) & $0.00$ ($0.108$) & $0.00$ ($0.109$) & $0.00$ ($0.105$) & none \\
$0.30$ & $0.00$ ($0.109$) & $0.00$ ($0.109$) & $0.00$ ($0.104$) & $0.00$ ($0.106$) & none \\
\bottomrule
\end{tabular}
\end{adjustbox}
\end{table}

\begin{table}[!htbp]
\centering\scriptsize
\caption{Grid of arm R, as in Table~\ref{tab:llm-UB}. The last row gives $\pmax$ of the stable mean-field branch of one leading answer. With exact sums, the release fold is $1.5926$ (Table~\ref{tab:symmetric}), so the branch ends between $c=1.4$ and $1.6$.}
\label{tab:llm-R}
\begin{adjustbox}{max width=\linewidth}
\begin{tabular}{llllllllll}
\toprule
start & $c=0.22$ & $0.6$ & $0.9$ & $1.2$ & $1.4$ & $1.6$ & $1.8$ & $2.0$ & $2.4$ \\
\midrule
broad & $1.00$ ($0.933$) & $1.00$ ($0.764$) & $1.00$ ($0.586$) & $1.00$ ($0.414$) & $0.00$ ($0.145$) & $0.00$ ($0.142$) & $0.00$ ($0.143$) & $0.00$ ($0.120$) & $0.00$ ($0.111$) \\
collapsed & $1.00$ ($0.936$) & $1.00$ ($0.755$) & $1.00$ ($0.597$) & $0.11$ ($0.214$) & $0.00$ ($0.149$) & $0.00$ ($0.138$) & $0.00$ ($0.139$) & $0.00$ ($0.137$) & $0.00$ ($0.131$) \\
branch & $0.949$ & $0.721$ & $0.547$ & $0.414$ & $0.342$ & $0.273$ & none & none & none \\
\bottomrule
\end{tabular}
\end{adjustbox}
\end{table}

\paragraph{Further seeds and learning rates.}
The broad start of arm U collapsed at $c=0.14$ in 6 of 7 runs over learning rates from $3 \times 10^{-5}$ to $3\times10^{-4}$ and stayed broad at $c=0.18$ in all 7 runs. The release threshold was at most $0.14$ at $3\times10^{-4}$, $0.18$ for seeds 0 to 2 at $10^{-4}$ and above $0.18$ for all three seeds at $3\times10^{-5}$. At $10^{-4}$ and $c=0.14$, the collapsed starts of seeds 1 and 2 stayed collapsed by the criterion, but with only 73 and 67 percent of the questions, and their mean $\pmax$ was $0.619$ and $0.604$, against $0.826$. At $3\times10^{-5}$ and $c=0.18$, all three seeds showed hysteresis, the collapsed starts ending at $0.68$ to $0.77$ against $0.755$. Seeds 3 and 4 at $10^{-4}$ stayed broad from the broad start at $c=0.14$, and their release thresholds were $0.18$ and $0.14$ (Appendix~\ref{app:seeds}).

\paragraph{Trajectories.}
In the broad adapter, 70 percent of the questions share their most likely letter, while the collapsed adapter covers nine of the ten letters. At $c=0.14$, the broad starts that collapsed kept a mean $\pmax$ below $0.2$ for 40 to 280 steps before rising to the spike branch, and at $3\times10^{-4}$ until the last evaluation. Ten broad starts of arm U with seeds 0 to 2 collapsed at fixed coefficients. Nine of them ended with the same most likely letter for every question, and one for 92 percent of the questions. Every collapsed start with seeds 0 to 2 that was released had a mean $\pmax$ below $0.4$ after 20 steps.

\paragraph{A ramped coefficient.}
The coefficient changed linearly over 900 steps, from $c=0.14$ to $0.30$ from the collapsed adapter and from $0.30$ to $0.06$ from the broad adapter. The runs use AdamW at $3\times10^{-5}$, seeds 0 to 3 and an evaluation every 20 steps.

\paragraph{Results of the ramps.}
From the collapsed adapter, the mean $\pmax$ of seeds 0 and 1 dropped to $0.815$ and $0.763$ at step 40 and then followed the spike branch. It stayed within $0.025$ of it from step 80 on while at least 95 percent of the questions were collapsed, up to $c=0.211$ for seed 0 and $0.225$ for seed 1. Release came at the fold $c=0.2431$ in seeds 0, 1 and 3 and one evaluation earlier, at $0.2396$, in seed 2 (Table~\ref{tab:llm-ramp-seeds}). From the broad adapter, the mean $\pmax$ of seeds 0 and 1 stayed between $0.105$ and $0.144$ down to $c=0.113$. The broad starts collapsed at $c=0.1027$ to $0.1187$, 17 to 36 percent above the lower boundary $0.0875$, with every question on one letter, and then followed the spike branch, ending at $0.938$ at $c=0.06$ in seeds 0 and 1 against $0.934$. No ramp was committed above the upper boundary at any evaluation.

\begin{table}[!htbp]
\centering\scriptsize
\caption{Agreement ramps of Figure~\ref{fig:regimes}b. Release is the first evaluation of the upward ramp with at most half of the questions collapsed, and collapse the first evaluation of the downward ramp with more than half.}
\label{tab:llm-ramp-seeds}
\begin{adjustbox}{max width=\linewidth}
\begin{tabular}{lllll}
\toprule
 & seed 0 & seed 1 & seed 2 & seed 3 \\
\midrule
release $c$ (fold $0.2431$) & $0.2431$ & $0.2431$ & $0.2396$ & $0.2431$ \\
collapse $c$ & $0.1080$ & $0.1080$ & $0.1187$ & $0.1027$ \\
loop, release minus collapse & $0.1351$ & $0.1351$ & $0.1209$ & $0.1404$ \\
collapse over the lower boundary $0.0875$ & $1.23$ & $1.23$ & $1.36$ & $1.17$ \\
\bottomrule
\end{tabular}
\end{adjustbox}
\end{table}

\subsection{Ramps with the majority reward}\label{app:majority-ramp}

These ramps test the regimes for the majority reward of TTRL (Section~\ref{sec:llm}). In arm U, release came near the fold in both seeds, and the release ratio of arms R and U lay near the value computed for this reward, not that for the agreement reward.

\paragraph{Design.}
These runs use the model, the 64 questions, the start adapters and the training of Appendix~\ref{app:llm-test} with the majority reward of TTRL. This reward gives $1$ to the samples of the most frequent letter of a group of 16 and $0$ to the others, with ties broken at random. The coefficient changes linearly over 900 steps, up from the collapsed adapter, from $0.3$ to $1.0$ in arm U and from $0.6$ to $2.0$ in arm R. It changes down from the broad adapter, from $1.0$ to $0.2$ and from $2.0$ to $0.4$. AdamW runs at $3\times10^{-5}$ with seeds 0 and 1, and $p$ is evaluated every 10 steps, every $0.0078$ to $0.0089$ in $c$ in arm U and $0.0156$ to $0.0178$ in arm R. A question counts as collapsed if its $\pmax$ exceeds $0.2$, since the folds lie near $M=0.29$. Release is the $c$ of the first evaluation of the last stretch of at least five consecutive evaluations, or of the final three or more, with at most half of the questions collapsed. Collapse is defined in the same way with more than half.

\paragraph{Checks.}
The implementation matches the exact enumeration of Appendix~\ref{app:majority-exact}.

\begin{table}[!htbp]
\centering\scriptsize
\caption{Transitions of the majority ramps. The release folds are $0.766$ in arm U and $1.587$ in arm R, and the lower boundaries $0.363$ and $0.806$. The release of arm R over that of arm U is $1.89$ in seed 0 and $2.13$ in seed 1, against $2.07$ computed for the majority reward and $6.55$ for the agreement reward.}
\label{tab:maj-ramps}
\begin{tabular}{lllll}
\toprule
arm & seed & release $c$ & collapse $c$ & loop, release minus collapse \\
\midrule
U & 0 & $0.7356$ & $0.5289$ & $0.207$ \\
U & 1 & $0.7044$ & $0.5467$ & $0.158$ \\
R & 0 & $1.3933$ & $1.3956$ & $-0.002$ \\
R & 1 & $1.5022$ & $1.1111$ & $0.391$ \\
\bottomrule
\end{tabular}
\end{table}

\paragraph{Results.}
Table~\ref{tab:maj-ramps} gives the transitions. By the criterion $\pmax>0.3$, only the release of arm R with seed 1, $1.4711$, and the collapses of arm R, $1.3778$ and $1.0933$, change. In the downward ramp of arm R with seed 0, more than half of the questions were collapsed at $c=1.6267$, above the fold $1.587$, for fewer than five evaluations. The persistent collapse came at $1.3956$. The loop, release minus collapse, was $0.207$ and $0.158$ wide in arm U, 51 and 39 percent of the window $0.403$ of the exact flow. In arm R, it was $-0.002$ and $0.391$ wide, the latter 50 percent of its window $0.780$, while the agreement ramps of seeds 0 and 1 covered 87 percent of theirs. Up to release, the mean $\pmax$ followed the leader branch within $0.027$ at every compared point except for arm U with seed 1 at $c=0.697$, just before its release at $0.7044$. At the end of the downward ramps, it was $0.829$ and $0.800$ in arm U against $0.8129$ on the branch at $c=0.2$. In arm R, it was $0.863$ and $0.846$ against $0.8598$ at $c=0.4$. Every question had the same most likely letter, and the accuracy was $0.12$.

\paragraph{Collapse of the downward ramps.}
Except for arm R with seed 0, the downward ramps collapsed when the unstable equilibrium on policies with one leading answer lay $0.039$ to $0.050$ above the uniform mass $0.1$. For the agreement ramp, this distance was $0.027$. This is consistent with noise crossing a shrinking barrier, but one distance does not fit both rewards.

\subsection{The answer-level bonus in the letter test}\label{app:letter-bonus}

This test asks whether the thresholds of the answer-level bonus of Appendix~\ref{app:bonus} hold in a language model across group sizes, both before the normalization (arm C) and after it (arm R). Sections~\ref{sec:statedep} and \ref{sec:llm} use its results. In both seeds and for every group size, broad starts collapsed at $0.8$ times the predicted threshold of each arm and did not collapse at $1.18$ to $1.25$ times it. After the normalization, a collapsed start stayed collapsed at a coefficient at which a broad start did not.

\paragraph{Design.}
\looseness=-1
The model, the 64 questions, the broad and collapsed adapters and the training settings are those of Appendix~\ref{app:llm-test}. Training uses AdamW at $3\times10^{-5}$ without weight decay for 900 steps, with an evaluation every 20 steps and the majority reward, ties broken at random. Each step draws $G$ letters for each of 32 questions, with $G\in\{8,16,32,64\}$. The smoothing term is replaced by the bonus $b_j=-\log(N_{a_j}/G)$ of Section~\ref{sec:statedep}. We write bars for group means, $\sd$ for the population standard deviation of a group and $\hat{\mathcal A}_j$ for the advantage of sample $j$. Arm C adds the bonus to the reward before the division,
\begin{equation}\label{eq:F-armC}
\hat{\mathcal A}_j=\frac{\tilde R_j-\overline{\tilde R}}{\sd(\tilde R)},\qquad \tilde R_j=R_j+c_{\rm b}\,b_j ,
\end{equation}
and arm R adds it after the division of the reward advantages,
\begin{equation}\label{eq:F-armR}
\hat{\mathcal A}_j=\frac{R_j-\bar R}{\sd(R)}+c_{\rm b}\,(b_j-\bar b) ,
\end{equation}
with a constant of $10^{-6}$ added to each denominator. These advantages equal those of the code of the TTRL test to $5\times10^{-13}$, and the leading velocity of this code matched exact enumeration in ten checks to within four standard errors.

\paragraph{Coefficients.}
The predicted threshold for a broad start is $c_{\rm b}^*=1/\log(G-1)$ in arm C and the end of the branch of interior equilibria in arm R (Table~\ref{tab:bonus-window}). Broad starts ran at $0.8$ and $1.25$ times this threshold. The upper coefficient of arm R with $G=8$ is $1.709$, $1.18$ times its threshold. At $1.25$ times its threshold, this coefficient would lie close to the stability threshold $1.858$ of the uniform policy, near which a broad start moves away from the uniform policy only slowly. Collapsed starts ran with $G=16$ and $64$ at four coefficients. In arm R, these are the upper coefficient of the broad starts, inside the window of Observation~\ref{obs:bonus-window}, and $1.25$ times the threshold near collapse. In arm C, they are the two coefficients of the broad starts. Table~\ref{tab:letter-bonus-thr} lists the coefficients.

\paragraph{Criteria.}
A run is collapsed at step 900 if more than half of the questions have $\pmax>0.95$. It is interior if it is not collapsed and its median $\pmax$ is at least $0.25$ and settled, which means that the mean of the medians at steps 820 to 900 differs from that at steps 720 to 800 by at most $0.05$. A threshold lies between the two coefficients of a group size if, in both seeds, the lower coefficient collapsed and the upper one ended interior. For collapsed starts, we measured, over the questions whose $\pmax$ at step 0 exceeded the unstable crossing $M_{\rm u}$ of the tabular flow, the mean drift
\begin{equation}\label{eq:F-drift}
\mathrm{drift}=\Big\langle \overline{\log(1-\pmax)}_{\,820\ldots900}-\log(1-\pmax)\big|_{0}\Big\rangle ,
\end{equation}
with $M_{\rm u}=0.952$ for $G=16$ and $0.919$ for $G=64$. The overline averages the five evaluations from step 820 to 900, and the brackets average over these questions.

\paragraph{Cost.}
All 48 runs finished, at about $1.16$ GPU-hours per run with two runs on each GPU.

\paragraph{Thresholds.}
Every run at the lower coefficient collapsed, with all questions above $\pmax=0.95$, and no run at the upper coefficient did (Table~\ref{tab:letter-bonus-thr}). Each threshold of both arms therefore lies between $0.8$ and $1.25$ times its prediction for every group size, and the ranges of arm C for $G=8$ and $64$ do not overlap, so the test also resolves the dependence on $G$. For $G\ge16$, arm R ended interior at its upper coefficient although this coefficient is only $0.88$, $0.65$ and $0.50$ times the threshold near collapse. Keeping a broad start from collapsing thus does not need the factor $G/\sqrt{G-1}$.

\begin{table}[!htbp]
\centering\scriptsize
\caption{Coefficients and outcomes of the bonus in the letter test. A threshold lies between the two coefficients if both seeds collapsed at the lower one and ended interior at the upper one. The ratio row divides the bracket of arm R by that of arm C, and the last row divides the upper coefficient of arm R by its threshold near collapse. Collapsed starts used $G=16$ and $64$ (Table~\ref{tab:letter-bonus-a2}).}
\label{tab:letter-bonus-thr}
\begin{tabular}{lllll}
\toprule
 & $G=8$ & $G=16$ & $G=32$ & $G=64$ \\
\midrule
arm C, threshold $c_{\rm b}^*$ & $0.514$ & $0.369$ & $0.291$ & $0.241$ \\
arm C, lower and upper coefficient & $0.411$, $0.642$ & $0.295$, $0.462$ & $0.233$, $0.364$ & $0.193$, $0.302$ \\
\midrule
arm R, threshold, end of branch & $1.454$ & $1.074$ & $0.876$ & $0.780$ \\
arm R, lower and upper coefficient & $1.163$, $1.709$ & $0.859$, $1.342$ & $0.701$, $1.096$ & $0.624$, $0.974$ \\
ratio of the brackets of arms R and C & $1.81$ to $4.16$ & $1.86$ to $4.55$ & $1.93$ to $4.70$ & $2.07$ to $5.05$ \\
arm R, threshold near collapse & $1.554$ & $1.526$ & $1.674$ & $1.946$ \\
arm R, upper over near collapse & $1.10$ & $0.88$ & $0.65$ & $0.50$ \\
\bottomrule
\end{tabular}
\end{table}

\paragraph{Interior states.}
At the upper coefficient, the runs settled by step 300, and the median $\pmax$ then varied between evaluations with a standard deviation of $0.01$ to $0.04$. In seven of the eight cells and in both seeds, its average over the evaluations at steps 300, 500, 700, 800 and 900 lay within $0.018$ of the predicted interior equilibrium (Table~\ref{tab:letter-bonus-m}). At step 900 alone, the same seven cells lay within $0.10$ of it. In these runs, all 64 questions shared the same leading letter, so each run is one sample, and each cell has two.

The exception is arm R with $G=8$. From step 300 on, its median $\pmax$ stayed between $0.31$ and $0.37$ instead of the predicted $0.452$. Its questions were spread over different letters, with $28$ and $17$ percent of them on the most common letter. Its coefficient $1.709$ lies above the threshold near collapse and at $0.92$ times the stability threshold of the uniform policy, where the pull away from the uniform policy is weak.

\begin{table}[!htbp]
\centering\scriptsize
\caption{Median $\pmax$ over questions at the upper coefficient, against the predicted interior equilibrium $M$ of the tabular flow. The time average is over the evaluations at steps 300, 500, 700, 800 and 900.}
\label{tab:letter-bonus-m}
\begin{adjustbox}{max width=\linewidth}
\begin{tabular}{lllll}
\toprule
cell & predicted $M$ & step 900, seeds 0 and 1 & time average, seeds 0 and 1 & time average minus $M$ \\
\midrule
arm C, $G=8$ & $0.587$ & $0.593$, $0.575$ & $0.569$, $0.579$ & $-0.018$, $-0.008$ \\
arm R, $G=8$ & $0.452$ & $0.344$, $0.315$ & $0.331$, $0.345$ & $-0.121$, $-0.107$ \\
arm C, $G=16$ & $0.686$ & $0.637$, $0.735$ & $0.683$, $0.701$ & $-0.003$, $+0.015$ \\
arm R, $G=16$ & $0.469$ & $0.429$, $0.465$ & $0.483$, $0.479$ & $+0.014$, $+0.010$ \\
arm C, $G=32$ & $0.760$ & $0.735$, $0.756$ & $0.765$, $0.756$ & $+0.005$, $-0.004$ \\
arm R, $G=32$ & $0.487$ & $0.461$, $0.469$ & $0.487$, $0.475$ & $0.000$, $-0.012$ \\
arm C, $G=64$ & $0.823$ & $0.800$, $0.792$ & $0.811$, $0.813$ & $-0.012$, $-0.010$ \\
arm R, $G=64$ & $0.503$ & $0.501$, $0.506$ & $0.497$, $0.503$ & $-0.006$, $0.000$ \\
\bottomrule
\end{tabular}
\end{adjustbox}
\end{table}

\paragraph{Collapsed starts.}
All sixteen runs from the collapsed adapter moved in the direction that the local threshold of their arm predicts (Table~\ref{tab:letter-bonus-a2}). Below the threshold, in arm R inside the window and in arm C at its lower coefficient, at least 88 percent of the questions stayed collapsed at every evaluation, and the drift was negative. Above it, every run was released within 20 to 240 steps. At the coefficient at which its broad start ended interior, arm R stayed collapsed. At its upper coefficient, arm C reached the interior state of its broad start, $0.65$ and $0.81$ against $0.69$ and $0.80$. With $G=64$, arm R above the threshold near collapse ended at $0.223$ and $0.204$, near the predicted interior equilibrium $0.221$. Divided by $c_{\rm b}^*$, the release threshold of arm R thus lies in $(3.63,5.16]$ for $G=16$ and in $(4.04,10.08]$ for $G=64$. Both brackets contain the factor $G/\sqrt{G-1}$, $4.13$ and $8.06$, but also every constant factor between $4.04$ and $5.16$, so the test does not resolve how this threshold depends on $G$.

\begin{table}[!htbp]
\centering\scriptsize
\caption{Collapsed starts with the bonus. The window coefficient of arm R is the upper coefficient of its broad starts, and its release coefficient $1.25$ times the threshold near collapse. Outcomes give class and median $\pmax$ at step 900. Release is the first evaluation with at most half of the questions collapsed, with the drift of \eqref{eq:F-drift} over 63 questions for $G=16$ and 64 for $G=64$. Cells give seeds 0 and 1.}
\label{tab:letter-bonus-a2}
\begin{adjustbox}{max width=\linewidth}
\begin{tabular}{lllllll}
\toprule
$G$ & run & $c_{\rm b}$ & prediction & outcome & release step & drift \\
\midrule
$16$ & arm R, window & $1.342$ & stays collapsed & collapsed, collapsed & none, none & $-1.40$, $-7.57$ \\
$16$ & arm R, release & $1.907$ & released to $\bar p$ & broad $0.194$, broad $0.199$ & $100$, $40$ & $+16.67$, $+16.66$ \\
$16$ & arm C, upper & $0.462$ & released to $0.686$ & interior $0.656$, $0.648$ & $60$, $240$ & $+15.75$, $+15.81$ \\
$16$ & arm C, lower & $0.295$ & stays collapsed & collapsed, collapsed & none, none & $-17.49$, $-7.29$ \\
$64$ & arm R, window & $0.974$ & stays collapsed & collapsed, collapsed & none, none & $-14.48$, $-14.48$ \\
$64$ & arm R, release & $2.433$ & released to $0.221$ & broad $0.223$, broad $0.204$ & $40$, $20$ & $+16.45$, $+16.48$ \\
$64$ & arm C, upper & $0.302$ & released to $0.823$ & interior $0.810$, $0.805$ & $80$, $60$ & $+14.98$, $+14.97$ \\
$64$ & arm C, lower & $0.193$ & stays collapsed & collapsed, collapsed & none, none & $-7.53$, $-19.68$ \\
\bottomrule
\end{tabular}
\end{adjustbox}
\end{table}

\paragraph{Caveats.}
The collapsed adapter was deep, with a median $1-\pmax$ of about $2\times10^{-8}$. Most questions therefore never sampled a second letter, so their advantages were zero and they moved only through the shared low-rank parameters. In the eight runs that stayed collapsed, 14 to 62 of the 64 questions also switched their leading letter between evaluations. The test thus concerns the set of collapsed policies, not the stability of each vertex. With $G=16$, arm R at its release coefficient $1.907$, above the largest coefficient $1.805$ with an interior equilibrium, ended at $0.19$ and $0.20$ instead of the uniform policy. A slow passage near the vanished fold at $M\approx0.22$ may explain this.

\subsection{Free-form test-time reinforcement learning}\label{app:ttrl}

With an answer-level bonus at $c_{\rm b}=0.6$ (Section~\ref{sec:ttrl}), arm C, which adds it before the normalization, did not collapse, and arm R, which adds it after, collapsed in every seed.

\paragraph{Model, data and training.}
The model is Qwen2.5-Math-1.5B \citep{qwen25math} with LoRA adapters of rank 32 and alpha 64. It is trained on 128 problems of MATH-500 \citep{hendrycks2021math,lightman2024verify} on which about half of the initial majority answers are wrong. Each step samples $G=8$ solutions at temperature 1 with at most 3072 tokens for each of 16 problems. The answer of a sample is its last boxed expression, and answers are grouped into classes of equivalent answers by a symbolic checker. Training uses the token-level clipped surrogate with clipping $0.2$ in four minibatches, AdamW at $10^{-4}$ without weight decay, gradient clipping at $1$ and no KL term, for 200 steps. The loss of a minibatch is the mean over its tokens, as in DAPO, so long samples weigh more. Each problem enters training about 25 times in 200 steps. Every 20 steps, 32 samples of each problem are evaluated at temperature 1 and top-p 1 with at most 3072 tokens, as in training, while the original TTRL evaluates at temperature $0.6$ and top-p $0.95$. After each run, every class is compared with the eight most frequent classes of its problem and merged if equivalent, which can only raise $p_{\rm mode}$. Merging changed no conclusion of this appendix.

\paragraph{Relation to the original TTRL.}
The original TTRL estimates the label by a vote over 64 samples and trains with GRPO on 32 of them \citep{zuo2025ttrl}. Its scripts for Qwen2.5-Math sample at temperature 1, as we do, but train all weights at the learning rate $5\times10^{-7}$. They also add the KL loss of GRPO with the coefficient $0.001$, outside the division by the standard deviation. Here, the vote and the update use the same 8 samples, and LoRA training has no KL term. Near collapse, a larger vote set elects the same leader, so the thresholds of a bonus counted within the training group depend only on its size.

This KL loss averages $\rho-\log\rho-1$ over the sampled tokens, where $\rho$ is the ratio of the reference probability to the current one. For one softmax over answers and samples of the current policy, its expected gradient in the logits is $\mathbb E_{a\sim p}[(1-p_{{\rm ref},a}/p_a)(e_a-p)]=p-p_{\rm ref}$. Descent thus pulls the policy toward $p_{\rm ref}$ as our smoothing on unordered answers pulls it toward the uniform policy. The coefficient $0.001$ is more than 300 times smaller than every lower boundary of Table~\ref{tab:symmetric} for the majority reward. The theory gives no reason to expect this KL loss to prevent collapse, in line with the collapse that \citet{wu2025spine} report for TTRL.

\paragraph{Samples without an answer.}
Rewards, bonuses, means and standard deviations use only the samples with an answer, and samples without one receive the fixed advantage $-0.1$ (Appendix~\ref{app:bonus}). The base model was first trained with a format reward, $1$ for a sample with an answer, in rounds of 30 steps until the mean fraction of valid samples reached $0.95$. Every run starts from this adapter, and runs with the same seed also share the generation seed.

\paragraph{Runs and metrics.}
The control and arm C at $c_{\rm b}=0.6$ ran with seeds 1 to 6, arm R at $c_{\rm b}=0.6$ with seeds 1, 2 and 4 to 6, and arm R at $c_{\rm b}=2.3$ and arm C at $c_{\rm b}=0.35$ with seed 1. A problem is eligible if its $p_{\rm mode}$ among valid samples is below $0.9$ at step 0, and covered if at least 90 percent of its samples have an answer. F95e is the share of covered eligible problems whose $p_{\rm mode}$ is at least $0.95$, and coverage is the share of eligible problems that are covered.

\paragraph{Results.}
Table~\ref{tab:ttrl} gives the means over runs, and Table~\ref{tab:ttrl-seeds} every run at step 200, including the new seeds of Appendix~\ref{app:seeds} and the coefficient scan of Appendix~\ref{app:sweep}. At step 0, the 19 runs had $0.990$ to $0.994$ valid samples, maj@32 $0.508$ to $0.562$, pass@1 $0.342$ to $0.361$, pass@32 $0.781$ to $0.844$ and nonmode $0.586$ to $0.630$. Generation is not deterministic, and the step-0 evaluations of runs with the same seed were identical on 0 to 1 percent of the problems. The coverage of arm C at $c_{\rm b}=0.6$ with seed 2 fell to $0.65$ at step 160, and that of the control with seed 3 to $0.68$ and $0.69$ at steps 140 and 160. No run collapsed onto a template answer, since the most common majority answer across problems covered at most 9 problems, and 17 in arm R at $c_{\rm b}=2.3$. Two of the six runs of the coefficient scan above a threshold lost the answer format, both with seed 1, against none of the other runs of Table~\ref{tab:ttrl-seeds}.

Without exploration, pass@32 fell from $0.816$ to $0.561$, a narrowing also seen with verifiable rewards \citep{yue2025limit}. Averaged over the six seeds, recovery of wrong majorities was $0.099$ in the control and $0.147$ in arm C at $c_{\rm b}=0.6$, and loss of right majorities $0.147$ and $0.166$. The sampling error is about $0.04$ per run. Changes of the majority answer, $0.346$ and $0.368$, include the sampling noise of the step-0 majority, which we did not measure.

Paired t statistics between arm C at $c_{\rm b}=0.6$ and the control have five degrees of freedom and no correction for multiple tests. They are $-8.7$ for pass@1, $+7.1$ for pass@32 and $+6.1$ for pass@8, with p-values below $0.002$. For maj@32, the loss of right majorities and the recovery of wrong ones, they are $+1.3$, $+1.7$ and $+2.2$, with p-values of about $0.26$, $0.14$ and $0.08$. For F95e at step 200, arm C minus the control is $-0.719$ on average, with a t statistic of $-30.5$, over all six seeds.

\begin{table}[!htbp]
\centering\scriptsize
\caption{Free-form TTRL at steps 0 and 200, means over runs (number in parentheses), 32 samples per problem. Runs that lost the answer format are left out, except in the rows marked all, which count them with their measured accuracy. F95e is the share of eligible problems with at least 90 percent valid samples whose most frequent answer holds at least 95 percent of them. The last two columns come from new samples of the same adapters, six sets for step 0, and were not measured for the coefficient scan. BoN is best-of-32 by a process reward model and SC a Borda vote over self-certainty.}
\label{tab:ttrl}
\begin{adjustbox}{max width=\linewidth}
\begin{tabular}{llllllll}
\toprule
run & F95e & $1-p_{\rm mode}$ & maj@32 & pass@1 & pass@32 & BoN & SC \\
\midrule
step 0 (19) & $0$ & $0.604$ & $0.537$ & $0.354$ & $0.816$ & $0.603$ & $0.535$ \\
$c_{\rm b}=0$ (6) & $0.725$ & $0.057$ & $0.504$ & $0.481$ & $0.561$ & $0.526$ & $0.501$ \\
arm C, $c_{\rm b}=0.35$ (1) & $0.752$ & $0.049$ & $0.508$ & $0.500$ & $0.562$ & $0.516$ & $0.508$ \\
arm C, $c_{\rm b}=0.45$ (3) & $0.630$ & $0.090$ & $0.513$ & $0.493$ & $0.581$ & & \\
arm C, $c_{\rm b}=0.6$ (6) & $0.006$ & $0.382$ & $0.518$ & $0.365$ & $0.710$ & $0.569$ & $0.521$ \\
arm C, $c_{\rm b}=0.8$ (2) & $0$ & $0.631$ & $0.496$ & $0.243$ & $0.762$ & & \\
arm C, $c_{\rm b}=0.8$, all (3) & & & $0.357$ & $0.166$ & $0.536$ & & \\
arm R, $c_{\rm b}=0.6$ (5) & $0.717$ & $0.069$ & $0.500$ & $0.487$ & $0.572$ & $0.528$ & $0.506$ \\
arm R, $c_{\rm b}=1.2$ (3) & $0.481$ & $0.135$ & $0.531$ & $0.481$ & $0.641$ & & \\
arm R, $c_{\rm b}=1.45$ (3) & $0.130$ & $0.336$ & $0.479$ & $0.404$ & $0.708$ & & \\
arm R, $c_{\rm b}=1.8$ (2) & $0$ & $0.723$ & $0.398$ & $0.158$ & $0.695$ & & \\
arm R, $c_{\rm b}=1.8$, all (3) & & & $0.266$ & $0.105$ & $0.464$ & & \\
arm R, $c_{\rm b}=2.3$ (1) & $0$ & $0.866$ & $0.398$ & $0.077$ & $0.633$ & $0.414$ & $0.352$ \\
\bottomrule
\end{tabular}
\end{adjustbox}
\end{table}

\begin{table}[!htbp]
\centering\scriptsize
\caption{Free-form TTRL at step 200 for every run, merged classes. Nonmode is $1-p_{\rm mode}$ among valid samples, averaged over eligible problems with at least 90 percent valid samples, and length is the mean number of tokens. Blank cells are undefined, since no eligible problem had 90 percent valid samples.}
\label{tab:ttrl-seeds}
\begin{adjustbox}{max width=\linewidth}
\begin{tabular}{lllllllllll}
\toprule
run & F95e & coverage & nonmode & maj@32 & pass@1 & pass@8 & pass@32 & valid & truncated & length \\
\midrule
control, seed 1 & $0.757$ & $0.96$ & $0.047$ & $0.453$ & $0.450$ & $0.469$ & $0.492$ & $0.985$ & $0.015$ & $488$ \\
control, seed 2 & $0.761$ & $0.77$ & $0.039$ & $0.508$ & $0.489$ & $0.529$ & $0.562$ & $0.922$ & $0.065$ & $823$ \\
control, seed 3 & $0.674$ & $0.75$ & $0.069$ & $0.539$ & $0.505$ & $0.562$ & $0.602$ & $0.915$ & $0.083$ & $969$ \\
control, seed 4 & $0.793$ & $0.76$ & $0.042$ & $0.500$ & $0.479$ & $0.507$ & $0.523$ & $0.903$ & $0.035$ & $866$ \\
control, seed 5 & $0.702$ & $0.87$ & $0.063$ & $0.492$ & $0.480$ & $0.529$ & $0.586$ & $0.967$ & $0.029$ & $586$ \\
control, seed 6 & $0.663$ & $0.79$ & $0.083$ & $0.531$ & $0.485$ & $0.550$ & $0.602$ & $0.917$ & $0.078$ & $924$ \\
arm C, $c_{\rm b}=0.6$, seed 1 & $0.000$ & $0.93$ & $0.443$ & $0.453$ & $0.331$ & $0.596$ & $0.703$ & $0.962$ & $0.117$ & $1129$ \\
arm C, $c_{\rm b}=0.6$, seed 2 & $0.011$ & $0.77$ & $0.363$ & $0.477$ & $0.312$ & $0.552$ & $0.625$ & $0.935$ & $0.127$ & $1080$ \\
arm C, $c_{\rm b}=0.6$, seed 3 & $0.008$ & $0.98$ & $0.396$ & $0.578$ & $0.406$ & $0.652$ & $0.750$ & $0.989$ & $0.152$ & $1169$ \\
arm C, $c_{\rm b}=0.6$, seed 4 & $0.000$ & $0.76$ & $0.330$ & $0.508$ & $0.390$ & $0.617$ & $0.711$ & $0.916$ & $0.134$ & $1104$ \\
arm C, $c_{\rm b}=0.6$, seed 5 & $0.000$ & $0.68$ & $0.329$ & $0.531$ & $0.387$ & $0.628$ & $0.719$ & $0.896$ & $0.153$ & $1202$ \\
arm C, $c_{\rm b}=0.6$, seed 6 & $0.019$ & $0.87$ & $0.433$ & $0.562$ & $0.366$ & $0.635$ & $0.750$ & $0.956$ & $0.069$ & $919$ \\
arm C, $c_{\rm b}=0.35$, seed 1 & $0.752$ & $0.98$ & $0.049$ & $0.508$ & $0.500$ & $0.538$ & $0.562$ & $0.988$ & $0.012$ & $424$ \\
arm C, $c_{\rm b}=0.45$, seed 1 & $0.761$ & $0.91$ & $0.050$ & $0.484$ & $0.475$ & $0.499$ & $0.516$ & $0.963$ & $0.035$ & $586$ \\
arm C, $c_{\rm b}=0.45$, seed 2 & $0.546$ & $0.98$ & $0.115$ & $0.547$ & $0.510$ & $0.573$ & $0.609$ & $0.993$ & $0.004$ & $652$ \\
arm C, $c_{\rm b}=0.45$, seed 3 & $0.583$ & $0.81$ & $0.105$ & $0.508$ & $0.493$ & $0.563$ & $0.617$ & $0.918$ & $0.083$ & $799$ \\
arm C, $c_{\rm b}=0.8$, seed 1 & $0.500$ & $0.02$ & $0.182$ & $0.078$ & $0.010$ & $0.046$ & $0.086$ & $0.045$ & $0.985$ & $3033$ \\
arm C, $c_{\rm b}=0.8$, seed 2 & $0.000$ & $0.97$ & $0.598$ & $0.555$ & $0.287$ & $0.689$ & $0.828$ & $0.974$ & $0.101$ & $1134$ \\
arm C, $c_{\rm b}=0.8$, seed 3 & $0.000$ & $0.98$ & $0.663$ & $0.438$ & $0.200$ & $0.543$ & $0.695$ & $0.989$ & $0.090$ & $1144$ \\
arm R, $c_{\rm b}=0.6$, seed 1 & $0.770$ & $0.98$ & $0.065$ & $0.531$ & $0.528$ & $0.560$ & $0.594$ & $0.991$ & $0.009$ & $523$ \\
arm R, $c_{\rm b}=0.6$, seed 2 & $0.772$ & $0.83$ & $0.055$ & $0.547$ & $0.526$ & $0.575$ & $0.609$ & $0.947$ & $0.053$ & $627$ \\
arm R, $c_{\rm b}=0.6$, seed 4 & $0.623$ & $0.89$ & $0.079$ & $0.492$ & $0.488$ & $0.528$ & $0.586$ & $0.968$ & $0.032$ & $553$ \\
arm R, $c_{\rm b}=0.6$, seed 5 & $0.747$ & $0.62$ & $0.060$ & $0.477$ & $0.444$ & $0.501$ & $0.523$ & $0.851$ & $0.148$ & $902$ \\
arm R, $c_{\rm b}=0.6$, seed 6 & $0.672$ & $0.95$ & $0.088$ & $0.453$ & $0.448$ & $0.500$ & $0.547$ & $0.988$ & $0.012$ & $527$ \\
arm R, $c_{\rm b}=1.2$, seed 1 & $0.565$ & $0.92$ & $0.105$ & $0.539$ & $0.496$ & $0.586$ & $0.648$ & $0.975$ & $0.023$ & $545$ \\
arm R, $c_{\rm b}=1.2$, seed 2 & $0.360$ & $0.93$ & $0.154$ & $0.508$ & $0.467$ & $0.578$ & $0.648$ & $0.972$ & $0.027$ & $819$ \\
arm R, $c_{\rm b}=1.2$, seed 3 & $0.518$ & $0.91$ & $0.147$ & $0.547$ & $0.480$ & $0.572$ & $0.625$ & $0.975$ & $0.022$ & $485$ \\
arm R, $c_{\rm b}=1.45$, seed 1 & $0.157$ & $0.92$ & $0.325$ & $0.461$ & $0.394$ & $0.575$ & $0.703$ & $0.971$ & $0.044$ & $657$ \\
arm R, $c_{\rm b}=1.45$, seed 2 & $0.129$ & $0.83$ & $0.338$ & $0.500$ & $0.407$ & $0.600$ & $0.719$ & $0.952$ & $0.056$ & $825$ \\
arm R, $c_{\rm b}=1.45$, seed 3 & $0.103$ & $0.88$ & $0.344$ & $0.477$ & $0.411$ & $0.595$ & $0.703$ & $0.958$ & $0.049$ & $776$ \\
arm R, $c_{\rm b}=1.8$, seed 1 & & $0.00$ & & $0.000$ & $0.000$ & $0.000$ & $0.000$ & $0.000$ & $1.000$ & $3072$ \\
arm R, $c_{\rm b}=1.8$, seed 2 & $0.000$ & $0.88$ & $0.737$ & $0.344$ & $0.144$ & $0.465$ & $0.633$ & $0.963$ & $0.267$ & $1620$ \\
arm R, $c_{\rm b}=1.8$, seed 3 & $0.000$ & $0.78$ & $0.708$ & $0.453$ & $0.171$ & $0.549$ & $0.758$ & $0.916$ & $0.157$ & $1292$ \\
arm R, $c_{\rm b}=2.3$, seed 1 & $0.000$ & $0.93$ & $0.866$ & $0.398$ & $0.077$ & $0.379$ & $0.633$ & $0.971$ & $0.403$ & $2189$ \\
\bottomrule
\end{tabular}
\end{adjustbox}
\end{table}

\paragraph{Coverage and interior states.}
Over all problems, the share with at least half valid samples and $p_{\rm mode}$ of at least $0.95$ stays at or below $0.062$ for arm C at $c_{\rm b}=0.6$ from step 100 on, while the controls end at $0.594$ to $0.742$, so the contrast does not depend on the covered problems. In some seeds, the share of samples with an answer fell during training, which the advantages of valid samples and the penalty did not fully prevent within 200 steps. Arm C at $c_{\rm b}=0.6$ kept $p_{\rm mode}$ among valid samples near $0.56$ to $0.67$ in the six seeds, below its predicted interior equilibrium of $0.71$ to $0.88$ (Appendix~\ref{app:bonus}). In the vanilla flow, it approaches this equilibrium by half the time in which $c_{\rm b}=0$ reaches $0.95$, so a slow approach does not explain the gap. Arm R at $c_{\rm b}=2.3$ prevented collapse in a degenerate state with $p_{\rm mode}$ about $0.13$, against a predicted low branch near $0.22$, and $25.9$ distinct answers in 32.

\paragraph{Selection from kept candidates.}
From the adapters of 19 runs and of the warm-up model, we drew 32 new samples per problem with the evaluation settings and new seeds, which reproduced the original evaluations. Qwen2.5-Math-PRM-7B scored every sample with an answer \citep{zhang2025prmlessons}, and the score of a sample is the product of its step scores. This model may have been trained on problems like these, so we compare arms rather than absolute accuracy. Best-of-32 accuracy was $0.569$ with the bonus and $0.526$ without it, higher in five of six seeds (Table~\ref{tab:selection-seeds}). Over seeds, the paired t test gave a p-value of $0.040$ and a sign test one of $0.22$, and the bootstrap interval $[0.009, 0.077]$ resamples problems only. Arm R at $c_{\rm b}=0.6$ gave $0.528$, close to the control, and the untrained model gave $0.603$. Best-of-32 with the bonus minus that before training was $-0.034$, with a p-value of $0.21$. A vote weighted by the process reward model beat majority voting in every arm, and a Borda vote over self-certainty \citep{kang2025selfcertainty} stayed within $0.01$ of it. Among questions whose step-0 majority was wrong, best-of-32 chose a correct answer in 21 percent with the bonus and 15 percent without it. Best-of-32 accuracy fell from $0.603$ before training to $0.526$ without the bonus in all six seeds, and the bonus recovered 56 percent of this loss.

\begin{table}[!htbp]
\centering\scriptsize
\caption{Best-of-32 accuracy by the process reward model, differences per seed over 128 problems.}
\label{tab:selection-seeds}
\begin{tabular}{llllllll}
\toprule
difference & seed 1 & seed 2 & seed 3 & seed 4 & seed 5 & seed 6 & mean \\
\midrule
arm C minus control & $+0.055$ & $-0.031$ & $+0.078$ & $+0.047$ & $+0.047$ & $+0.063$ & $+0.043$ \\
arm R minus control & $+0.039$ & $+0.039$ & & $+0.016$ & $-0.031$ & $-0.031$ & $+0.006$ \\
arm C minus before training & $-0.023$ & $-0.141$ & $+0.008$ & $-0.055$ & $+0.008$ & $0.000$ & $-0.034$ \\
\bottomrule
\end{tabular}
\end{table}

\subsection{Coefficient scan}\label{app:sweep}

In the coefficient scan of Section~\ref{sec:ttrl}, each arm switched from collapse to prevention between the coefficients that bracket its predicted threshold, and arm R did so on the problems predicted to reach an interior equilibrium.

\paragraph{Design.}
Arm R ran at $c_{\rm b}=1.2$, $1.45$ and $1.8$, or $0.77$, $0.93$ and $1.16$ times its threshold near collapse, and arm C at $0.45$ and $0.8$, or $0.88$ and $1.56$ times its own, with seeds 1 to 3. Every run started from the format-trained adapter with the settings of Appendix~\ref{app:ttrl} and differed from the runs there only in the coefficient. Against the control of its seed, a run is collapsed if its F95e at step 200 is at least $0.40$ or at least that of the control minus $0.05$, and prevented if its F95e is at most $0.15$ at every evaluation from step 100 to 200 and at least $0.25$ below the control at step 200. The drift $d$ is the rise of the mean $p_{\rm mode}$ from step 100 to 200 on the eligible problems covered at both steps. Arm R at $1.45$ is also evaluated on the window set, the 104 problems whose start reaches an interior equilibrium of the tabular flow at this coefficient. The coefficient $1.45$ lies within 6 percent of the ends $1.375$ and $1.454$ of the interior branch for free-form answers and for ten answers.

\begin{table}[!htbp]
\centering\scriptsize
\caption{Coefficient scan in free-form TTRL. Each cell gives the outcome of a run at step 200, its F95e and that of the control of the same seed. Multiples in parentheses are of the threshold near collapse $1.554$ for arm R and of $c_{\rm b}^*=0.514$ for arm C.}
\label{tab:sweep}
\begin{adjustbox}{max width=\linewidth}
\begin{tabular}{llll}
\toprule
run & seed 1 & seed 2 & seed 3 \\
\midrule
arm R, $c_{\rm b}=1.8$ ($1.16$) & answer format lost & prevented, $0.000$, $0.761$ & prevented, $0.000$, $0.674$ \\
arm R, $c_{\rm b}=1.45$ ($0.93$), window set & prevented, $0.125$, $0.747$ & prevented, $0.108$, $0.767$ & prevented, $0.114$, $0.667$ \\
arm R, $c_{\rm b}=1.2$ ($0.77$) & collapsed, $0.565$, $0.757$ & in between, $0.360$, $0.761$, $d=0.186$ & collapsed, $0.518$, $0.674$ \\
arm C, $c_{\rm b}=0.8$ ($1.56$) & answer format lost & prevented, $0.000$, $0.761$ & prevented, $0.000$, $0.674$ \\
arm C, $c_{\rm b}=0.45$ ($0.88$) & collapsed, $0.761$, $0.757$ & collapsed, $0.546$, $0.761$ & collapsed, $0.583$, $0.674$ \\
\bottomrule
\end{tabular}
\end{adjustbox}
\end{table}

\begin{figure}[!htbp]
\centering
\includegraphics[width=0.3\linewidth]{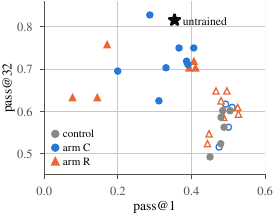}
\caption{Free-form TTRL at step 200 for every run whose answer format held, with the untrained model at step 0. Gray markers are controls, and colored markers are filled at the coefficients that prevented collapse, for arm R at $1.45$ on the window set.}
\label{fig:ttrl-app}
\end{figure}

\paragraph{Results.}
All runs reached step 200 (Table~\ref{tab:sweep}). With seed 1, arm R at $1.8$ and arm C at $0.8$ lost the answer format at steps 160 and 200 after their coverage had fallen, and nearly every sample then reached the limit of 3072 tokens. With seeds 2 and 3, both prevented collapse. Arm C at $0.45$ collapsed in all three seeds, in seed 2 to an F95e of $0.546$, $0.215$ below its control. Arm R at $1.2$ collapsed in seeds 1 and 3 and drifted toward collapse in seed 2, with $d=0.186$ against $0.021$ for arm R at $1.8$. With the runs at $0.6$ of Appendix~\ref{app:ttrl}, the threshold of arm C therefore lies in $(0.45,0.6]$. On the window set, arm R at $1.45$ prevented collapse in every seed, so its threshold lies in $(1.2,1.45]$ there, and the ratio of the two thresholds lies in $(2.0,3.22)$. The bracket $(1.2,1.45]$ of arm R holds the predicted end $1.375$ of the interior branch for free-form answers and excludes the threshold $1.554$ near collapse, as Section~\ref{sec:statedep} predicts for broad starts. Over all problems, arm R at $1.45$ ended with an F95e of $0.157$ in seed 1, one problem above the criterion, and the coefficients $1.2$ and $1.8$ of arm R give a ratio in $(2.0,4.0)$.

Arm R at $1.2$ collapsed more slowly than the control. Its F95e reached $0.40$ at steps 140 and 180 with seeds 1 and 3 and not by step 200 with seed 2, while the controls and arm R at $0.6$ reached it at steps 80 to 120. In the runs whose answer format held, pass@1 fell as the coefficient grew above the threshold (Table~\ref{tab:ttrl}, Figure~\ref{fig:ttrl-app}). Arm R at $1.45$, inside its window, kept a pass@1 of $0.404$, with solutions as long as those of the control.

\paragraph{Problems outside the window set.}
The tabular flow predicts the problems outside the window set to collapse at $c_{\rm b}=1.45$. Their F95e reached only $0.05$ to $0.30$ at step 200, against $0.71$ to $0.80$ for the controls on the same problems. For most of them, $1.45$ lies at $0.93$ to $1.0$ times the end of their interior branch, where the tabular flow collapses slowly, and the others start near collapse. On the window set, F95e also rose, from $0.01$ to $0.05$ at step 100 to $0.11$ to $0.13$ at step 200, so 200 steps do not separate an interior equilibrium from a slow collapse. Averaged over steps 160 to 200, the mean $p_{\rm mode}$ of arm R lay $0.03$ to $0.13$ below the interior equilibrium of the tabular flow at $1.8$, above the threshold near collapse, and within $0.035$ of it at $1.45$, inside the window. The interior equilibrium of arm C falls from $0.71$ at $c_{\rm b}=0.6$ to $0.42$ at $0.8$ for free-form answers, and pass@1 fell with it.

\subsection{Replication with new seeds}\label{app:seeds}

This replication repeats the runs of Table~\ref{tab:regimes} with new seeds. No run violated a boundary, and inside the windows, single seeds moved the thresholds by one grid step.

\paragraph{Design.}
The replication repeats the protocols above with new seeds and changes no code, data, start, hyperparameter or criterion of a run. In the testbed, it repeats the grid at $w=3$ of Appendix~\ref{app:testbed-c1x} and both grids of Appendix~\ref{app:testbed-norm} with seeds 2 to 4, in 336 runs, including the SGD cells that seed 0 had skipped with both terms normalized. In the language model, it repeats the agreement grids of arms U and R with seeds 3 and 4, in 64 runs, and the majority ramps with seeds 2 and 3, in 8 runs. In free-form TTRL, it repeats the control and arms C and R at $c_{\rm b}=0.6$ with seeds 4 to 6, in 9 runs. A run violates an upper boundary if it ends collapsed above it, counted for each start, or if a ramp stays committed above it for five consecutive evaluations. It violates a lower boundary if a start ends broad below it, if an upward ramp stays broad below it for five consecutive evaluations or if a downward ramp is broad at its last evaluation.

\begin{table}[!htbp]
\centering\scriptsize
\caption{Regime tests with new seeds. Runs above the upper boundary, runs below the lower boundary from broad and from collapsed starts, their violations and the thresholds of the new seeds. Ramps count once above and once below, so the 140 entries come from 132 distinct runs, 124 at fixed coefficients and 8 ramps.}
\label{tab:seeds}
\begin{adjustbox}{max width=\linewidth}
\begin{tabular}{lllllll}
\toprule
setting & seeds & above the upper & below the lower, broad & below the lower, collapsed & violations & thresholds of the new seeds, prevention and release \\
\midrule
testbed, $w=3$ & 2 to 4 & $24$ & $6$ & $6$ & $0$ & SGD $0.16$ and $0.26$ to $0.28$, AdamW $0.18$ to $0.20$ and $0.28$ \\
testbed, both normalized & 2 to 4 & $24$ & $12$ & $12$ & $0$ & SGD $0.14$ to $0.16$ and $0.28$, AdamW $0.16$ to $0.18$ and $0.26$ to $0.28$ \\
LLM agreement, arm U & 3, 4 & $8$ & $2$ & $2$ & $0$ & $0.14$ and $0.18$, $0.14$ and $0.14$ \\
LLM agreement, arm R & 3, 4 & $16$ & $6$ & $6$ & $0$ & $1.2$ and $1.2$, $1.4$ and $1.2$ \\
LLM majority, arm U, ramps & 2, 3 & $4$ & $2$ & $2$ & $0$ & collapse $0.600$, $0.476$, release $0.720$, $0.728$ \\
LLM majority, arm R, ramps & 2, 3 & $4$ & $2$ & $2$ & $0$ & collapse $1.218$, $1.271$, release $1.533$, $1.533$ \\
\midrule
total & & $80$ & $30$ & $30$ & $0$ & \\
\bottomrule
\end{tabular}
\end{adjustbox}
\end{table}

\paragraph{Results.}
No run among the 132 violated a boundary (Table~\ref{tab:seeds}), including the cells closest to a boundary. These are $c=0.14$ at $w=3$, $2.2$ percent below the lower boundary, $c=0.28$ with both terms normalized, $1.4$ percent above the upper one, and $c=1.6$ in arm R, $0.5$ percent above it. At the fold of arm R, the branch of one leading answer has $\pmax=0.247$, below the criterion $0.3$, so a committed state at this coefficient would not count as collapsed. The runs at $c=1.6$ nevertheless ended with mean $\pmax$ between $0.138$ and $0.150$ in every seed and start, below this branch. No run of these tests ended with a fraction of collapsed states or questions between $0.3$ and $0.7$. Inside the windows, single seeds moved the thresholds by one grid step. In the testbed, the seed means of the thresholds at $w=3$ equaled those of seeds 0 and 1. With both terms normalized, every pair of new seed and optimizer showed a window, and with only the advantages normalized, the two starts differed by at most $0.024$.

In the majority ramps, release came 3 to 6 percent below the folds, and arm R kept its loop in both seeds. With seed 3, the mean $\pmax$ of arm U lay $0.152$ below the branch at $c=0.697$, four evaluations before release, likely because its questions were released one at a time.

In free-form TTRL, arm C kept F95e at or below $0.057$ from step 100 on in all three new seeds, while the controls ended at $0.663$ to $0.793$ and arm R at $0.623$ to $0.747$, and every run kept a coverage of at least $0.62$ from step 100 on. Merging equivalent answers did not change the F95e of arm C from step 100 on.

\section{Additional numerics}\label{app:additional}\label{app:numerics}

Table~\ref{tab:capped-adam} compares update rules that bound the speed.\label{app:numerics-outside}

\begin{table}[!htbp]
\centering\footnotesize
\caption{Discrete Adam, sign and SGD with the leak applied after the optimizer step, on a line of $K=512$ bins with vanilla policy gradient, the agreement reward and $\sigma=\omega$. Peak height $K\pmax$ and position, with bd.\ for a peak at the boundary and int.\ for a peak in the interior. The Adam rows give the learning rate $\eta$ and $(\beta_1,\beta_2)$.}
\label{tab:capped-adam}
\begin{adjustbox}{max width=\linewidth}
\begin{tabular}{llllllll}
\toprule
optimizer & $\sigma=0.003$ & $0.01$ & $0.03$ & $0.05$ & $0.097$ & $0.15$ & $0.3$ \\
\midrule
Adam, $\eta=0.01$, $(0.9,0.999)$ & $12.7$ (bd.) & $7.07$ (bd.) & $3.94$ & $3.05$ & $2.29$ & $1.94$ & $1.46$ \\
Adam, $(0,0.999)$ & $12.3$ & $6.56$ & $4.23$ & $3.11$ & $2.20$ & $1.78$ & $1.45$ \\
Adam, $(0.9,0.99)$ & & & $3.98$ & $3.14$ & $2.29$ & $1.87$ & $1.43$ \\
Adam, $\eta=0.003$ & $12.4$ & $6.98$ & $3.90$ & $3.23$ & $2.19$ & $1.85$ & $1.43$ \\
Adam, $\eta=0.04$ & $9.39$ (int.) & $6.92$ & $4.02$ & $3.20$ & $2.17$ & $1.83$ & $1.44$ \\
sign, $\eta=0.01$ & $6.37$ (int.) & $3.49$ (int.) & $4.03$ & $3.13$ & $2.26$ & $1.86$ & $1.44$ \\
SGD (ideal flow) & collapse & $109$ & $114$ (bd.) & $90$ (bd.) & $2.40$ (pattern) & $1$ & $1$ \\
\bottomrule
\end{tabular}
\end{adjustbox}
\end{table}

\subsection{Hysteresis of a leak}\label{sec:hysteresis}\label{app:prereg-hyst}\label{app:hysteresis-full}

Under vanilla policy gradient, a leak bypasses the geometry. Numerical continuation of the lattice equation \eqref{eq:lattice} with $q=1$ then gives a window. Below its onset, broad data collapse, and above the fold of the branch of spikes, a collapsed policy is released. We tested this window on a ring of $K=64$ bins, with $K=128$ as a check, under vanilla policy gradient and the agreement reward. 

One step updates the logits and then leaks,
\begin{equation}
z\leftarrow z+\eta\hat V,\qquad p_i\leftarrow(1-2\varepsilon)p_i+\varepsilon(p_{i-1}+p_{i+1}),
\end{equation}
with $\hat V$ the expected or sampled gradient. The broad start is a Gaussian peak of width $0.15$ centered at $0.4$, and the collapsed start has $0.99$ in one bin and $0.005$ in each neighbor. After $3\times10^5$ steps, a state with $\pmax>0.2$ counts as collapsed. The prevention and release thresholds are the smallest values of exploration at which the broad start does not collapse and the collapsed start is released. They are bisected in $\omega$, or in $\varepsilon$ for Adam, with a majority over three seeds for sampled flows.

V1 uses the expected gradient with SGD at $\eta=1$, rechecked at $\eta=0.25$. V2 samples groups of $G=16$ actions with the leave-one-out reward. V3 adds group normalization, and V4 replaces SGD in V3 by Adam at $0.01$.

With expected gradients, the prevention threshold is $0.0240$, within one percent of the predicted $0.0239$, at both learning rates. At $\eta=1$, the release threshold lies nine percent below the predicted $0.0422$, the fold of the flow. The fold of the fully discrete map is $0.03823$ at $\eta=1$ and $0.04115$ at $\eta=0.25$, against the measured $0.0383$ and $0.0411$. On a ring of 128 bins, the thresholds were $0.0236$ and $0.0383$. Sampled rewards keep the hysteresis with a ratio of $1.40$, normalization lets the policy collapse even at $\varepsilon=0.5$, and Adam removes the hysteresis, with thresholds $0.01378$ and $0.01381$.

\paragraph{Unordered answers.}
Two results for unordered answers held in simulation. Temperature only rescaled time for $\tau\in\{0.6,1,1.2,2\}$, in 62 of 64 runs (Proposition~\ref{prop:temperature}). The Potts window $(1/n,c_{\rm H}^{\rm s}(n))$ for the agreement reward and the fixed point for the majority reward of large groups held for $n=4$, $16$, $64$ and $256$ (Appendix~\ref{app:potts}).

\subsection{Numerical methods}\label{app:details-numerics}

The lattice flows are integrated by explicit Euler steps of adaptive size, set by the largest logit velocity on bins with non-negligible mass. On a line, diffusion is treated implicitly by the discrete cosine transform. Lattice branches are continued in the peak value $\pmax$, with stability from the Jacobian on the tangent space of the simplex, computed in double precision. For a broad peak, the eigenvalue of the mode that shifts the peak is exponentially small in its width, so this computation can miss an unstable direction of a wide peak. We report only unstable directions that double precision resolves, and the statements about broad data below $c^*$ rest on direct simulations (Appendix~\ref{app:inloss-dial}).

\paragraph{Exploration in the loss.}\label{app:numerics-inloss}
The kernel $D_w$ uses the periodic distance on the ring. Explicit Euler steps on the logits use the step $0.5/(\pmax^2+2c\,\pmax)$ and run for $40K^2$ time units or $6\times10^6$ steps. After this time, the residual $\max_i|\partial_tp_i|/\pmax$ was below $4\times10^{-8}$ in every reported run. Broad data are a Gaussian of width $0.1$. Collapsed data have $0.99$ in one bin, $0.005$ in each neighbor and $10^{-8}$ elsewhere before normalization. A state counts as collapsed when $\pmax>0.2$.

\section{Further related work}\label{app:related}

\paragraph{Self-referential rewards for language models.}
In the limit of large groups, the majority reward of TTRL \citep{zuo2025ttrl} and of self-rewarded training \citep{shafayat2025srt} is the indicator $R_i=[i=\arg\max_jp_j]$. The likelihood of a semantic cluster \citep{zhang2025empo} is the agreement reward on clusters. Offline, fine-tuning on consistent answers \citep{huang2023selfimprove} and preference optimization toward them \citep{prasad2025scpo} are not policy flows, but each moves probability toward the current majority. Self-rewarding language models \citep{yuan2024selfrewarding} judge their own answers, which lies outside our model, and consensus rewards have also been derived from an equilibrium \citep{mehrish2026core}.

Several works observe the collapse that our theory describes. Self-rewarding baselines reach trivial consensus solutions \citep{zhang2025corewarding}, and self-improvement sharpens the policy toward its own mode \citep{huang2025sharpening}. Intrinsic rewards without labels drive the policy toward its initial majority \citep{he2026unsupervised}, and test-time training on self-consistency tilts answers toward their mode \citep{corderoencinar2025certified}. Concurrent work attributes the collapse that appears only with many training problems to the coupling of the parameters they share \citep{ding2026collective}. It stabilizes the uniform collective mode with a bonus on the entropy of the marginal. Our theory treats one problem at a time and leaves this coupling out. Asymmetric clipping keeps low-probability tokens alive \citep{yu2025dapo}, and exploration for reasoning has been added as bonuses on outcomes \citep{song2025outcome}. With a verifier, differential smoothing treats correct and incorrect answers differently and improves accuracy and diversity \citep{gai2026differential}. Random or spurious rewards also improve Qwen2.5-Math models \citep{shao2026spurious}, because the clipping of GRPO favors behaviors the model already prefers, an effect of finite steps that our expected flows leave out. Gains under self-referential rewards should therefore be checked across models. Token-level methods, such as the covariance-based clipping and KL penalty of \citet{cui2025entropy}, work below the level of answers, outside the single softmax that our theory treats.

\paragraph{The normalization of GRPO.}
GRPO divides the advantages of a group by the standard deviation of its rewards \citep{shao2024deepseekmath}, and the division decides whether an exploration term keeps its thresholds (Section~\ref{sec:statedep}). Dividing the advantages by the standard deviation of the group can speed convergence \citep{ge2026grponorm}, makes predictions overconfident on stochastic outcomes \citep{bereket2025uncalibrated} and inflates the advantages of a weak consensus \citep{yan2026scrl}. Removing it can raise diversity \citep{li2025darling}. After the division, GEPO shrinks advantages according to the entropy of the group \citep{cheng2026gepo}, and ExPPO reshapes them with the surprisal of the rollout policy and the pass rate of the prompt \citep{jin2026exppo}. DARE adds the bonus $\alpha(1-n(y)/M)(1-u(y))$, with $n(y)$ the count and $u(y)$ the mean token entropy of answer $y$ among $M$ rollouts, to a reward that GRPO then normalizes \citep{du2026dare}. Its bonus thus enters before the division, as in our arm C, and its weights by entropy do not treat answers alike. A bonus added after a normalization over the batch, as REINFORCE++ uses \citep{hu2025reinforcepp}, keeps one threshold for broad and collapsed policies, divided by a batch statistic that one question hardly moves (Proposition~\ref{prop:norm}(b)). This statistic still falls as all questions concentrate. A bonus normalized by its own statistics, as in PAPO \citep{tan2026papo}, is standardized like the reward near collapse. The sign computation of Section~\ref{sec:statedep} then gives the threshold $c_{\rm b}=1$ there for every $G$, between those of arms C and R.

\paragraph{Methods that break the symmetry.}\phantomsection\label{app:related-symmetry}
Several recent methods for TTRL do not treat the answers alike. T3RL weights the vote by tool verification, such as code execution, so that a verified answer can outvote a more frequent one \citep{liao2026t3rl}. SERPO replaces the vote by rubrics that it evolves from its own rollouts, for open-ended tasks without a single answer \citep{wang2026serpo}. Learned rankers and selectors that use a graph of rollouts also choose the pseudo-label from the content of the answers \citep{marina2026risc,mehrish2026core}. Each of these signals depends on the content of an answer, not only on its count, so it can change the leading answer. Our theory does not say how reliable such a signal is. SPINE updates only high-entropy forking tokens and keeps their entropy within a band \citep{wu2025spine}, and $\lambda$-GRPO learns how the advantage of a response is spread over its tokens \citep{wang2025lambdagrpo}. Both act below the level of answers, where the tabular flow of one softmax over answers does not apply, and their weights need not treat answers alike. TTRL-Guard scales down the update of a problem as its flip rate declines, preserves samples of minority answers and suspends updates on polarized problems \citep{lin2026extinction}. As far as its description shows, these rules use the votes only through statistics that do not change when the answers are relabeled. A positive weight on the whole update of a problem only rescales time, as in Theorem~\ref{thm:std}, and a rule that treats the answers alike commutes with their permutations. In the tabular flow, both keep the order of the answers (Proposition~\ref{prop:order}), so they can delay commitment but not change the leading answer.

\paragraph{Vision-language-action models.}
VLA models such as RT-2 \citep{brohan2023rt2} and OpenVLA \citep{kim2024openvla} predict each action dimension as one of 256 ordered bins, as in our lattice equations, with clipping as the reflecting boundary. Reinforcement learning of VLA models uses task success with GRPO and a raised temperature and upper clip \citep{li2025simplevlarl}, or a learned process reward with PPO \citep{lu2025vlarl}. At test time, TT-VLA \citep{ttvla2026} rewards estimated progress, and T$^2$VLA \citep{t2vla2026} rewards similarity to its own confident trajectories, a self-referential reward.

\paragraph{Collapse under self-training.}
Generative models trained recursively on their own outputs lose the tails of the distribution \citep{shumailov2024collapse}, with a threshold in the proportion of fresh data \citep{bertrand2024stability} rather than in exploration. Self-play also collapses once the gate on its training data admits too many tasks whose execution failed \citep{pu2026survive}.

\paragraph{Learning dynamics.}
Policy gradient methods converge globally under external rewards \citep{mei2020softmax,cen2022fast}, and softmax policy gradient does so at a speed that depends strongly on the initialization \citep{mei2020escaping}. A self-referential reward moves with the policy, so the question is which equilibria exist and which are reached. As a gradient flow, GRPO with a noisy verifier has a transition where Youden's index is zero \citep{rad2026rate}. With a verifier and a KL penalty, iterated GRPO moves the probability of success to a fixed point set by the reference and the penalty, with or without the variance in its normalization \citep{mroueh2026grpo}.

\paragraph{Selection, mutation and Potts models.}
Under NPG, the agreement reward with smoothing in the loss gives the flow $\partial_tp_i=p_i(p_i-S_2)+c\,((Dp)_i-p_i)$ of the probabilities, the replicator equation of the pure coordination game with an additive mutation term. Self-referential fitness also appears in the language dynamics of \citet{komarova2001grammar}, and for random fitness matrices, its coherence threshold tends to a nonzero constant as the number of grammars grows \citep{komarova2004replicator}. In the Shahshahani geometry, Fisher's fundamental theorem survives when mutation rates depend only on the target, and other rates allow stable limit cycles \citep{hofbauer1985selection}. Under vanilla policy gradient, a leak between neighboring bins has rates of the second kind, and no Lyapunov function is known. Under NPG, a leak and a smoothing term in the loss give the same flow, with the potential $F_c$. In the mean-field Potts model with three or more states, the transition is first order \citep{wu1982potts}, and in a window around it, the disordered and ordered phases are both stable. Theorem~\ref{thm:potts} is an instance of this window and Proposition~\ref{prop:potts-uniform} an analogue. Coordination with learning errors has such a window too \citep{komarova2001grammar,mitchener2003bifurcation}, and in smooth Q-learning, exploration changes the number and stability of equilibria \citep{tuyls2003selection,kianercy2012boltzmann} and, on weighted potential games, can select among them \citep{leonardos2021exploration}. \citet{mitchener2003bifurcation} classifies the fixed points and proves convergence when learning errors scale with the mean fitness, which is our smoothing with a policy-dependent coefficient, as in Proposition~\ref{prop:norm}(b). In replicator equations in continuous space with diffusion inside the fitness, interior stationary solutions are uniform once every diffusion coefficient exceeds the largest eigenvalue of the interaction matrix over the first Neumann eigenvalue \citep{bratus2014replicator}. Computer-assisted proofs of saddle-node bifurcations in lattice dynamical systems \citep{sander2016validated} are the closest precedent for Theorem~\ref{thm:inloss-cap}. These results do not cover the path of the reward, that is, how the geometry, the preconditioner and the normalization of a learning algorithm move or remove the window.

\section{A practical guide}\label{app:guide}

For setting exploration in practice, Appendix~\ref{app:guide-recipe} gives a recipe with the boundaries of Table~\ref{tab:guide-boundaries}, and Appendix~\ref{app:guide-cap} explains what the certified folds show.

\subsection{A recipe}\label{app:guide-recipe}

\begin{enumerate}[leftmargin=*,itemsep=1pt]
\item \emph{Decide what exploration should do.} Exploration that treats the answers alike can keep a question from collapsing and keep candidates among the samples. It cannot change the leading answer in the tabular flow (Proposition~\ref{prop:order}, Corollary~\ref{cor:plurality}). If the initial plurality is often wrong and the goal is a higher pass@1, add a signal that breaks the symmetry (Appendix~\ref{app:related-symmetry}).
\item \emph{Keep exploration on the path of the reward.} Add a bonus to the reward before the normalization, as in arm C, or divide a term of the loss by the same statistic as the reward, as in arm B. Training without the division, as in Dr.\ GRPO, also keeps exploration on the path. A bonus after the division, a KL penalty outside it or a leak after the optimizer step has a threshold that changes with the policy or the learning rate (Section~\ref{sec:offpath}).
\item \emph{Look up the boundaries of the setting.} Table~\ref{tab:guide-boundaries} collects them. For other numbers of answers or group sizes, the exact sums of Appendix~\ref{app:majority-exact} give them on policies with one leading answer.
\item \emph{Set the coefficient above the lower boundary, with a margin, from the start.} Inside the window, the outcome depended on the start, the seed, the learning rate and the optimizer (Section~\ref{sec:reading}). The coefficient that prevents commitment grows with the leading answer (Proposition~\ref{prop:potts-invariant}), so exploration costs least early. In the testbed, the measured thresholds lay 3 to 29 percent above the folds, and in free-form TTRL, arm C prevented collapse at $1.17$ times its threshold.
\item \emph{Do not count on other settings.} Temperature only rescales time, clipping and Adam move no equilibrium of exploration in the loss, and the KL coefficient $0.001$ of TTRL lies far below the boundaries (Table~\ref{tab:remedies}).
\item \emph{Monitor every question.} Track the largest answer probability of each question, the share of questions above a cutoff such as $0.95$ and the share of samples with an answer. A drop in coverage can hide collapse (Section~\ref{sec:ttrl}). Many questions collapsing onto the same answer at once point to shared parameters, which the tabular flow leaves out \citep{ding2026collective}.
\end{enumerate}

\begin{table}[!htbp]
\centering\scriptsize
\caption{Boundaries of the exploration coefficient in the tabular flow, without normalization unless stated. Prevention needs at least the lower boundary, and release follows above the upper one. The majority reward and the bonus are computed on policies with one leading answer, with $c_{\rm b}^*=1/\log(G-1)$, and the threshold of arm R near collapse comes from the sign of a drift.}
\label{tab:guide-boundaries}
\begin{tabular}{@{}>{\raggedright\arraybackslash}p{0.33\linewidth}>{\raggedright\arraybackslash}p{0.22\linewidth}>{\raggedright\arraybackslash}p{0.22\linewidth}>{\raggedright\arraybackslash}p{0.15\linewidth}@{}}
\toprule
setting & lower boundary & upper boundary & source \\
\midrule
agreement reward, smoothing toward $\bar p$, $n$ answers & $\frac1n$ & $\frac n{4(n-1)}$ & Proposition~\ref{prop:potts-uniform} \\
same, leave-one-out reward of groups of $G$ & $\frac{G-2}{G}\cdot\frac1n$ & $\frac{G-2}{G}\cdot\frac n{4(n-1)}$ & Section~\ref{sec:unordered} \\
majority reward, $n=10$, $G=16$, smoothing toward $\bar p$ & $0.363$ & $0.766$ & Appendix~\ref{app:majority-exact} \\
agreement reward, entropy bonus under NPG & $\frac1n$ & $c_{\rm H}^{\rm s}(n)$, $0.193$ for $n=16$ & Theorem~\ref{thm:potts} \\
bonus before the division, arm C & $c_{\rm b}^*$, $0.514$ for $G=8$ & $c_{\rm b}^*$ & Section~\ref{sec:statedep} \\
bonus after the division, arm R & end of the interior branch, $2.3$ to $3.2\,c_{\rm b}^*$ & about $\frac{G}{\sqrt{G-1}}\,c_{\rm b}^*$ & Table~\ref{tab:bonus-main} \\
Gaussian kernel of width $w$ on ordered bins & at most $c^*$, $0.1477$ for $w=3$ & at least $c_{\rm rel}$, $0.2919$ for $w=3$ & Theorem~\ref{thm:inloss-cap} \\
leak on a line of $K$ bins & $\omega_{\rm lin}(K)$ in $\varepsilon/\eta$ for the uniform policy & not given & Theorem~\ref{thm:D5} \\
\bottomrule
\end{tabular}
\end{table}

\subsection{What the certified folds show}\label{app:guide-cap}

On ordered bins, the equilibria of \eqref{eq:inloss} have no closed form, so Theorem~\ref{thm:inloss-cap} rests on computation. At a fold, a branch of equilibria turns back, and a stable equilibrium meets one with one unstable direction. The lower fold $c^*$ ends the branch of broad peaks, and the upper fold $c_{\rm rel}$ ends the branch of spikes. Floating point solves the equations of an equilibrium only approximately, so we use ball arithmetic, which carries a rigorous error bound with every number \citep{johansson2017arb}. The Krawczyk test then shows that a tiny box around the numerical fold contains exactly one exact solution of the equations that define a fold \citep{krawczyk1969,rump2010}, and further interval computations certify the signs of the eigenvalues there (Appendix~\ref{app:inloss-proofs}). The result therefore concerns the exact equation \eqref{eq:inloss} rather than a simulation of it. The enclosure of each fold has a radius below $10^{-39}$, rings and lines of different sizes give the same folds to $10^{-27}$, and each certificate takes at most 11 seconds. The certificate is local and covers states symmetric about one bin, so the claims that broad data reach the certified branch and that no other state lowers the boundaries rest on simulations (Appendix~\ref{app:inloss-dial}).

\section{Open problems}\label{app:open}

\begin{enumerate}[leftmargin=*,itemsep=0pt,parsep=0pt,topsep=0pt]
\item Prove that broad data reach the broad branch for every $c$ above $c^*$, so that the folds are thresholds of the dynamics.
\item Show that stability transfers from the tabular flow to networks on ordered bins, and connect the regimes of one state to the collective instability of states that share parameters \citep{ding2026collective}.
\item Treat a KL penalty toward a nonuniform reference policy, which breaks the symmetry of the answers.
\item Compare symmetric exploration with methods that break the symmetry on problems whose initial plurality is right and wrong, at matched compute.
\end{enumerate}

\end{document}